\documentclass[acmsmall,screen,nonacm]{acmart}

\AtBeginDocument{%
  }

\setcopyright{acmlicensed}
\copyrightyear{2026}
\acmYear{2026}
\acmMonth{5}
\acmVolume{}
\acmNumber{}
\acmArticle{}
\acmDOI{XXXXXXX.XXXXXXX}
\acmConference[Conference acronym 'XX]{Make sure to enter the correct
  conference title from your rights confirmation email}{June 03--05,
  2018}{Woodstock, NY}
\acmISBN{978-1-4503-XXXX-X/2018/06}

\AtBeginDocument{%
  \fancypagestyle{firstpagestyle}{%
    \fancyhf{}%
    \fancyfoot[R]{\footnotesize Preprint, May 2026}%
  }%
}

\newcommand{\dataset}{\mathcal{D}}
\newcommand{\devset}{\dataset_\text{dev}}
\newcommand{\fewshotset}{\dataset_\text{shots}}
\newcommand{\testset}{\dataset_\text{test}}

\newcommand{\llm}{\Phi}
\newcommand{\metallm}{\llm_{\text{meta}}}
\newcommand{\evalllm}{\llm_{\text{eval}}}

\newcommand{\instruction}[1]{i_{#1}}
\newcommand{\instructionset}{\mathcal{I}}
\newcommand{\initialinstructions}{\instructionset_0}
\newcommand{\offspringinstruction}{\instruction{\text{off}}}
\newcommand{\mutatedinstruction}{\instruction{\text{mut}}}

\newcommand{\prompt}[1]{p_{#1}}
\newcommand{\promptset}{\mathcal{P}}
\newcommand{\population}{\promptset_{\mu}}
\newcommand{\offspringprompt}{\prompt{\text{off}}}
\newcommand{\offspringpromptset}{\promptset_\text{off}}
\newcommand{\mutatedprompt}{\prompt{\text{mut}}}
\newcommand{\mutatedpromptset}{\promptset_\text{mut}}
\newcommand{\crossoverprompt}{\prompt{C}}
\newcommand{\mutationprompt}{\prompt{M}}
\newcommand{\challenger}{\prompt{\text{chal}}}
\newcommand{\incumbent}{\prompt{\text{inc}}}

\newcommand{\paretoset}{\promptset_\text{par}}
\newcommand{\paretofront}{\mathcal{F}_\text{par}}
\newcommand{\paretosetapprox}{\hat{\promptset}_\text{par}}
\newcommand{\paretofrontapprox}{\hat{\mathcal{F}}_\text{par}}
\newcommand{\incset}{\promptset_\text{inc}}

\newcommand{\runhistory}{\mathcal{H}}

\newcommand{\fewshot}[1]{e_{#1}}
\newcommand{\fewshots}[1]{\boldsymbol{e_{#1}}}
\newcommand{\offspringfewshots}{\fewshots{\text{off}}}
\newcommand{\fewshotspace}{\mathcal{E}}
\newcommand{\nshots}{k}
\newcommand{\maxshots}{\nshots_{\max}}

\newcommand{\populationsize}{\mu}
\newcommand{\blocksize}{b}
\newcommand{\blockset}{\mathcal{B}}
\newcommand{\block}[1]{B_{#1}}
\newcommand{\blockscommon}{\blockset_{\cap}}
\newcommand{\blockschal}{\blockset_{\text{chal}}}
\newcommand{\blocksmin}{\blockset_{\text{min}}}
\newcommand{\blockstoeval}{\mathcal{B}_{\text{rest}}}
\newcommand{\maxblockevals}{z_{\max}}
\newcommand{\niters}{T}
\newcommand{\iter}{t}
\newcommand{\ncrossovers}{c}

\newcommand{\scorefunction}{\sigma}
\newcommand{\costfunction}{c}

\newcommand{\lengthpenalty}{\gamma}

\newcommand{\weightin}{w_{\text{in}}}  %
\newcommand{\weightout}{w_{\text{out}}}  %

\newcommand{\HyperCall}[3]{%
  \hyperlink{#3}{\Call{#1}{#2}}%
}

\DeclareMathOperator*{\argmax}{arg\,max}
\DeclareMathOperator*{\argmin}{arg\,min}

\newcommand{\mocaponsgaII}{\texttt{NSGA-II-PO}}
\newcommand{\mocapointensify}{\texttt{MO-CAPO}}
\newcommand{\mistral}{Mistral-3.2-24B}
\newcommand{\qwen}{Qwen3-30B}
\newcommand{\gpt}{GPT-OSS-120B}

\definecolor{CapoGreen}{HTML}{1b9e77}
\definecolor{EvoPurple}{HTML}{7570b3}

\newcommand{\findingbox}[1]{
\vspace{5pt}
\begin{tikzpicture}
\node [fill=white, draw=black, rounded corners=0.5em] {
\begin{tabular}{p{0.9\linewidth}}
\textit{\textbf{Finding:} #1}
\end{tabular}
};
\end{tikzpicture}
}

\usepackage{algorithm}
\usepackage{algorithmicx} 
\usepackage{booktabs}
\usepackage{multirow}
\usepackage{float}
\usepackage{adjustbox}
\usepackage[noEnd=true,indLines=true,italicComments=true]{algpseudocodex}
\usepackage{subcaption}
\usepackage{wrapfig}
\usepackage{tabularx}
\usepackage{enumitem}
\usepackage{tikz}
\usepackage{siunitx}
\usepackage{placeins}

\begin{document}

\title{MO-CAPO: Multi-Objective Cost-Aware Prompt Optimization}

\author{Jan B{\"u}ssing}
\affiliation{%
  \institution{LMU Munich}
  \city{Munich}
  \country{Germany}}
\email{jan.buessing@campus.lmu.de}

\author{Moritz Schlager}
\authornote{These authors contributed equally.}
\affiliation{%
 \institution{Technical University of Munich (TUM), Munich Center for Machine Learning (MCML)}
 \city{Munich}
 \country{Germany}}
\email{moritz.schlager@tum.de}

\author{Timo Heiß}
\authornotemark[1]
\affiliation{%
  \institution{LMU Munich, Munich Center for Machine Learning (MCML)}
  \city{Munich}
  \country{Germany}}
\email{timo.heiss@stat.uni-muenchen.de}

\author{Tom Zehle}
\authornotemark[1]
\affiliation{%
  \institution{University of Freiburg, ELLIS Institute Tübingen}
  \city{Freiburg}
  \country{Germany}}
\email{tom.zehle@tue.ellis.eu}

\author{Matthias Feurer}
\affiliation{%
  \institution{TU Dortmund University, Lamarr Institute for Machine Learning and Artificial Intelligence}
  \city{Dortmund}
  \country{Germany}}
\email{matthias.feurer@tu-dortmund.de}

\renewcommand{\shortauthors}{Büssing et al.}

\begin{abstract}
Large language models (LLMs) achieve strong performance across a wide range of tasks but are highly sensitive to prompt design, motivating the need for automatic prompt optimization. Existing methods predominantly focus on performance alone, ignoring competing objectives such as inference cost or latency. At the same time, existing work on multi-objective prompt optimization relies on off-the-shelf NSGA-II, ignoring optimization efficiency. As a remedy, we introduce \mocapointensify{}, a novel multi-objective prompt optimization algorithm that jointly optimizes performance and inference cost while leveraging budget allocation for cost-efficient optimization. We further propose a deployment-oriented cost objective that captures the full computational profile of LLM inference.
We evaluate our approach across four tasks and three LLMs and compare it to an NSGA-II-based multi-objective method and state-of-the-art single-objective prompt optimizers.
Results show that \mocapointensify{} consistently identifies strong, robust, and diverse Pareto front approximations while maintaining cost-efficiency.
It outperforms the NSGA-II baseline on 8 out of 12 cases in terms of the noisy R2 metric and achieves competitive performances often already at a considerably lower budget. The discovered solution sets span diverse performance-cost trade-offs that are omitted by single-objective optimizers, yet the top-performance candidates remain competitive with single-objective solutions. 
Additionally, we conduct the first evaluation of multi-objective machine learning experiments that considers generalization and robustness through noisy R2 and approximation gap, enabling a more realistic assessment of solution quality.
\mocapointensify{} enables practitioners to select from an efficiently discovered set of multiple prompts offering different trade-offs between performance and cost.
\end{abstract}


\setcopyright{none}
\settopmatter{printacmref=false, printccs=false, printfolios=true}
\renewcommand\footnotetextcopyrightpermission[1]{}
\authorsaddresses{}
\maketitle

\section{Introduction}

Nowadays, large language models (LLMs) yield impressive results on a wide variety of tasks~\citep{radford-openai19a,ouyang-neurips22a,touvron-arxiv23a}.
They can be adapted to solve new tasks simply through a \textit{prompt} as input, consisting of a textual instruction and optionally few-shot examples~\citep{brown-neurips20a,liu-acmcs23a}.
However, the performance of an LLM on a given task is highly sensitive to the quality and formulation of this prompt, as well as the selection and order of few-shot examples~\citep{zhao-icml21b,lu-aclam22a,zhou-iclr23a}.
Optimizing prompts can therefore substantially improve performance~\citep{zehle-automl25a}.
Because manual prompt engineering is tedious and unreliable~\citep{jiang-acl20a,liu-acmcs23a}, the field of automatic prompt optimization has emerged~\citep{cui-acl25a,li-arxiv25a}.

Most existing prompt optimization algorithms focus solely on optimizing performance.
In practice, however, there are often several competing objectives for which we want to optimize prompts, such as cost or latency, in addition to performance~\citep{murthy-arxiv25a}.
At the same time, the optimization process itself can be costly, as evaluating prompts requires numerous LLM calls~\citep{agarwal-arxiv24a}.
CAPO~\citep{zehle-automl25a} recognizes both challenges, but primarily addresses the cost-efficiency of the optimization process through a racing mechanism, and considers the cost of a prompt only by penalizing its length. This penalty is controlled by an \textit{a priori}-chosen weight, and choosing an appropriate weight is a non-trivial task for users.
Existing multi-objective prompt optimizers mostly rely on off-the-shelf NSGA-II to discover trade-offs between multiple objectives, such as performance and prompt length, and are not cost-efficient in their optimization procedures~\citep{yang-emnlp23a,camara-stil25a,lopes-arxiv25a}.
In addition, costs are not only produced by prompt length. With the rise of thinking models that leverage internal reasoning~\citep{guo-nature25a,jaech-arxiv24a} and prompting techniques such as chain-of-thought~\citep{wei-neurips22a}, output tokens have become a considerable driver of cost and latency.

\begin{wrapfigure}{r}{0.45\textwidth}
    \centering
    \includegraphics[width=0.45\textwidth]{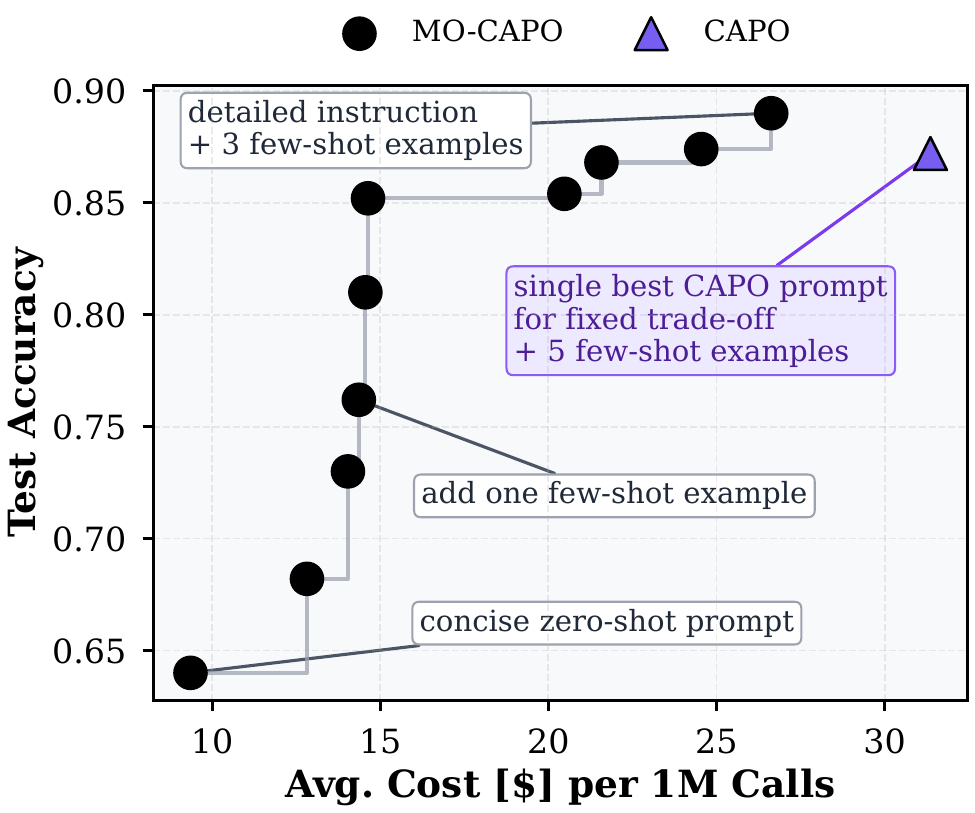}
    \caption{Prompts for \mistral{} on Subj: CAPO is limited to a single solution, whereas \mocapointensify{} discovers a Pareto front. The x-axis denotes the average cost in US dollars per 1M calls with a prompt, the y-axis its test set accuracy.}
    \label{fig:intro_pareto}
    
\end{wrapfigure} 

From an optimization perspective, prompt optimization can naturally be viewed as an expensive black-box optimization problem~\citep{zhou-iclr23a,cheng-acl24}.
The objective function (LLM performance under a given prompt) can only be evaluated through costly LLM calls, while its internal structure remains unknown.
This setting closely resembles established black-box optimization problems such as automatic algorithm configuration ~\citep[AAC;][]{hutter-jair09a} and hyperparameter optimization (HPO) in machine learning~\citep{feurer-book19a}, where function evaluations are likewise expensive.
Moreover, these problems are frequently multi-objective, e.g., trading off runtime vs. solution quality in AAC~\citep{blot-lion16a}, or performance vs. interpretability in HPO~\citep{karl-telo23a}.
These fields have developed dedicated budget allocation strategies specifically for multi-objective optimization~\citep{blot-lion16a,rook-ec25a}.

In this work, we propose a new multi-objective prompt optimization method that builds on the evolutionary strategy of CAPO~\citep{zehle-automl25a}, extends it to explicitly handle multiple objectives, and integrates an intensification mechanism for budget allocation inspired by the multi-objective AAC method MO-SMAC~\citep{rook-ec25a}. Our main contributions are as follows:
\begin{enumerate}[nosep]
    \item We introduce \mocapointensify{}, an \textit{a posteriori} multi-objective prompt optimization algorithm that enables the discovery of full Pareto front approximations of prompts that balance performance and inference cost (see Fig.~\ref{fig:intro_pareto}) while keeping optimization itself cost-efficient.
    \item We propose a new objective function that accounts for the full computational profile of LLM inference by weighting input and output tokens, providing a realistic measure of deployment costs for modern LLMs.
    \item We are the first study to conduct a large multi-objective
    generalization evaluation by studying optimistic and pessimistic HV, approximation gap~\citep{feurer-ida23a}, and the noisy R2 metric~\citep{branke-ieeetec25a}.
    \item We compare our method against SOTA single-objective prompt optimizers and an NSGA-II-based baseline across multiple LLMs and tasks, demonstrating superiority in multi-objective metrics, efficiency, solution set diversity, and competitive top performances. 
    We verify the effect of the algorithmic components and new cost objective in dedicated ablation studies.\footnote{Code and experimental results are provided as supplementary material.}
\end{enumerate}

\section{Notation and Problem Statement}

Let $\prompt{} = (\instruction{}, \fewshots{})$ be a prompt consisting of an instruction $\instruction{}$ from the instruction space $\instructionset$ and an ordered tuple of $\nshots$ few-shot examples $\fewshots{} = (\fewshot{1}, \dots, \fewshot{\nshots})$ from the example space $\fewshotspace^\nshots$. The complete search space $\promptset = \instructionset \times\bigcup_{\nshots=0}^{\maxshots} \fewshotspace^\nshots$ includes prompts from zero to $\maxshots$ shots. In automatic prompt optimization, this search space (or subspace when only instructions are tuned) is systematically explored using automatic optimization strategies~\cite{cui-acl25a,li-arxiv25a,wan-neurips24a}.

Assume an instance $x \in \mathcal{X}$ with an associated ground truth $y \in \mathcal{Y}$ from an unknown data-generating process $P_{xy}$. Let $\dataset = {(x_i, y_i)}_{i=1}^n$ denote a dataset of size $n$ drawn from $P_{xy}$. We consider two disjoint instantiations: a development set $\devset$ for optimization and a holdout test set $\testset$ for final evaluation.

An LLM is treated as a black-box function $\llm:\promptset \times \mathcal{X} \to \mathcal{Y}$ that maps a prompt $\prompt{} \in \promptset$ and an input instance $x \in \mathcal{X}$ to a predicted output $\hat{y} \in \mathcal{Y}$.\footnote{In the following, we differentiate between (1) the evaluation-LLM $\evalllm(\prompt{},x)$, which returns the predictions $\hat y$ for an instance $x$ using $\prompt{}$, and (2) the meta-LLM $\metallm(\prompt{\text{meta}}, \instruction{})$, which is used to alter instructions $\instruction{}$ and $\hat y$ is again an instruction.}
Performance is measured by a point-wise scoring function $\scorefunction: \mathcal{Y} \cup (\mathcal{Y} \times \mathcal{Y}) \rightarrow \mathbb{R}$ that evaluates each output independently and returns a numerical score.\footnote{This includes both supervised scenarios with available ground-truth labels, where we can test, e.g., for exact match $\scorefunction(y, \hat{y}) = \mathbb{I}_{[y = \hat{y}]}$, or unsupervised scenarios without ground-truth, where the scoring function can be an LLM-as-a-Judge or a custom numerical reward function that maps the LLM-output to a score to be maximized.}
In (single-objective) prompt optimization, the goal is to find an optimal solution $\prompt{}$ that maximizes this score in expectation over the true unknown data distribution $P_{xy}$: $\argmax_{p \in \mathcal{P}} \mathbb{E}_{(x,y) \sim P_{xy}} [\scorefunction(y, \evalllm(p, x))]$.
In practice, we have to resort to maximizing an empirical score by estimating it on the finite development set $\devset$: 
\begin{equation}\label{eq:f1}
\tilde f_1(p; \devset) = \frac{1}{n_\text{dev}} \sum_{i=1}^{n_\text{dev}} \scorefunction(y_i, \evalllm(p, x_i)).
\end{equation}

However, when optimizing prompts, there can be more than one objective of interest. Often, the cost (computational or financial) of a prompt is another important consideration \cite{zehle-automl25a,camara-stil25a,yang-emnlp23a,lopes-arxiv25a}. Then the aim is to reduce the expected costs, i.e., $\argmin_{p \in \mathcal{P}} \mathbb{E}_{x \sim P_x} [\costfunction_\llm(p; x)]$ for an LLM-specific cost function $c_\llm:\promptset \times \mathcal{X} \to \mathbb R$ as a direct measure (e.g., API costs) or proxy (e.g., token count) for the cost of a prompt. Estimating this empirically on $\devset$ yields:
\begin{equation}\label{eq:f2}
f_2(p; \devset) = \frac{1}{n_\text{dev}} \sum_{i=1}^{n_\text{dev}} \costfunction_\llm(p; x_i).
\end{equation}

Further objectives $f_j$, such as additional metrics, can also be of interest. The problem now becomes a multi-objective optimization (MOO) problem. To maintain consistency with MOO conventions, we write $f_1(p; \devset) = -\tilde{f}_1(p; \devset)$.
Let $f: \promptset \rightarrow \mathbb{R}^m$ be a vector-valued objective function. The $m$-dimensional MOO problem (minimization WLOG) is then defined as~\citep{ehrgott-book05a,emmerich-naturecomp18a,karl-telo23a}:
\begin{equation}
\argmin_{\prompt{} \in \promptset} f(\prompt{}; \devset) = \argmin_{\prompt{} \in \promptset}(f_1(\prompt{}; \devset), f_2(\prompt{}; \devset), \ldots, f_m(\prompt{}; \devset)). 
\end{equation}
Objectives $f_j$ can conflict; no single solution generally minimizes all simultaneously. To compare solutions, the concept of Pareto dominance is used. Let $\prompt{1}, \prompt{2} \in \promptset$ be two candidates and $f(\prompt{1})$ and $f(\prompt{2})$ their respective objective vectors (omitting $\devset$ for brevity). $\prompt{1}$ Pareto dominates $\prompt{2}$ ($\prompt{1} \prec \prompt{2}$), iff:
\begin{equation}
\forall i \in \{1,\ldots,m\}: f_i(\prompt{1}) \leq f_i(\prompt{2}) \land \exists j \in \{1,\ldots,m\}: f_j(\prompt{1}) < f_j(\prompt{2}).
\end{equation}
A candidate $\prompt{}^*$ is \textit{Pareto-optimal} if $\not\exists \prompt{}' \in \promptset$ such that $\prompt{}' \prec \prompt{}^*$. The \textit{Pareto set} $\paretoset \subseteq \promptset$ contains all Pareto-optimal prompts. Its image $\paretofront=f(\paretoset)$ is the \textit{Pareto front}. In MOO, an optimization algorithm returns an approximation $\paretosetapprox$. Its quality is characterized by convergence of $\paretofrontapprox=f(\paretosetapprox)$ to $\paretofront$ and diversity across trade-offs~\citep{karl-telo23a}.

In this work, we focus on finding good approximations $\paretosetapprox$ for the $2$-dimensional prompt optimization problem $\argmin_{\prompt{} \in \promptset} f(\prompt{}; \devset) = \argmin_{\prompt{} \in \promptset} (f_1(\prompt{}; \devset), f_2(\prompt{}; \devset))$ with the objectives defined in (\ref{eq:f1}) and (\ref{eq:f2}), and the implicit goal to generalize to the unseen test set $\testset$.

\section{Background \& Related Work}
\label{sec:related-work}

\subsection{Multi-objective optimization}

Established literature on multi-objective optimization (MOO) exists across several areas, including AAC~\citep{blot-lion16a,rook-ec25a} and HPO~\citep{karl-telo23a}.
A large body of work in MOO relies on Multi-Objective Evolutionary Algorithms (MOEAs), which maintain a population of candidate solutions and evolve them using selection, variation, and survival operators~\citep{karl-telo23a}. NSGA-II is a widely adopted MOEA that ranks solutions via non-dominated sorting (NDS) and employs crowding distance (CD) to preserve diversity along the Pareto front~\citep{deb-ieeeevocomp02a}. We explain both NDS and CD in more detail in Appendix~\ref{app:add}. Alternative paradigms include archive-based approaches such as SPEA2~\citep{zitzler-tik01a} and indicator-based methods such as SMS-EMOA~\citep{beume-or07a}, using hypervolume (HV) contribution for selection.

Within multi-objective AAC, several methods extend established single-objective algorithms. MO-ParamILS \citep{blot-lion16a} adapts ParamILS~\citep{hutter-jair09a} to the multi-objective setting by maintaining an archive of non-dominated configurations and employing iterated local search with adaptive evaluation strategies. MO-SMAC~\citep{rook-ec25a} extends SMAC~\citep{hutter-lion11a} by introducing a multi-objective acquisition strategy and intensification mechanism that allocates evaluation effort adaptively and terminates dominated configurations early. Its MO-Intensify routine compares challengers against their nearest incumbent in objective space, terminates evaluation early when dominated, and prunes the incumbent set using CD to maintain Pareto front coverage.\footnote{In this context, an incumbent refers to a member of the incumbent set, which contains the currently non-dominated solutions maintained by the algorithm as its evolving Pareto front approximation.}

Related ideas combining evolutionary search with budget allocation via multi-fidelity approaches include TPOT-SH~\citep{parmentier-ictai19a} for AutoML pipeline construction and MO-DEHB~\citep{awad-arxiv23a}, an extension of DEHB~\citep{awad-ijcai21a}, for multi-objective hyperparameter optimization. Both integrate evolutionary operators with a resource-allocation mechanism such as successive halving~\citep{jamieson-aistats16a} or hyperband~\citep{li-jmlr18a}.
Racing-based methods~\citep{birattari-gecco02a,lopez-ibanez-orp16a} such as S-Race~\citep{zhang-gecco13a} and its extensions~\citep{miranda-esann15a,zhang-gecco15a} further explore statistically principled elimination strategies for multi-objective settings.

\subsection{Metrics in Multi-Objective Optimization} \label{sec:mo-metrics}

Assessing optimizer performance in multi-objective settings requires careful consideration of both solution quality and stability across runs. The hypervolume indicator $\mathrm{HV}(\paretosetapprox)$, which measures the volume of the region in objective space dominated by $\paretosetapprox$ w.r.t. a reference point $r$, is typically used to quantify solution quality in terms of convergence to and diversity along the Pareto front in a single scalar value~\citep{zitzler-ieeeevocomp03a} (for details, see \S\ref{app:hv}), but can obscure stability aspects like generalization errors, selection effects, and stochastic variability.

\paragraph{Approximation Gap for Generalization Robustness} In MOO for machine learning, solutions selected on the development set $\devset$ may not generalize to test data $\testset$ due to estimation error on finite samples. Solutions that were part of $\paretosetapprox$ on $\devset$ can be dominated on $\testset$. Using only the non-dominated solutions to assess solution quality can lead to overestimation. In particular, HV masks potential degradations when dominated solutions are included in the volume.

To quantify the robustness of $\paretosetapprox$, \citet{feurer-ida23a} propose the approximation gap. Two sets are constructed: an optimistic $\paretosetapprox^\text{opt}$, containing solutions from $\paretosetapprox$ that remain non-dominated on $\testset$, and a pessimistic $\paretosetapprox^\text{pes}$ with solutions that do not dominate any other in $\paretosetapprox$. This enables (1) method ranking when the optimistic HV of method $B$ falls below the pessimistic of $A$, (2) Pareto dominance when the pessimistic front of $A$ dominates the optimistic of $B$, and (3) the approximation gap $\Delta_{\text{gap}} = \mathrm{HV}(\paretosetapprox^\text{opt}) - \mathrm{HV}(\paretosetapprox^\text{pes})$ with smaller gaps indicating more robust approximations.

\paragraph{Noisy R2 Metric for Measuring Selection Robustness}
Single-objective evaluation protocols select the best solutions based on performance on $\devset$ and report their performance on $\testset$, mirroring how practitioners select solutions based on optimization performance and experience generalization performance. While straightforward for a single objective, MOO complicates this protocol: users select a single solution from $\paretosetapprox$ based on their preferences over objective trade-offs. However, standard metrics like HV on $\testset$ evaluate the (aggregated) quality of the entire solution set, potentially overstating practical performance when the selected solution generalizes poorly.

To incorporate this selection step, \citet{branke-ieeetec25a} proposes the noisy R2 (nR2) metric. For each preference vector $\boldsymbol{\lambda}_i$ parameterizing a utility function $U$,\footnote{The utility function $U$ maps multi-objective outcomes to a scalar value and is parameterized by a preference vector $\boldsymbol{\lambda}_i$, which specifies the trade-off preference between objectives. Potential utility functions are, e.g., a weighted linear combination of the objectives or a Chebychev utility with weights $\boldsymbol{\lambda}_i$.} the solution minimizing utility during optimization $\prompt{i}^* = \argmin_{\prompt{} \in \paretosetapprox} U(f(\prompt{}; \devset), \boldsymbol{\lambda}_i)$ is selected, and its utility $U(f(\prompt{i}^*; \testset), \boldsymbol{\lambda}_i)$ is evaluated on test data. Averaging these values over many sampled preference vectors yields nR2.

\paragraph{Empirical Attainment Surfaces for Stochastic Variability}
Single Pareto fronts obscure performance stability across runs, while plotting all fronts can be visually cluttered. Empirical Attainment Surfaces (EAS)~\cite{watanabe-arxiv23a} aggregate $S$ runs into interpretable front shapes with performance bands. 
The empirical attainment function $\alpha(\mathbf{y}) := \frac{1}{S}\sum_{s=1}^{S}\mathbb{I}[\paretofrontapprox^{s} \preceq \mathbf{y}]$ measures the fraction of runs whose front approximation $\paretofrontapprox^{s}$ weakly dominates $\mathbf{y}$ (i.e., at least as good as $\mathbf{y}$ in all objectives). The $L/S$-EAS contains points with $\alpha(\mathbf{y}) \ge L/S$, meaning at least $L$ out of $S$ runs attained this surface.
Plotting several EAS, such as median, best ($1/S$), and worst ($S/S$), compactly visualizes performance and variability: narrow bands signal stable behavior across runs, wide bands reveal variability.
However, like HV, EAS are computed on the optimistic solution set and therefore mask the generalization gap between $\devset$ and $\testset$. A narrow EAS band may suggest stable, high-quality performance, but can conceal that some of the attained solutions are dominated on unseen data.

\subsection{Related Work on Prompt Optimization}

Automatic prompt optimization is typically categorized into continuous methods, learning soft prompts in embedding space, and discrete methods, directly modifying textual prompts~\cite{cui-acl25a}. Due to prompt interpretability and their LLM-agnostic nature, we focus on discrete methods.

Many discrete methods rely on a \textit{meta-LLM} guided by a \textit{meta-prompt} to modify prompts. This includes Monte Carlo search like APE~\cite{zhou-iclr23a}, evolutionary approaches like EvoPrompt~\cite{guo-iclr24a} and PromptBreeder~\cite{fernando-icml24a}, using the meta-LLM for evolutionary operators~\citep{meyerson-telo24a}, or OPRO~\cite{yang-iclr24a}, which employs the meta-LLM directly as optimizer with performance histories in the meta-prompt.
A complementary line of work applies gradient-based optimization ideas to textual prompts, including ProTeGi~\cite{pryzant-emnlp23a}, TextGrad~\cite{yuksekgonul-nature25a}, and LLM-AutoDiff~\cite{yin-arxiv25a}.

While these works optimize only instructions, i.e., over $\promptset = \instructionset$, recent approaches jointly optimize prompts and few-shot examples~\citep{agarwal-arxiv24a,zehle-automl25a}. This has been shown to yield strong synergies~\citep{wan-neurips24a} and is used by leading prompt optimizers such as MIPROv2~\citep{opsahl-ong-emnlp24a}.
GEPA~\citep{agrawal-arxiv25a} instead uses reflective prompt evolution, iteratively diagnosing failures from system trajectories and refining high-level prompt rules, while outperforming MIPROv2. Both are included in DSPy~\citep{khattab-iclr24a}.

In parallel, efficiency considerations have become central. 
EASE~\citep{wu-neurips24a} applies neural bandits to efficiently optimize few-shot example selection and can be extended to jointly optimize examples and instructions.
CAPO~\citep{zehle-automl25a} is an evolutionary approach that combines instruction and few-shot optimization, integrates racing to discard underperforming prompts early, and penalizes prompt length to control inference cost. A recent comparison~\citep{zehle-arxiv25a} shows that CAPO can outperform SOTA optimizers like GEPA.

These optimizers are all single-objective optimizers targeting prompt performance. Although GEPA~\cite{agrawal-arxiv25a} introduces a multi-objective-inspired Pareto-based selection mechanism for population diversity, the optimization is single-objective.
CAPO~\citep{zehle-automl25a} as well as Promptomatix~\citep{murthy-arxiv25a} consider prompt length as an additional objective, but use scalarization with a fixed trade-off parameter that must be chosen \textit{a priori}, which can be challenging.

Nonetheless, there exist multi-objective methods that yield Pareto fronts of prompts to trade off performance and length. InstOptima~\cite{yang-emnlp23a} uses NSGA-II with meta-prompts to optimize prompt performance vs. length vs. perplexity. It includes separate mutation and crossover for instructions and examples, and the meta-LLM is guided by the parents' performances. 
MOPrompt~\citep{camara-stil25a} also optimizes prompts for accuracy and token length with NSGA-II. This is also done for joint model selection and prompt construction from predefined blocks~\cite{lopes-arxiv25a}, but without proper evaluation on unseen test data.

Other works consider alternative objective trade-offs. Survival of the Safest~\cite{sinha-emnlpit24a} proposes a multi-objective, feedback-based mutation strategy to trade off performance and security, with the final solution chosen via a weighted combination of the objectives.
EMO-Prompt~\cite{baumann-aec24a} applies NSGA-II and SMS-EMOA with meta-prompts for crossover and mutation, evaluated on story generation with conflicting sentiment objectives.
MOPO~\cite{menchaca-resendiz-iccl25a} similarly employs NSGA-II for affective text generation.
FairPrompt~\cite{shafqat-ieeeaccess25a} uses NSGA-II to select prompts that jointly maximize semantic diversity, bias detection rate, and balanced demographic coverage for fairness testing.

Using reinforcement learning strategies to tune small networks on top of an LLM whose weights are frozen, MORL-Prompt~\cite{jafari-femnlp24a} compares three different scalarization and indicator methods, while Pareto Prompt Optimization~\cite{zhao-iclr25a} proposes a multi-objective weight update for the neural network. Both works focus on optimizing performance and style metrics.

However, these works do not address the cost-efficiency of the optimization process, despite the substantial cost of LLM evaluations, and they lack rigorous multi-objective evaluation protocols.

\paragraph{Positioning of \mocapointensify}
While the idea of trading off prompt quality and prompt length through MOO is not new, we are the first to explicitly examine how to efficiently arrive at solutions by incorporating a multi-objective budget allocation mechanism from MO-AAC into evolutionary prompt optimization. Moreover, we go beyond using prompt length as a proxy for cost and explicitly consider input and output token usage, covering all costs involved in LLM inference.
In addition to algorithmic contributions, we provide a systematic evaluation of Pareto front approximations and how selected solutions generalize to unseen test data, whereas prior work either picks a single solution by scalarizing the objectives or omits a dedicated test-set analysis altogether. To the best of our knowledge, this constitutes the first thorough evaluation of multi-objective optimization for machine learning, extending the preliminary work of \citet{feurer-ida23a} and applying the methodology from \citet{branke-ieeetec25a}.

\section{Multi-Objective Cost-Aware Prompt Optimization}
\label{sec:mocapo}
We introduce \mocapointensify{}, a multi-objective prompt optimization algorithm that jointly optimizes performance and inference cost.
Our approach extends CAPO by moving from searching for the optimal solution at a fixed trade-off to performing true multi-objective optimization. 
For this purpose, we propose a deployment-oriented cost objective that explicitly accounts for both input and output token consumption, reflecting the full computational profile of modern LLM inference.
In addition, it integrates an intensification mechanism for budget allocation in an evolutionary algorithm to ensure cost-efficient search. Together, these components enable the discovery of robust Pareto front approximations while maintaining high optimization efficiency.

\subsection{Cost Objective}
\label{sec:cost-objective}

We propose $c_\llm(\prompt{},x)=w_\text{in}tok_\text{in}(p; x) + w_\text{out}tok_\text{out}(p; x)$ to measure the cost of a prompt candidate $\prompt{}$ on an instance $x$. Here, $tok_\text{in}(p; x)$ and $tok_\text{out}(p; x)$ represent the number of input and output tokens processed by the LLM during inference with $\prompt{}$ on $x$. $ w_\text{in}$ and $w_\text{out}$ are (potentially LLM-specific) weights. The corresponding cost-objective on a dataset $\dataset$ is then:
\begin{equation}
f_2(\prompt{}; \dataset) = \frac{1}{n}\sum_{i=1}^n \left [ w_\text{in}tok_\text{in}(\prompt{}; x_i) + w_\text{out}tok_\text{out}(\prompt{}; x_i) \right ]=  w_\text{in}\widebar{tok_\text{in}}(\prompt{}; \dataset) + w_\text{out}\widebar{tok_\text{out}}(\prompt{}; \dataset).
\end{equation}

\begin{wraptable}{r}{0.5\textwidth}
    \centering
    \caption{Cost weights for various LLMs derived from \textit{OpenRouter}. We average prices per million tokens across all listed providers per model (cf. \S\ref{app:pricing}).}
    \vspace{-0.5em}
    \begin{tabular}{l r@{.}l r@{.}l}
    \toprule
    \textbf{Model} 
        & \multicolumn{2}{c}{$\weightin$} 
        & \multicolumn{2}{c}{$\weightout$} \\ 
    \midrule
    \mistral        & 0  & 08  & 0   & 32  \\
    \qwen           & 0  & 11  & 0   & 41  \\ 
    \gpt            & 0  & 12  & 0   & 49  \\
    Claude Opus 4.6 & 2  & 72  & 25  & 43  \\
    GPT-5.2 Pro     & 21 & 00  & 168 & 00  \\
    \bottomrule
    \end{tabular}
    \label{tab:cost-weights}
\end{wraptable}

This formulation allows accounting for the distinct computational profiles of input and output tokens. In current autoregressive LLMs input tokens are usually cheaper, as processed in parallel during pre-fill, whereas output tokens are more expensive, as generation proceeds sequentially through autoregressive decoding. Token weights are defined based on the inference costs prior to optimization. For local deployments, weights can be determined from latency metrics (e.g., time-to-first-token, tokens-per-second) combined with resource costs. For deployments via vendor APIs, weights can be obtained from the pricing structure. To illustrate this, we derive model-specific weights $\weightin$ and $\weightout$ for exemplary LLMs from the pricing data of all vendors listed on \textit{OpenRouter}\footnote{\textit{OpenRouter} (\url{https://openrouter.ai}) is a unified API platform providing access to over 400 AI models through a single endpoint, passing through provider-native pricing with distinct input and output rates.} in Tab.~\ref{tab:cost-weights}. This serves as a proxy for inference costs and reflects the distinct computational profiles described above. We also employ these weights later in our experiments in \S\ref{sec:experiments}.

\subsection{A Multi-Objective Prompt Optimization Baseline Using NSGA-II}\label{sec:nsga-ii-po}

To represent the existing multi-objective prompt optimization methods that use NSGA-II~\cite{deb-ieeeevocomp02a} in combination with a meta-LLM to optimize prompts for a variety of objectives (as outlined in \S\ref{sec:related-work}), we introduce a baseline algorithm for our work, \mocaponsgaII{}. It uses NSGA-II with the evolutionary operators from CAPO~\citep{zehle-automl25a} to optimize prompts consisting of instructions and few-shot examples for both performance and our cost objective from \S\ref{sec:cost-objective}. 

In particular, it employs a binary tournament with front rank and CD as selection criteria for parent selection, contrasting with CAPO's random selection.
Environmental selection is performed via non-dominated sorting (NDS) with crowding distance (CD) as a tie-breaker to include partial fronts. We use a steady-state version of NSGA-II, recalculating CD after each selection, since removing candidates alters neighbor distances. Pseudo-code for this baseline is provided in \S\ref{app:nsgaii}.

\subsection{MO-CAPO}

We now present our new method \mocapointensify{}. It advances our baseline from \S\ref{sec:nsga-ii-po} by introducing a budget allocation mechanism to terminate the evaluation of unpromising prompts early, given the high cost of evaluations in prompt optimization. The method integrates and adapts core concepts from NSGA-II~\cite{deb-ieeeevocomp02a}, MO-SMAC~\citep{rook-ec25a}, and CAPO~\citep{zehle-automl25a} to develop a cost-efficient algorithm for multi-objective prompt optimization.

\paragraph{Overview} 
\mocapointensify{} follows a standard genetic algorithm (see Algorithm~\ref{algo:mo-capo-smac:main}). After prompt initialization and their initial evaluation, the algorithm creates an incumbent set $\incset\subseteq\population$ of non-dominated solutions. In each iteration of the evolutionary loop, the genetic operators \HyperCall{crossover}{}{func:cross_over} and \HyperCall{mutate}{}{func:mutate} generate $\ncrossovers$ offspring $\offspringpromptset$. Parents are selected based on binary tournament selection. Each offspring enters an adapted version of MO-SMAC's intensification mechanism as a ``challenger'', which is progressively evaluated on additional data and compared against the incumbent set. If a prompt survives intensification, it is added to $\incset$. After $T$ iterations, $\incset$ is returned as the final Pareto front approximation.
\mocapointensify{} differs from typical evolutionary algorithms and MO-SMAC by retaining both a population $\population$ and a dedicated incumbent set $\incset\subseteq\population$.\footnote{The incumbent set differs conceptually from the first front identified by non-dominated sorting, as it can contain candidates that are evaluated for different budgets.} The incumbent set serves as the current Pareto front approximation and reference for intensification. However, in contrast to MO-SMAC, rejected candidates cannot simply be all discarded after early termination to retain sufficient genetic material for crossover and mutation.\footnote{Discarding all dominated solutions would deplete genetic diversity and limit exploration, while using the complete prompt history $\runhistory$ would introduce all unpromising candidates. Our separation into an evolving population and incumbent set balances diversity preservation with efficient dominance-based pruning.}
In the following, we describe each component in more detail. Full algorithmic specifications are provided in \S\ref{app:algo-details}.
  
\begin{algorithm}[tb]
\small
\caption{\mocapointensify{}: Multi-Objective Cost-Aware Prompt Optimization}
\begin{algorithmic}[1]
\Require {datasets $\devset$ \& $\fewshotset$, 
    meta-LLM $\metallm$, 
    evaluation-LLM $\evalllm$, 
    initial instructions $\initialinstructions = \{\instruction{1}, \dots, \instruction{\populationsize}\}$,
    population size $\populationsize$,
    block size $\blocksize$,
    no. iterations $\niters$,
    no. crossovers per iteration $\ncrossovers$,
    max. no. shots $\maxshots$,
    input \& output token-weights $\weightin$ \& $\weightout$,
    crossover-meta-prompt $\crossoverprompt$,
    mutation-meta-prompt $\mutationprompt$
    }
\State Divide dataset $\devset$ into blocks $\blockset = \{\block{1}, ..., \block{z}\}$ where $|\block{i}| = \blocksize$
\State $\population \gets \HyperCall{initialize\_pop}{\initialinstructions, \fewshotset, \maxshots, \evalllm}{func:initialize_population}$
\State $\block{} \gets \Call{sample}{\blockset}$
\State $\runhistory \gets \Call{evaluate}{\population, \block{}, \evalllm, \weightin, \weightout}$
\State $\incset \gets \Call{get\_non\_dominated}{\population, \runhistory,  \block{}}$
\For{$\iter=1$ to $\niters$}
    \State $\offspringpromptset \gets \HyperCall{crossover}{\population, \runhistory, \metallm, \crossoverprompt, \ncrossovers}{func:cross_over}$
    \State $\offspringpromptset \gets \HyperCall{mutate}{\offspringpromptset, \metallm, \evalllm, \mutationprompt, \fewshotset, \maxshots}{func:mutate}$
    \For{$\challenger$ in $\offspringpromptset$}
        \State $\incset, \population, \runhistory \gets \HyperCall{intensify}{\incset, \population, \challenger, \blockset, \evalllm, \runhistory, \weightin, \weightout, \populationsize}{func:intensify}$ 
    \EndFor
\EndFor
\State \Return $\incset$
\end{algorithmic}
\label{algo:mo-capo-smac:main}
\end{algorithm}

\paragraph{Initialization}
Following the original CAPO method, \mocapointensify{} partitions the development set $\devset$ into blocks $\blockset$ of equal size $\blocksize$ and creates an initial population $\population$ in \HyperCall{initialize\_pop}{}{func:initialize_population} by augmenting the initial instructions $\initialinstructions$ with up to $\maxshots$ randomly sampled few-shot examples $\fewshots{}$ (see \S\ref{app:algo-details} for details). Each prompt in $\population$ is evaluated on the same single randomly sampled block $\block{}$, yielding their objective function vectors $f(\prompt{}; \block{})$. Based on these evaluations, the incumbent set $\incset$ of all non-dominated solutions is populated using NDS. A run history $\runhistory$ records all evaluations performed for each candidate on each block to prevent redundant computations in future steps.

\paragraph{Parent Selection}
We design the parent selection to favor candidates that are close to the Pareto front, while preserving diversity and accounting for different levels of evaluation. This differs from CAPO's random parent selection. \mocapointensify{} performs a binary tournament, adapted from NSGA-II to handle heterogeneous evaluation levels. Two independent tournaments are conducted to obtain distinct parents. In each tournament, two candidates are sampled from $\population$ and compared:
(1) If exactly one candidate is an incumbent, it wins, favoring the candidate that contributes to the Pareto front approximation.
(2) If both candidates are incumbents, the higher CD decides in order to maintain diversity of the incumbent front.
(3) If both are non-incumbents and share identical evaluation levels, they are compared using NDS and CD.
(4) For non-incumbents with different evaluation levels, a weaker dominance criterion from MO-ParamILS~\cite{blot-lion16a} is applied: if $\blockset_{\prompt{i}} \subset \blockset_{\prompt{j}}$, $\prompt{j}$ wins if it dominates $\prompt{i}$ on the subset $\blockset_{\prompt{i}}$. This ensures a fair comparison, as candidates evaluated on fewer blocks could otherwise appear superior by chance alone due to the high variance in performance estimates from small samples.
(5) When the previous criteria do not provide a clear preference, random selection decides the winner (for details, see \S\ref{app:algo-details}).

\paragraph{Offspring Creation}
The genetic operators \HyperCall{crossover}{}{func:cross_over} and \HyperCall{mutate}{}{func:mutate} follow the established mechanisms from CAPO~\citep{zehle-automl25a}. Instructions of the selected parents are crossed using a meta-LLM, while the offspring's few-shot examples are sampled from the union of the parents' few-shot pool, creating $\ncrossovers$  offspring instructions in $\offspringpromptset$. These are subsequently mutated by altering the instruction with a meta-LLM and by randomly removing or adding examples (see \S\ref{app:algo-details}).

\begin{algorithm}[tb]
\small
\caption{\mocapointensify: Intensification}
\begin{algorithmic}[1]
\Function{intensify}{$\incset, \population, \challenger, \blockset, \evalllm, \runhistory, \weightin, \weightout, \populationsize$}
\hypertarget{func:intensify}{}
    \State $\blockscommon \gets \bigcap_{\prompt{} \in \incset} \Call{get\_evaluated\_blocks}{\prompt{}, \runhistory}$ \Comment{Get shared evaluation basis}
    \State $\blockschal \gets [\,]$
    \State $f_\text{new} \gets (\infty, \dots, \infty)$
    \While{\textbf{true}}
        \State $f_\text{old} \gets f_\text{new}$
        \Repeat \Comment{Progressively add block evaluations of $\challenger$}
            \State $\blockschal \gets \blockschal \cup \{\Call{sample}{\blockscommon \setminus \blockschal}\}$
            \State $\runhistory \gets \Call{evaluate}{\challenger, \blockschal, \evalllm, \weightin, \weightout}$
            \State $f_\text{new} \gets \Call{get\_objective\_values}{\challenger, \runhistory, \blockschal}$
        \Until{$f_\text{new} \prec f_\text{old}$ \textbf{or} $\blockscommon = \blockschal$}
        \If{$\blockscommon = \blockschal$} \Comment{Case (b): temporal acceptance}
            \State $\population \gets \population \cup \{\challenger\}$
            \State $\incset \gets \Call{get\_non\_dominated}{\incset \cup \{\challenger\}, \runhistory, \blockschal}$ \Comment{Re-assess incumbents}
            \State \textbf{break}
        \Else \Comment{Case (a): early rejection check} %
            \State $\incumbent \gets \Call{get\_closest\_inc}{\incset, \challenger, \blockschal}$
            \State $f_\text{inc} \gets \Call{get\_objective\_values}{\incumbent, \runhistory, \blockschal}$
            \If{$f_\text{inc} \prec f_\text{new}$} \Comment{If closest incumbent dominates, ``reject'' $\challenger$}
                \State $\population \gets \population \cup \{\challenger\}$
                \State \textbf{break}
            \EndIf
        \EndIf
    \EndWhile
    \If{$|\population| > \populationsize$} \Comment{Maintain population size}
        \State $\population, \incset \gets \HyperCall{environmental\_selection}{\population, \incset, \runhistory, \populationsize}{func:mocapo_envselection}$
    \EndIf
    \State $\runhistory \gets \HyperCall{advance\_incumbents}{\incset, \runhistory, \evalllm, \weightin, \weightout}{func:mocapo_advinc}$ \Comment{Advance incumbent}
    \State \Return $\incset, \population, \runhistory$
\EndFunction
\end{algorithmic}
\label{algo:intensification}
\end{algorithm}

\paragraph{Intensification}
Similar to CAPO's single-objective racing, we aim to avoid wasting budget by evaluating weak candidates on the full data. To still enable fair comparisons with the incumbent set, we evaluate challengers progressively on more data blocks. The multi-objective intensification mechanism in Algorithm~\ref{algo:intensification} starts by identifying the common blocks $\blockscommon$ on which all incumbents have been evaluated so far. The challenger must be evaluated on this same subset before it can be added to $\incset$. Therefore, it is evaluated progressively on more blocks that are drawn at random from $\blockscommon$. A comparison against the incumbents is triggered either if (a) the challenger's performance has improved, i.e., the new objective vector $f_\text{new}$ after adding further evaluations dominates the objective vector $f_\text{old}$ from the previous comparison,\footnote{A comparison is always triggered after the first block evaluation as we initialize $f_\text{old}=(\infty,\ldots,\infty)$.} or (b) the challenger has been evaluated on all blocks from $\blockscommon$.
In case (a), the challenger is compared to the closest incumbent\footnote{The nearest incumbent is selected based on the Euclidean distance in objective space, where objective vectors are min-max normalized over all objective values seen during the optimization process so far. \citet{rook-ec25a} argue that comparing only against the closest incumbent instead of all incumbents reduces the type I error, i.e., the error of falsely rejecting a good challenger based on the comparison on only a subset of the data.} on the challenger's blocks.
If the challenger is dominated by the closest incumbent, it is ``rejected'' in the sense that its intensification is terminated and the challenger is not added to $\incset$, but is still added to the population $\population$ as genetic material. If it is not dominated, the loop continues by adding further evaluations.
In case (b), the challenger survived the intensification loop and was evaluated on all blocks in $\blockscommon$. It is added to the population and temporarily to the set of potential incumbents. This set is re-evaluated using NDS on the commonly evaluated blocks to ensure that only non-dominated solutions remain in $\incset$.

Finally, we perform environmental selection (see next paragraph) to maintain population size $\populationsize$. Moreover, we advance the evaluation of the incumbent set. The incumbent with the fewest evaluations is selected. If it lags behind in evaluations, it is evaluated on one of the missing blocks. If all incumbents share the same block set, it is evaluated on a new block sampled from $\blockset$. This mechanism ensures comparable evaluation levels for the incumbents while gradually expanding the evaluation basis as the optimization progresses.

\paragraph{Environmental Selection} Environmental selection ensures that the population size does not exceed $\populationsize$ by eliminating poorly performing candidates from the population.
Candidates not in $\incset$ are removed preferentially, as they are dominated.
If such candidates exist and have all been evaluated on the same blocks, NDS and CD are applied, eliminating candidates from the worst front with CD as tie-breaker. If such candidates do not share the same evaluation blocks, such a comparison would not be reliable. In this case, candidates with the fewest evaluations are discarded first, as their objective estimates are based on less evidence.
If this applies to multiple candidates, a random elimination is performed among them.
When only candidates in $\incset$ remain, CD over $\blockscommon$ is used as an elimination criterion to promote diversity among the front (see \S\ref{app:algo-details}).

\section{Experimental Design}\label{sec:experiments}

\paragraph{Setup}
We evaluate \mocapointensify{} using three open-weight LLMs of different sizes and organizational origins: \gpt{}~\citep{gptoss-report25}, \qwen{}~\cite{qwen-technical25}, and \mistral{}~\cite{mistral31-blog25}. These model sizes enable deployment on single-GPU instances while delivering strong performance. Each experiment employs the same model for prompt alteration ($\metallm$) and evaluation ($\evalllm$). We provide the complete model specifications in \S\ref{app:model-details}.

We employ four datasets spanning a diverse range of tasks, which vary in domain, output cardinality, and reasoning complexity: (1) \textit{AG News} (4-class topic classification~\citep{zhang-neurips15a}), (2) \textit{GSM8K} (grade school math word problems~\citep{cobbe-arxiv21a}), (3) \textit{Subj} (binary subjectivity classification~\citep{pang-acl04a}), and (4) \textit{MBPP} (programming problems~\citep{austin-arxiv21}).
Each dataset is partitioned into three subsets: a development set $\devset$ of 300 instances for optimization, a few-shot pool $\fewshotset$ of 100 instances, and a hold-out test set $\testset$ of 500 instances for generalization assessment. Detailed descriptions of datasets and performance measures are provided in \S\ref{app:dataset-details}.

We benchmark the proposed \mocapointensify{} against the \mocaponsgaII{} baseline, as a representative of existing multi-objective prompt optimizers, and against unoptimized initial instructions. In addition, we compare against three state-of-the-art single-objective prompt optimizers: CAPO~\cite{zehle-automl25a} (primary baseline as our method directly builds upon it), EvoPromptGA~\cite{guo-iclr24a} (widely known evolutionary optimizer), and GEPA~\cite{agrawal-arxiv25a} (popular due to implementation in DSPy).
The evolutionary optimizers require ten initial instructions as starting point, and GEPA a single initial instruction. In addition, CAPO, \mocapointensify{}, and \mocaponsgaII{} use a textual task description in the meta-prompts.
For this, task descriptions and a pool of 15 initial instructions per dataset are taken from \citet{zehle-automl25a} for datasets (1)-(3). For the new task (4), we manually create a task description (cf. \S\ref{app:task-descriptions}) and automatically generate an initial instruction pool using Anthropic's Claude (cf. \S\ref{app:init-instructions}). In each optimization run, initial instructions for the optimizers are sampled from this pool. The same ten sampled initial instructions are used as a baseline for the comparison. Despite local LLM deployment, we use the averaged API prices from Tab.~\ref{tab:cost-weights} as a proxy for the model-specific token weights $\weightin$ and $\weightout$. This choice allows interpreting the cost objective value as the average potential API cost in US dollars per 1M calls with this prompt. We provide an overview of the parameterizations of all optimizers in \S\ref{app:implementation-details}.

Each experiment configuration is executed with three random seeds controlling initial instruction sampling, dataset partitioning, LLM decoding, and stochastic algorithm components. All optimization runs terminate under a budget constraint of $7.5 \times 10^6$ total tokens consumed by $\evalllm$.

\paragraph{Evaluation}
From the multi-objective optimizers and the initial instruction set, we can naturally form Pareto fronts, which we evaluate using the multi-objective nR2 and HV metrics.\footnote{To ensure comparability of objectives, optimizers, and seeds, we employ global min-max normalization per dataset. The normalization bounds are derived from the union of all incumbent solutions across development and test set scores of all multi-objective optimizers and seeds.} To obtain the nR2 metric, we use the Chebychev utility (following ~\citet{branke-ieeetec25a}) and average across 500 uniformly sampled preference vectors. Moreover, we use EAS plots from the package by \citet{watanabe-arxiv23a}. With $S=3$ runs per task, we plot the $2/3$-EAS (median) as a line with bands spanning $1/3$-EAS (best) and $3/3$-EAS (worst).
Single-objective optimizers naturally only yield a single incumbent in their last step. Therefore, we evaluate single-objective methods solely in terms of their accuracy and compare them with the maximum-accuracy solutions of the multi-objective methods.
Unless stated otherwise, reported metrics represent the mean and standard deviation over three seeds, where candidates are selected by their performance on $\devset$, and their objective vectors are subsequently evaluated on $\testset$.

\section{Results \& Analysis}

\subsection{Benchmark Results}

\begin{table}[b]
\centering
\caption{Results for multi-objective prompt optimizers compared to initial instructions across all datasets and models at 7.5M token budget.
nR2 measures convergence (lower is better). HV$_{opt}$ (optimistic) and HV$_{pes}$ (pessimistic) measure Pareto front quality (higher is better).
Gap measures HV uncertainty (lower is better).
Bold indicates best value per model-dataset combination.}
\label{tab:results_main}
\small
\setlength{\tabcolsep}{2pt}
\renewcommand{\arraystretch}{0.9}
\begin{adjustbox}{max width=\textwidth}
\begin{tabular}{ll cccc cccc}
\toprule
& & \multicolumn{4}{c}{\textbf{GSM8K}} & \multicolumn{4}{c}{\textbf{Subj}} \\
\cmidrule(lr){3-6} \cmidrule(lr){7-10}
\textbf{Model} & \textbf{Optimizer} & nR2 $\downarrow$ & HV$_{opt}$ $\uparrow$ & HV$_{pes}$ $\uparrow$ & Gap $\downarrow$ & nR2 $\downarrow$ & HV$_{opt}$ $\uparrow$ & HV$_{pes}$ $\uparrow$ & Gap $\downarrow$ \\
\midrule
\multirow{3}{*}{\gpt{}} & Initial & $0.319_{\pm.078}$ & $0.438_{\pm.121}$ & $0.426_{\pm.115}$ & $0.012_{\pm.021}$ & $0.214_{\pm.023}$ & $0.719_{\pm.047}$ & $0.708_{\pm.043}$ & $\mathbf{0.012}_{\pm.020}$ \\
 & \mocaponsgaII & $0.098_{\pm.045}$ & $0.973_{\pm.116}$ & $0.920_{\pm.088}$ & $0.053_{\pm.092}$ & $0.098_{\pm.008}$ & $0.987_{\pm.023}$ & $\mathbf{0.966}_{\pm.022}$ & $0.021_{\pm.004}$ \\
 & \mocapointensify & $\mathbf{0.038}_{\pm.005}$ & $\mathbf{1.150}_{\pm.020}$ & $\mathbf{1.145}_{\pm.019}$ & $\mathbf{0.005}_{\pm.007}$ & $\mathbf{0.095}_{\pm.009}$ & $\mathbf{0.998}_{\pm.009}$ & $0.949_{\pm.024}$ & $0.049_{\pm.033}$ \\
\cmidrule{1-10}
\multirow{3}{*}{\mistral{}} & Initial & $0.387_{\pm.013}$ & $0.352_{\pm.029}$ & $0.352_{\pm.029}$ & $\mathbf{0.000}_{\pm.000}$ & $0.383_{\pm.071}$ & $0.361_{\pm.153}$ & $0.361_{\pm.153}$ & $\mathbf{0.000}_{\pm.001}$ \\
 & \mocaponsgaII & $0.122_{\pm.038}$ & $0.920_{\pm.086}$ & $0.896_{\pm.096}$ & $0.024_{\pm.023}$ & $0.130_{\pm.024}$ & $0.880_{\pm.077}$ & $0.842_{\pm.085}$ & $0.038_{\pm.054}$ \\
 & \mocapointensify & $\mathbf{0.110}_{\pm.034}$ & $\mathbf{0.943}_{\pm.080}$ & $\mathbf{0.930}_{\pm.073}$ & $0.013_{\pm.015}$ & $\mathbf{0.129}_{\pm.026}$ & $\mathbf{0.898}_{\pm.093}$ & $\mathbf{0.884}_{\pm.109}$ & $0.014_{\pm.016}$ \\
\cmidrule{1-10}
\multirow{3}{*}{\qwen{}} & Initial & $0.374_{\pm.011}$ & $0.361_{\pm.018}$ & $0.361_{\pm.018}$ & $\mathbf{0.000}_{\pm.000}$ & $0.264_{\pm.061}$ & $0.590_{\pm.126}$ & $0.590_{\pm.126}$ & $\mathbf{0.000}_{\pm.000}$ \\
 & \mocaponsgaII & $0.343_{\pm.006}$ & $0.399_{\pm.009}$ & $0.399_{\pm.009}$ & $\mathbf{0.000}_{\pm.000}$ & $0.111_{\pm.023}$ & $0.927_{\pm.061}$ & $0.913_{\pm.079}$ & $0.014_{\pm.020}$ \\
 & \mocapointensify & $\mathbf{0.340}_{\pm.006}$ & $\mathbf{0.410}_{\pm.018}$ & $\mathbf{0.406}_{\pm.021}$ & $0.004_{\pm.004}$ & $\mathbf{0.106}_{\pm.005}$ & $\mathbf{0.930}_{\pm.015}$ & $\mathbf{0.925}_{\pm.022}$ & $0.005_{\pm.006}$ \\
\end{tabular}
\end{adjustbox}

\begin{adjustbox}{max width=\textwidth}
\begin{tabular}{ll cccc cccc}
\toprule
& & \multicolumn{4}{c}{\textbf{AG News}} & \multicolumn{4}{c}{\textbf{MBPP}} \\
\cmidrule(lr){3-6} \cmidrule(lr){7-10}
\textbf{Model} & \textbf{Optimizer} & nR2 $\downarrow$ & HV$_{opt}$ $\uparrow$ & HV$_{pes}$ $\uparrow$ & Gap $\downarrow$ & nR2 $\downarrow$ & HV$_{opt}$ $\uparrow$ & HV$_{pes}$ $\uparrow$ & Gap $\downarrow$ \\
\midrule
\multirow{3}{*}{\gpt{}} & Initial & $0.052_{\pm.008}$ & $1.086_{\pm.029}$ & $1.010_{\pm.096}$ & $0.076_{\pm.102}$ & $0.429_{\pm.035}$ & $0.278_{\pm.032}$ & $0.259_{\pm.063}$ & $0.019_{\pm.032}$ \\
 & \mocaponsgaII & $0.068_{\pm.025}$ & $1.044_{\pm.046}$ & $\mathbf{1.031}_{\pm.044}$ & $\mathbf{0.014}_{\pm.022}$ & $0.306_{\pm.011}$ & $0.504_{\pm.077}$ & $0.469_{\pm.023}$ & $0.035_{\pm.054}$ \\
 & \mocapointensify & $\mathbf{0.045}_{\pm.012}$ & $\mathbf{1.116}_{\pm.045}$ & $0.939_{\pm.276}$ & $0.177_{\pm.238}$ & $\mathbf{0.144}_{\pm.120}$ & $\mathbf{0.840}_{\pm.293}$ & $\mathbf{0.834}_{\pm.295}$ & $\mathbf{0.005}_{\pm.005}$ \\
\cmidrule{1-10}
\multirow{3}{*}{\mistral{}} & Initial & $\mathbf{0.177}_{\pm.047}$ & $\mathbf{0.806}_{\pm.105}$ & $\mathbf{0.806}_{\pm.105}$ & $\mathbf{0.000}_{\pm.000}$ & $0.196_{\pm.068}$ & $0.731_{\pm.157}$ & $0.671_{\pm.178}$ & $0.060_{\pm.105}$ \\
 & \mocaponsgaII & $0.224_{\pm.066}$ & $0.713_{\pm.137}$ & $0.649_{\pm.168}$ & $0.064_{\pm.063}$ & $\mathbf{0.092}_{\pm.010}$ & $\mathbf{0.995}_{\pm.038}$ & $\mathbf{0.936}_{\pm.060}$ & $\mathbf{0.059}_{\pm.069}$ \\
 & \mocapointensify & $0.231_{\pm.031}$ & $0.704_{\pm.099}$ & $0.457_{\pm.360}$ & $0.247_{\pm.393}$ & $0.107_{\pm.044}$ & $0.979_{\pm.131}$ & $0.859_{\pm.211}$ & $0.120_{\pm.083}$ \\
\cmidrule{1-10}
\multirow{3}{*}{\qwen{}} & Initial & $0.269_{\pm.020}$ & $0.628_{\pm.040}$ & $0.613_{\pm.046}$ & $\mathbf{0.014}_{\pm.012}$ & $0.380_{\pm.039}$ & $0.392_{\pm.086}$ & $0.331_{\pm.045}$ & $0.061_{\pm.064}$ \\
 & \mocaponsgaII & $\mathbf{0.177}_{\pm.021}$ & $\mathbf{0.826}_{\pm.063}$ & $\mathbf{0.782}_{\pm.029}$ & $0.044_{\pm.042}$ & $\mathbf{0.144}_{\pm.046}$ & $\mathbf{0.860}_{\pm.128}$ & $\mathbf{0.848}_{\pm.119}$ & $\mathbf{0.012}_{\pm.010}$ \\
 & \mocapointensify & $0.190_{\pm.027}$ & $0.808_{\pm.067}$ & $0.704_{\pm.082}$ & $0.104_{\pm.134}$ & $0.165_{\pm.027}$ & $0.832_{\pm.102}$ & $0.730_{\pm.104}$ & $0.102_{\pm.100}$ \\
\bottomrule
\end{tabular}
\end{adjustbox}
\end{table}

\paragraph{Comparison on Multi-objective Metrics.}
First, we compare \mocapointensify{} with \mocaponsgaII{} and initial instructions after the full budget in Table~\ref{tab:results_main}.
According to the nR2 metric, \mocapointensify{} achieves the best performance in 8 out of 12 model–dataset combinations, while \mocaponsgaII{} leads in 3 cases. In these cases, \mocapointensify{} is still within one standard deviation.
In one setting (\mistral{} on AG News), the unoptimized initial instructions perform best.
However, the overall results on AG News show generally limited gains from optimization, suggesting that its relatively lower task complexity leads to saturation with the employed LLMs, while the more complex reasoning and generation tasks benefit substantially from prompt optimization.
On those, we consistently outperform the initial instructions.
This underlines that prompt choice alone has a substantial impact on the final performance–cost trade-off, and relying on manually designed or default instructions leaves considerable untapped potential. Automatic prompt optimization is therefore not merely beneficial, but often necessary to obtain competitive solution sets in the multi-objective setting.

\findingbox{\mocapointensify{} outperforms the NSGA-II-based baseline in 8 out of 12 settings.}
\vspace{5pt}

These findings also reflect in other multi-objective metrics.
In terms of optimistic HV, \mocapointensify{} also achieves the highest value in 8 out of 12 model–dataset combinations.
In particular, on MBPP, \mocapointensify{} achieves performance improvements of up to a factor of 3 in HV compared to the initial instructions (for \gpt{}), highlighting substantial gains.

We now also discuss the three decision criteria by \citet{feurer-ida23a} (cf. \S\ref{sec:mo-metrics}). Following criterion (1), we find that \mocapointensify's pessimistic HV exceeds the optimistic HV of \mocaponsgaII{} in 5 out of 12 cases, while the opposite relationship only occurs once. At the same time, \mocapointensify's pessimistic HV exceeds the optimistic HV of the initial instructions in 10 out of 12 cases, with only one opposite case. Regarding criterion (2), we find that \mocapointensify{} pessimistic fronts dominate \mocaponsgaII{} optimistic fronts on 5 out of 36 seeds, while the reverse case happens only once.\footnote{For details, we refer to the supplementary material.} Lastly, criterion (3) assesses the robustness of a solution set. We observe relatively small approximation gaps for both \mocapointensify{} and \mocaponsgaII{}.
This indicates that the discovered fronts generalize well from development to test data and are not driven by a few unstable boundary solutions. 
Notable exceptions appear on the saturated AG News task and on MBPP for \mocapointensify{}.
These findings are in line with those from the original work introducing these metrics: domination of the optimistic HV by the pessimistic HV is likely, but dominance of the Pareto front is rather unlikely.

Overall, these results show that \mocapointensify{} consistently discovers strong, robust Pareto fronts at a budget of 7.5M tokens, particularly on more complex tasks, often outperforming its competitor.

\begin{figure}[h]
    \centering
    \includegraphics[width=0.6\textwidth]{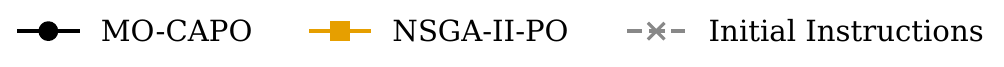}

    \begin{subfigure}[b]{0.3\textwidth}
        \centering
        \includegraphics[height=4.2cm]{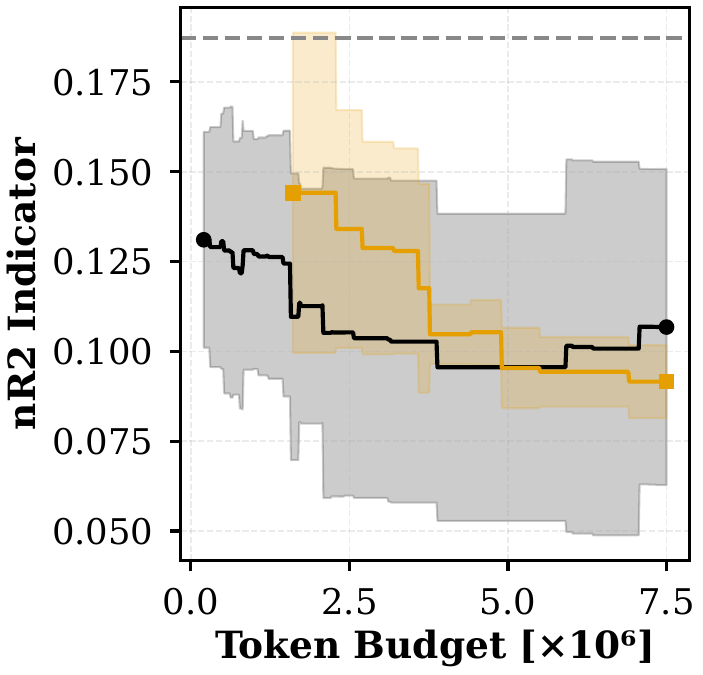}
        \caption{\mistral{} on MBPP}
        \label{fig:traj-grid-a}
        \Description[Optimization trace of Mistral-32-24B on MBPP.]{Optimization trace of Mistral-32-24B on MBPP for the nR2 indicator. Both methods converge to a similar score, but MO-CAPO exhibits a higher variance.}
    \end{subfigure}
    \hfill
    \begin{subfigure}[b]{0.3\textwidth}
        \centering
        \includegraphics[height=4.2cm]{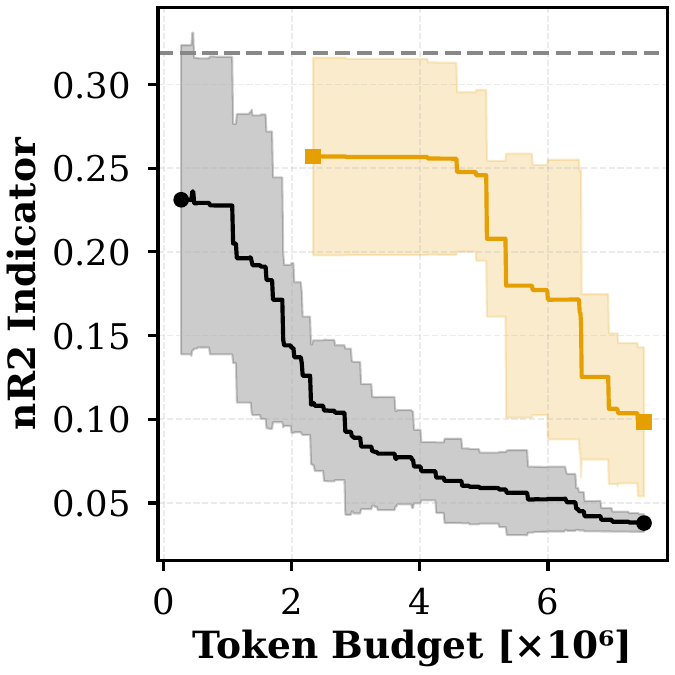}
        \caption{\gpt{} on GSM8K}
        \label{fig:traj-grid-b}
        \Description[Optimization trace of GPT-OSS-120B on GSM8K.]{Optimization trace of GPT-OSS-120B on GSM8K for the nR2 indicator. MO-CAPO converges significantly faster to a lower score than NSGA-II-PO.}
    \end{subfigure}
    \hfill
    \begin{subfigure}[b]{0.3\textwidth}
        \centering
        \includegraphics[height=4.2cm]{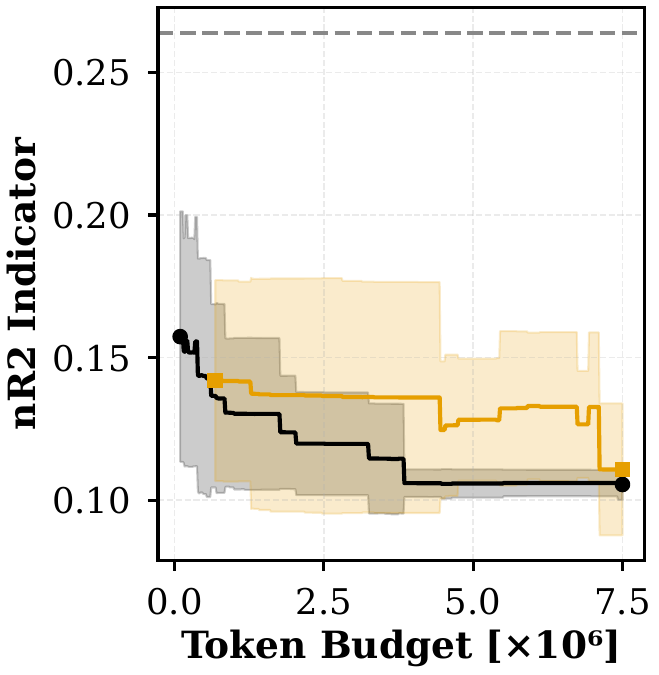}
        \caption{\qwen{} on Subj}
        \label{fig:traj-grid-c}
        \Description[Optimization trace of Qwen3-30B on Subj.]{Optimization trace of Qwen3-30B on Subj for the nR2 indicator. MO-CAPO converges significantly faster than NSGA-II-PO and exhibits substantially less variance.}
    \end{subfigure}

    \caption{Optimization trajectories for the nR2 indicator across three dataset/model combinations. Lines and shaded regions denote mean $\pm$ std across three independent runs.}
    \label{fig:traj-grid}
\end{figure}

\paragraph{Optimization Trajectories.}
Despite promising results at the full token budget, we are also interested in evaluations at intermediate budgets.
The optimization trajectories for nR2 in Figure~\ref{fig:traj-grid} reveal that the cost-efficient design of \mocapointensify{} enables it to produce competitive solutions already at small budgets.
Across the three selected representative settings, \mocapointensify{} is always the first optimizer to finalize the first optimization step and return an improved solution set.
Moreover, it often reaches near-final performances already well before $4\times10^6$ tokens (Figure~\ref{fig:traj-grid-a} and \ref{fig:traj-grid-c}) and dominates \mocaponsgaII{} over the entire budget range, with one exception for higher budgets in Figure~\ref{fig:traj-grid-a}.
Particularly, in Figure~\ref{fig:traj-grid-b}, \mocapointensify{} finds a solution set with $\text{nR2}\approx0.17$ on average already before 2M input tokens, while \mocaponsgaII{} requires roughly thrice the budget for the same performance.  
Within the first tokens, \mocapointensify{} sharply reduces nR2, while \mocaponsgaII{} either improves slowly or remains largely flat.
Trajectory plots for the remaining settings in \S\ref{app:further-results} confirm these findings.

\findingbox{\mocapointensify{} finds competitive solutions already at much lower budgets compared to \mocaponsgaII{}, underscoring its efficiency.}

\begin{figure}[h]
    \centering
    \includegraphics[width=0.95\textwidth]{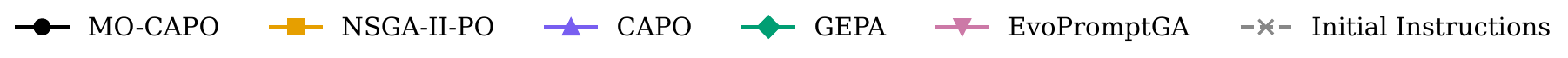} 
    \begin{subfigure}[b]{0.3\textwidth}
        \centering
        \includegraphics[height=4.5cm]{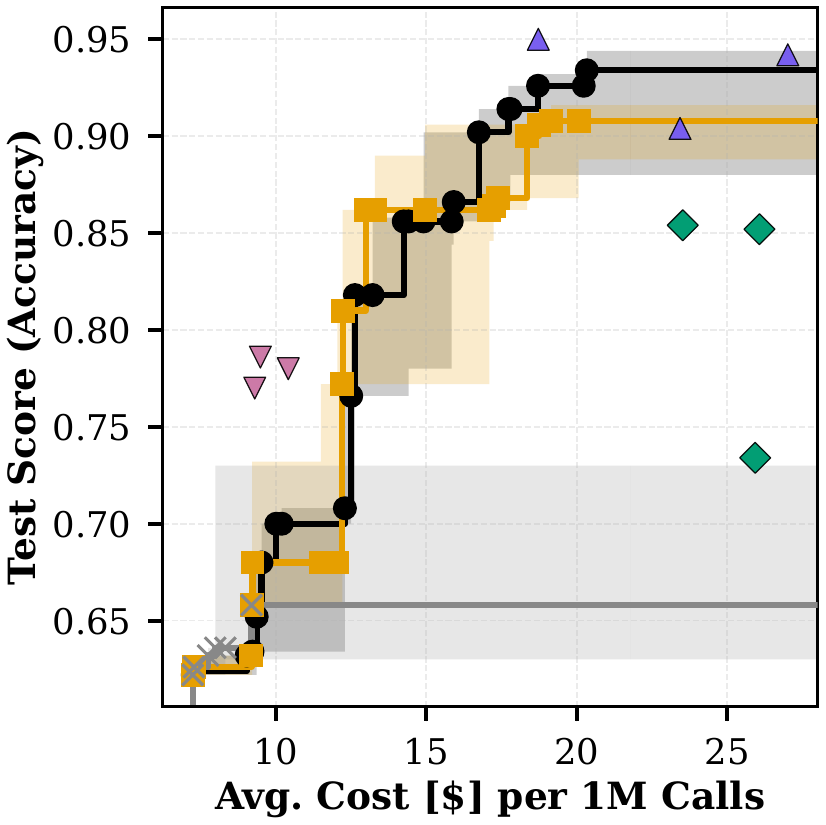}
        \caption{\mistral{} on Subj}
        \label{fig:eas-a}
        \Description[Empirical attainment surface for Mistral-3.2-24B on Subj.]{Empirical attainment surface for Mistral-3.2-24B on Subj. The attainment surfaces of MO-CAPO and NSGA-II-PO  are very similar, but MO-CAPO achieves a higher accuracy for higher token costs. The initial instructions performs significantly worse, and the solution of single-objective optimization algorithms CAPO, GEPA and EvoPromptGA exhibit very different tradeoffs between accuracy and token costs.}
    \end{subfigure}
    \hspace{1.5cm} 
    \begin{subfigure}[b]{0.3\textwidth} 
        \centering
        \includegraphics[height=4.5cm]{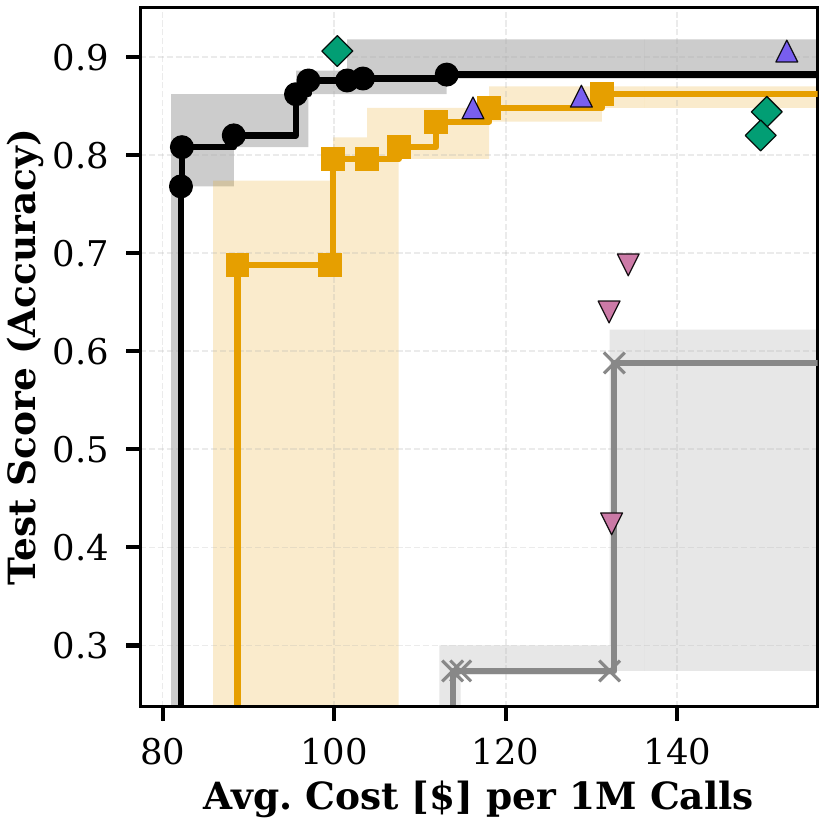}
        \caption{\gpt{} on GSM8K}
        \label{fig:eas-b}
        \Description[Empirical attainment surface for GPO-OSS-120B on GSM8K.]{Empirical attainment surface for GPO-OSS-120B on GSM8K. The attainment surface of MO-CAPO dominates the attainment surface of NSGA-II-PO. The initial instructions performs significantly worse, and the solution of single-objective optimization algorithms CAPO, GEPA and EvoPromptGA exhibit very different tradeoffs between accuracy and token costs.}
    \end{subfigure}
    \caption{Empirical attainment surfaces for MO optimizer at a budget of 7.5M tokens. Lines indicate median attainment across three independent seeds, shaded bands span minimum to maximum attainment, showing the variability across seeds. For single-objective optimizers, we plot the single solutions from each seed.}
    \label{fig:eas}
\end{figure}

\paragraph{Attainment Surfaces.}
The EAS plots in Figure~\ref{fig:eas} after the full budget provide more detailed insights into the different objective trade-offs the final solutions entail.
We see that multi-objective optimizers find diverse fronts of different trade-offs, covering both low-cost solutions with reasonable accuracy and more costly high-accuracy solutions.
The mostly narrow bands of the EAS plots indicate relatively stable results across seeds.
Initial instructions are either limited to lower-cost areas or entirely dominated, and \mocapointensify{} fronts often partly exceed or even dominate \mocaponsgaII{} fronts.
Although solutions found by single-objective optimizers sometimes dominate solutions in the multi-objective fronts, they clearly represent fixed trade-offs: for example, EvoPrompt, as an instruction-only optimizer, finds only low-cost solutions with medium performance in Figure~\ref{fig:eas-a}, while CAPO yields the best performing but also expensive solutions.
\mocapointensify{} finds solutions in both areas and also in between, which are highly relevant from a practical perspective: in many deployment scenarios, marginal gains in accuracy do not justify substantial increases in inference cost.
By explicitly optimizing the performance–cost trade-off, \mocapointensify{} provides practitioners with viable low-cost alternatives that are ignored by performance-only methods. %
Further EAS plots in \S\ref{app:further-results} confirm these findings. 
Nonetheless, it is important to consider that EAS mask the pessimistic HV as they only show non-dominated solutions.

\findingbox{\mocapointensify{} discovers a wide variety of performance-cost-trade-offs, exploiting both low-cost solutions omitted by single-objective optimizers as well as highly performant ones.}

\begin{figure}[b]
    \centering
    \begin{subfigure}[b]{0.3\textwidth}
        \centering
        \includegraphics[width=\linewidth]{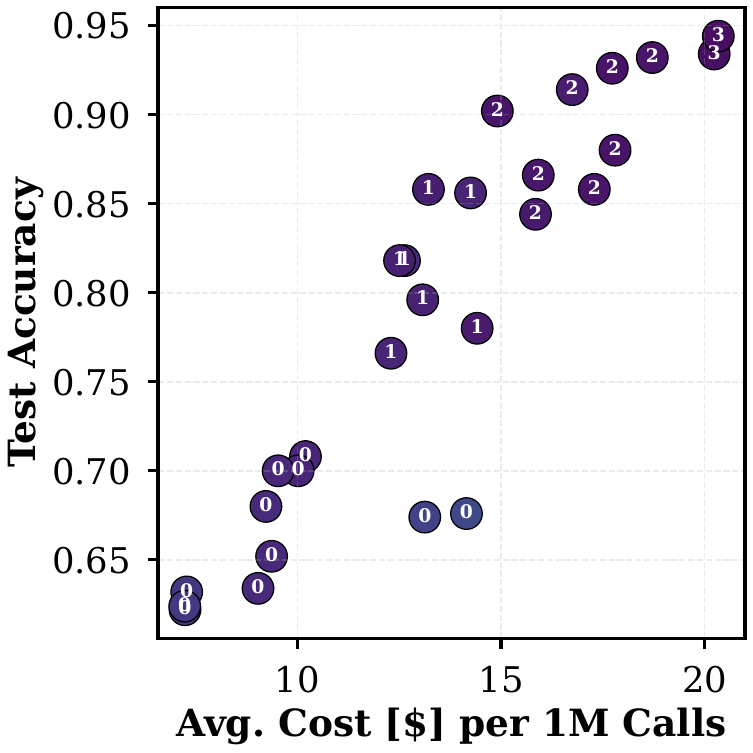}
        \caption{\mistral{} on Subj}
        \label{fig:mistral_fs}
        \Description[Few-shot example counts for Mistral-32-24B on Subj.]{Few-shot example counts for Mistral-32-24B on Subj. The plot shows the tradeoffs between token cost and test accuracy, and displays varying tradeoffs when using zero and up to three few-shot examples.}
    \end{subfigure}
    \hfill 
    \begin{subfigure}[b]{0.3\textwidth}
        \centering
        \includegraphics[width=\linewidth]{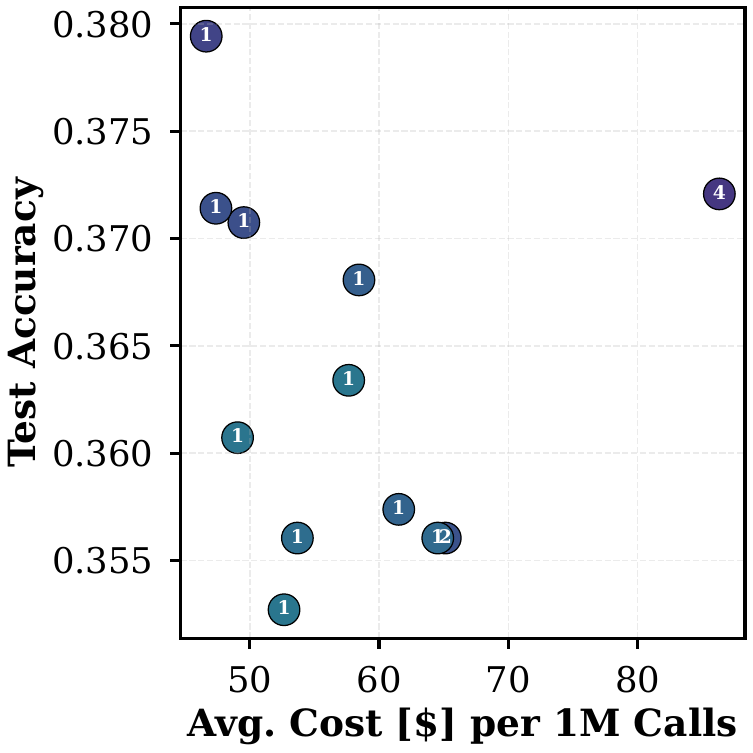}
        \caption{\qwen{} on MBPP}
        \label{fig:qwen_fs}
        \Description[Few-shot example counts for Qwen3-30B on MBPP.]{Few-shot example counts for Qwen3-30B on MBPP. The plot shows the tradeoffs between token cost and test accuracy, and displays varying tradeoffs when using zero and up to four few-shot examples. Interestingly, using more than one few-shot example does not lead to improved performance.}
    \end{subfigure}
    \hfill
    \begin{subfigure}[b]{0.3\textwidth}
        \centering
        \includegraphics[width=\linewidth]{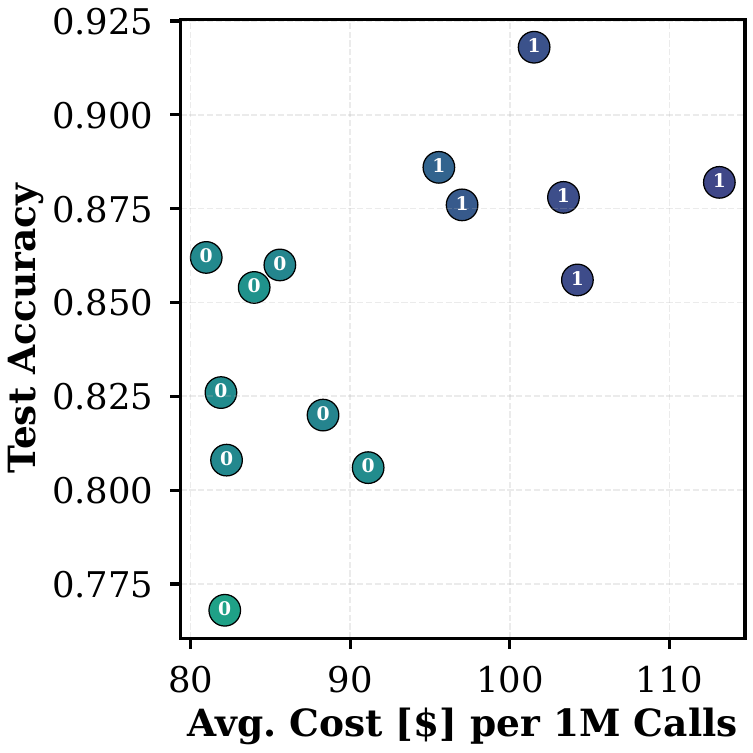}
        \caption{\gpt{} on GSM8K} %
        \label{fig:gpt_oss_fs}
        \Description[Few-shot example counts for GPT-OSS-120B on GSM8K.]{Few-shot example counts for GPT-OSS-120B on GSM8K. The plot shows the tradeoffs between token cost and test accuracy, and displays varying tradeoffs when using zero and one few-shot examples. Interestingly, using more than one few-shot example does not lead to improved performance, and there are two different clusters for prompt with zero and one few-shot example.}
    \end{subfigure}
    \begin{subfigure}[b]{0.08\textwidth}
        \includegraphics[width=\textwidth]{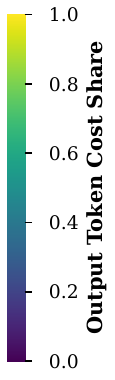}
        \vspace{.65cm}
    \end{subfigure}
    \caption{Few-shot example counts for \mocapointensify{} for final incumbent set members of all three random seeds. Reported objective values are evaluated on the test set. Example counts are displayed as numbers, and the color scale indicates the share of costs produced by output tokens.}
    \label{fig:fs_counts}
\end{figure}

\paragraph{Prompt Composition Analysis.}
Next, we examine the diversity of the prompt compositions in the solution sets.
Prompts can be diverse in (i) the number of few-shot examples, (ii) the instruction's level of detail, and (iii) the verbosity of the associated output (direct output vs. reasoning).
We analyze these aspects in Figure~\ref{fig:fs_counts} by displaying the number of few-shot examples in the prompt and the share of cost associated with the output tokens vs. the input tokens.
Regarding the number of few-shot examples, we find that \mocapointensify{} often discovers a diverse set with varying numbers of examples (e.g., between 0 and 3 in Figure~\ref{fig:mistral_fs}), where prompts with more examples naturally incur higher costs.
Nonetheless, prompts with the same number of examples often incur different costs and even prompts with fewer examples (e.g., zero-shot) can have higher costs than prompts with more examples (e.g., one-shot; see Figure~\ref{fig:mistral_fs}), suggesting differences in (ii) and (iii).
For the ratio of input to output tokens, we find small differences for prompts with the same example count and large differences across tasks.
Nonetheless, this analysis highlights the diversity in prompt composition across the solution sets. Concrete prompt examples are provided in \S\ref{app:prompt-examples}.

\findingbox{\mocapointensify{} discovers diverse prompt sets with a varying number of few-shot examples.}

\begin{table}[b]
\centering
\small
\caption{Accuracy achieved by each optimizer at 7.5M token budget (bold = best accuracy per model-dataset). We select the best prompt based on dev-performance and report the test-performance.}
\label{tab:boundary}
\begin{adjustbox}{max width=\textwidth}
\setlength{\tabcolsep}{2pt}
\renewcommand{\arraystretch}{0.875}
\begin{tabular}{ll cccc}
\toprule
\textbf{Model} & \textbf{Optimizer} & \textbf{GSM8K} & \textbf{Subj} & \textbf{AG News} & \textbf{MBPP} \\
\midrule
\multirow{6}{*}{\gpt} & Initial & $0.495_{\pm.192}$ & $0.741_{\pm.021}$ & $0.856_{\pm.005}$ & $0.072_{\pm.026}$ \\
 & CAPO & $0.871_{\pm.031}$ & $0.858_{\pm.004}$ & $\mathbf{0.869}_{\pm.011}$ & $0.169_{\pm.021}$ \\
 & GEPA & $0.857_{\pm.044}$ & $0.870_{\pm.032}$ & $0.863_{\pm.017}$ & $0.253_{\pm.107}$ \\
 & EvoPromptGA & $0.584_{\pm.141}$ & $0.744_{\pm.017}$ & $0.861_{\pm.009}$ & $0.080_{\pm.016}$ \\
 & \mocaponsgaII{} & $0.852_{\pm.009}$ & $\mathbf{0.871}_{\pm.008}$ & $0.855_{\pm.014}$ & $0.165_{\pm.010}$ \\
 & \mocapointensify{} & $\mathbf{0.893}_{\pm.022}$ & $0.869_{\pm.007}$ & $0.867_{\pm.013}$ & $\mathbf{0.257}_{\pm.062}$ \\
\cmidrule{1-6}
\multirow{6}{*}{\mistral} & Initial & $0.463_{\pm.013}$ & $0.673_{\pm.052}$ & $0.859_{\pm.011}$ & $0.304_{\pm.022}$ \\
 & CAPO & $\mathbf{0.752}_{\pm.020}$ & $\mathbf{0.932}_{\pm.025}$ & $0.844_{\pm.023}$ & $\mathbf{0.331}_{\pm.002}$ \\
 & GEPA & $0.571_{\pm.146}$ & $0.813_{\pm.069}$ & $\mathbf{0.860}_{\pm.018}$ & $0.318_{\pm.006}$ \\
 & EvoPromptGA & $0.439_{\pm.031}$ & $0.779_{\pm.008}$ & $0.856_{\pm.016}$ & $0.301_{\pm.029}$ \\
 & \mocaponsgaII{} & $0.743_{\pm.045}$ & $0.904_{\pm.014}$ & $0.849_{\pm.014}$ & $\mathbf{0.331}_{\pm.000}$ \\
 & \mocapointensify{} & $0.751_{\pm.036}$ & $0.919_{\pm.034}$ & $0.848_{\pm.016}$ & $0.320_{\pm.017}$ \\
\cmidrule{1-6}
\multirow{6}{*}{\qwen} & Initial & $0.540_{\pm.022}$ & $0.754_{\pm.044}$ & $0.850_{\pm.009}$ & $0.329_{\pm.006}$ \\
 & CAPO & $\mathbf{0.665}_{\pm.025}$ & $\mathbf{0.909}_{\pm.033}$ & $\mathbf{0.900}_{\pm.014}$ & $\mathbf{0.381}_{\pm.010}$ \\
 & GEPA & $0.659_{\pm.024}$ & $0.775_{\pm.078}$ & $0.870_{\pm.004}$ & $0.360_{\pm.026}$ \\
 & EvoPromptGA & $0.537_{\pm.010}$ & $0.783_{\pm.019}$ & $0.860_{\pm.029}$ & $0.370_{\pm.004}$ \\
 & \mocaponsgaII{} & $0.657_{\pm.033}$ & $0.885_{\pm.030}$ & $0.893_{\pm.010}$ & $0.372_{\pm.014}$ \\
 & \mocapointensify{} & $0.662_{\pm.036}$ & $0.885_{\pm.008}$ & $0.885_{\pm.009}$ & $0.369_{\pm.005}$ \\
\bottomrule
\end{tabular}
\end{adjustbox}
\end{table}

\begin{figure}[ht]
    \centering
    \includegraphics[width=0.8\linewidth]{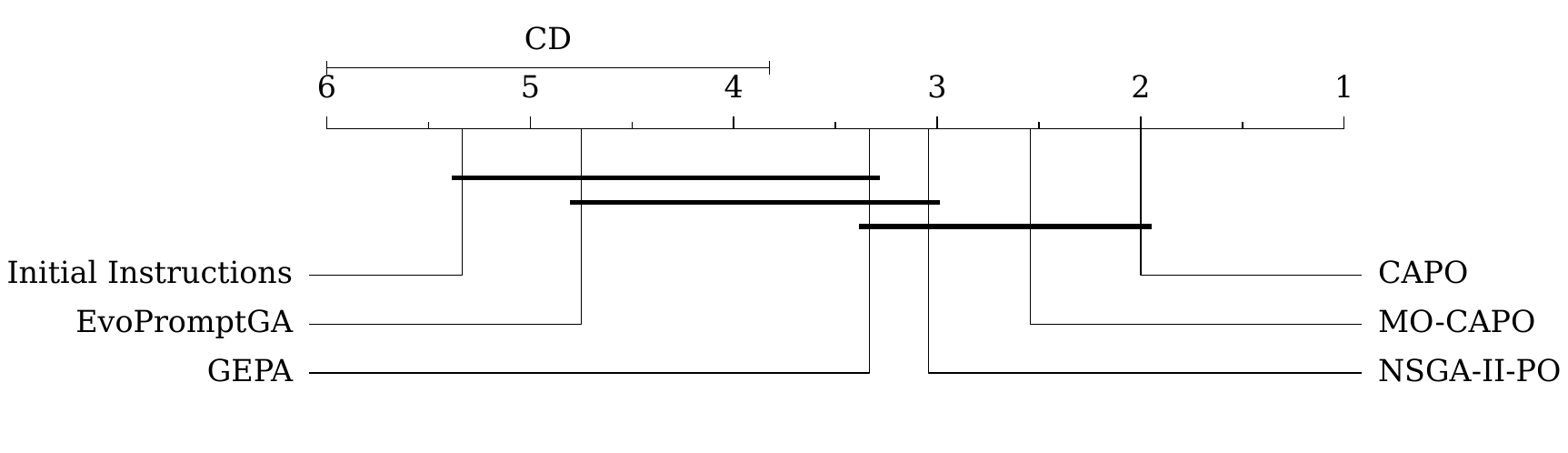}
    \caption{
    Critical difference diagram ($\alpha=0.05$) for test accuracy. 
    Ranks are computed per model--dataset combination based on accuracy and averaged across seeds. Horizontal bars connect methods that are not significantly different according to the Friedman test, followed by a post-hoc Nemenyi test.
    }
    \Description[Critical difference diagram for test accuracy.]{Critical difference diagram for test accuracy. The ranks (from best to worst) are CAPO, MO-CAPO, NSGA-II-PO, GEPA, EvoPromptGA and Initial Instructions. Importantly, CAPO, MO-CAPO and NSGA-II-PO exhibit a very similar performance.}
    \label{fig:cd_accuracy}
\end{figure}

\paragraph{Accuracy Comparison with Single-objective Optimizers.}
Lastly, we examine how the maximum-accuracy prompts discovered by the multi-objective methods compare with those of SOTA single-objective optimizers.
Table~\ref{tab:boundary} confirms that \mocapointensify{} is highly competitive in terms of accuracy, yielding the best performance in 2/12 cases and showing results close to the best (mostly within one standard deviation) in the remaining settings.
This makes \mocapointensify{} ranked second on average, behind CAPO and ahead of GEPA, as shown in the critical differences plot in Figure~\ref{fig:cd_accuracy}.
Importantly, this plot shows that at the $\alpha = 0.05$ level, there are no significant differences in accuracy between our multi-objective methods and the SOTA single-objective optimizers CAPO and GEPA, as determined by a post-hoc Nemenyi test~\citep{demsar-jmlr06a,herbold-joss20a}. \mocapointensify{} significantly outperforms both EvoPrompt and the initial instructions on this single objective.

\findingbox{The maximum accuracy solutions found by \mocapointensify{} are highly competitive with state-of-the-art single-objective optimizers.}

\subsection{Ablation Studies}

To better understand the new components in \mocapointensify{}, we present ablation studies of the intensification mechanism and the cost-objective, and summarize results from ablations on minor components, such as the parent selection mechanism.

\begin{table}[b]
\centering
\caption{Efficiency comparison between \mocapointensify{} and \mocaponsgaII{} across all datasets and models at 7.5M token budget.
\#Cand is the average number of evaluated candidates.
B/Cand is the average number of blocks on which candidates were evaluated.
Iter1 is the budget consumed (in thousand tokens) when the first iteration is completed.
TT80 is time-to-80\%, i.e., the budget (in \% of the total budget) required to achieve 80\% of the final pessimistic HV.
Bold values indicate the best per model-dataset pair.}
\label{tab:efficiency}
\begin{adjustbox}{max width=\textwidth}
\setlength{\tabcolsep}{2pt}
\renewcommand{\arraystretch}{0.875}
\begin{tabular}{cl cccc cccc}
\toprule
& & \multicolumn{4}{c}{\textbf{GSM8K}} & \multicolumn{4}{c}{\textbf{Subj}} \\
\cmidrule(lr){3-6} \cmidrule(lr){7-10}
\textbf{Model} & \textbf{Optimizer} & \textbf{\#Cand} & \textbf{B/Cand} & \textbf{Iter1 (kTok)} $\downarrow$ & \textbf{TT80 (\%)} $\downarrow$ & \textbf{\#Cand} & \textbf{B/Cand} & \textbf{Iter1 (kTok)} $\downarrow$ & \textbf{TT80 (\%)} $\downarrow$ \\
\midrule
\multirow{2}{*}{\gpt} 
& \mocapointensify{} 
& $209_{\pm\phantom{0}15}$ 
& $3.0_{\pm.3}$ 
& $\mathbf{\phantom{0}296}_{\pm\phantom{0}21}$ 
& $\mathbf{\phantom{0}32.3}_{\pm18.6}$ 
& $237_{\pm\phantom{0}8}$ 
& $3.3_{\pm.4}$ 
& $\mathbf{\phantom{0}251}_{\pm\phantom{0}26}$ 
& $\mathbf{\phantom{0}\phantom{0}3.3}_{\pm\phantom{00}.3}$ \\

& \mocaponsgaII{} 
& $\phantom{0}46_{\pm\phantom{0}8}$ 
& --- 
& $2503_{\pm278}$ 
& $\phantom{0}90.8_{\pm\phantom{0}8.0}$ 
& $\phantom{0}58_{\pm\phantom{0}7}$ 
& --- 
& $2070_{\pm169}$ 
& $\phantom{0}27.3_{\pm\phantom{0}2.3}$ \\

\cmidrule{1-10}

\multirow{2}{*}{\mistral} 
& \mocapointensify{} 
& $168_{\pm\phantom{0}18}$ 
& $3.1_{\pm.3}$ 
& $\mathbf{\phantom{0}332}_{\pm\phantom{0}53}$ 
& $\mathbf{\phantom{0}\phantom{0}8.4}_{\pm\phantom{0}7.5}$ 
& $442_{\pm\phantom{0}30}$ 
& $3.4_{\pm.1}$ 
& $\mathbf{\phantom{0}109}_{\pm\phantom{0}10}$ 
& $\mathbf{\phantom{0}34.5}_{\pm56.7}$ \\

& \mocaponsgaII{} 
& $\phantom{0}45_{\pm\phantom{0}10}$ 
& --- 
& $2683_{\pm356}$ 
& $\phantom{0}37.3_{\pm\phantom{0}8.5}$ 
& $155_{\pm\phantom{0}19}$ 
& --- 
& $\phantom{0}829_{\pm\phantom{0}34}$ 
& $\phantom{0}37.8_{\pm41.7}$ \\

\cmidrule{1-10}

\multirow{2}{*}{\qwen} 
& \mocapointensify{} 
& $234_{\pm\phantom{0}94}$ 
& $2.1_{\pm.5}$ 
& $\mathbf{\phantom{0}350}_{\pm\phantom{0}39}$ 
& $\mathbf{\phantom{0}\phantom{0}7.2}_{\pm\phantom{0}4.8}$ 
& $530_{\pm109}$ 
& $3.0_{\pm.4}$ 
& $\mathbf{\phantom{0}100}_{\pm\phantom{0}\phantom{0}7}$ 
& $\mathbf{\phantom{0}\phantom{0}6.0}_{\pm\phantom{0}5.0}$ \\

& \mocaponsgaII{} 
& $\phantom{0}41_{\pm\phantom{0}6}$ 
& --- 
& $2836_{\pm156}$ 
& $\phantom{0}37.1_{\pm\phantom{0}2.5}$ 
& $187_{\pm\phantom{0}48}$ 
& --- 
& $\phantom{0}725_{\pm\phantom{0}44}$ 
& $\phantom{0}25.8_{\pm28.2}$ \\

\midrule
& & \multicolumn{4}{c}{\textbf{AG News}} & \multicolumn{4}{c}{\textbf{MBPP}} \\
\cmidrule(lr){3-6} \cmidrule(lr){7-10}

\multirow{2}{*}{\gpt} 
& \mocapointensify{} 
& $349_{\pm170}$ 
& $3.0_{\pm1.0}$ 
& $\mathbf{\phantom{0}187}_{\pm\phantom{0}23}$ 
& $\mathbf{\phantom{0}\phantom{0}2.9}_{\pm\phantom{0}1.0}$ 
& $218_{\pm\phantom{0}19}$ 
& $2.9_{\pm.3}$ 
& $\mathbf{\phantom{0}292}_{\pm\phantom{0}33}$ 
& $\mathbf{\phantom{0}70.6}_{\pm25.8}$ \\

& \mocaponsgaII{} 
& $\phantom{0}90_{\pm\phantom{0}4}$ 
& --- 
& $1463_{\pm\phantom{0}28}$ 
& $\phantom{0}19.0_{\pm\phantom{0}\phantom{0}.8}$ 
& $\phantom{0}55_{\pm\phantom{0}2}$ 
& --- 
& $2134_{\pm143}$ 
& $100.0_{\pm\phantom{0}\phantom{0}.0}$ \\

\cmidrule{1-10}

\multirow{2}{*}{\mistral} 
& \mocapointensify{} 
& $513_{\pm147}$ 
& $2.9_{\pm.6}$ 
& $\mathbf{\phantom{0}130}_{\pm\phantom{0}11}$ 
& $\mathbf{\phantom{0}34.4}_{\pm56.8}$ 
& $233_{\pm\phantom{0}92}$ 
& $2.7_{\pm1.1}$ 
& $\mathbf{\phantom{0}244}_{\pm\phantom{0}53}$ 
& $\mathbf{\phantom{0}\phantom{0}4.0}_{\pm\phantom{0}2.0}$ \\

& \mocaponsgaII{} 
& $170_{\pm\phantom{0}32}$ 
& --- 
& $1149_{\pm138}$ 
& $\phantom{0}42.8_{\pm49.5}$ 
& $\phantom{0}62_{\pm\phantom{0}12}$ 
& --- 
& $2003_{\pm470}$ 
& $\phantom{0}34.0_{\pm13.2}$ \\

\cmidrule{1-10}

\multirow{2}{*}{\qwen} 
& \mocapointensify{} 
& $367_{\pm\phantom{0}14}$ 
& $3.3_{\pm.6}$ 
& $\mathbf{\phantom{0}133}_{\pm\phantom{0}25}$ 
& $\mathbf{\phantom{0}\phantom{0}1.8}_{\pm\phantom{0}.3}$ 
& $221_{\pm\phantom{0}42}$ 
& $3.0_{\pm.6}$ 
& $\mathbf{\phantom{0}221}_{\pm\phantom{0}28}$ 
& $\phantom{0}90.2_{\pm17.1}$ \\

& \mocaponsgaII{} 
& $141_{\pm\phantom{0}36}$ 
& --- 
& $1064_{\pm\phantom{0}57}$ 
& $\phantom{0}13.9_{\pm\phantom{0}.7}$ 
& $\phantom{0}61_{\pm\phantom{0}6}$ 
& --- 
& $1815_{\pm330}$ 
& $\mathbf{\phantom{0}74.7}_{\pm35.5}$ \\

\bottomrule
\end{tabular}
\end{adjustbox}
\end{table}

\paragraph{Impact of Intensification}
Although we introduced \mocaponsgaII{} as a baseline for benchmarking, it also serves as an ablation, since its main difference to \mocapointensify{} is the intensification mechanism.
Thus, a comparison isolates the contribution of this budget-allocation component.
The benchmark results (Table~\ref{tab:results_main}) showed that \mocapointensify{} mostly outperforms \mocaponsgaII{} at the full budget, and trajectory plots (Figure~\ref{fig:traj-grid}) demonstrated that \mocapointensify{} finds competitive solution sets earlier -- gains stemming from intensification and partial evaluation, allocating the budget intelligently.

We now provide a more detailed analysis of the efficiency gains through intensification in Table~\ref{tab:efficiency}.
First, the efficient budget allocation allows \mocapointensify{} to discover up to a factor of $5.7$ more candidates, enabling broader exploration of the search space.
This is because it performs only around 3 block evaluations (90 samples) per candidate on average, whereas \mocaponsgaII{} without intensification always performs full evaluations (300 samples).
Moreover, Table~\ref{tab:efficiency} confirms that the average budget required to yield a first solution set after the first step is reduced by up to 88\% through intensification and initial evaluation on only a single block.
In addition, \mocapointensify{} converges faster, often requiring less than 1/3 of the token budget to achieve 80\% of the final performance. Without intensification, this takes substantially longer in most cases. 
These findings confirm the intended effect of intensification.

\findingbox{\mocapointensify's intensification substantially improves efficiency by enabling broader exploration, considerably reducing the budget needed for the first solution set, and achieving near-final performance with less token budget.}

\paragraph{Effect of Cost Weights}
Another novel component of \mocapointensify{} is the cost objective.
We isolate the influence of the input and output token weights by systematically setting them to zero.
The experiments are conducted on GSM8K and Subj with \qwen{} and a budget of 7.5M tokens.
For comparability, the nR2 metric is always computed using the default cost weights. 

The results in Table~\ref{tab:token_increase} show that explicitly penalizing input tokens is essential to control prompt length.
When setting $w_{\text{in}} = 0$, input token usage increases dramatically compared to the default (+118.9\% on GSM8K and +225.7\% on Subj).
In contrast, removing only the output penalty ($w_{\text{out}} = 0$) has a comparatively modest effect on output token usage (+0.93\% on GSM8K and +38.8\% on Subj).
When both weights are removed, the increase in input tokens remains high while the changes in output are comparably small.
This behavior is intuitive, as the optimizer directly controls prompt length, while output length is only indirectly influenced by the instruction design.
The impact is also reflected in the optimization quality. Removing the input penalty substantially degrades nR2, while removing only the output penalty leaves nR2 nearly unchanged, suggesting that input control is the dominant driver of the cost objective.
Since setting both cost weights to 0 reduces \mocapointensify{} to a single-objective optimizer, we compare its accuracy with other SOTA single-objective optimizers in \S\ref{app:further-ablations}. The results show that it is the best-performing optimizer together with CAPO, outperforming GEPA and EvoPrompt. This demonstrates its applicability to single-objective settings.

\begin{table}[tb]
\centering
\caption{Token increase analysis for cost ablation experiments on \qwen{} at 7.5M token budget.
We measure the percentage increase in average input/output token increase compared to \mocapointensify{} with cost weights active.
nR2 measures optimization quality (lower is better).}
\label{tab:token_increase}
\begin{adjustbox}{max width=\textwidth}
\setlength{\tabcolsep}{4pt}
\renewcommand{\arraystretch}{0.9}
\small
\begin{tabular}{llccc}
\toprule
\textbf{Dataset} & \textbf{Experiment} & \textbf{Input} (\%) & \textbf{Output} (\%) & \textbf{nR2} $\downarrow$ \\
\midrule
\multirow{4}{*}{GSM8K} 
& \mocapointensify{} (default) & - & - & $\mathbf{0.340}_{\pm.006}$ \\
& $w_{\text{in}}=0$ & $125.03_{\pm48.14}$ & $-6.14_{\pm0.72}$ & $0.434_{\pm.060}$ \\
& $w_{\text{out}}=0$ & $\phantom{00}5.69_{\pm\phantom{0}8.90}$ & $\phantom{00}0.93_{\pm1.94}$ & $0.363_{\pm.008}$ \\
& $w_{\text{in}}=0, w_{\text{out}}=0$ & $112.50_{\pm27.00}$ & $-5.08_{\pm1.08}$ & $0.439_{\pm.071}$ \\
\cmidrule{1-5}
\multirow{4}{*}{Subj} 
& \mocapointensify{} (default) & - & - & $0.106_{\pm.005}$ \\
& $w_{\text{in}}=0$ & $225.13_{\pm38.11}$ & $-0.45_{\pm13.57}$ & $0.210_{\pm.028}$ \\
& $w_{\text{out}}=0$ & $\phantom{00}4.83_{\pm14.23}$ & $\phantom{0}40.90_{\pm45.90}$ & $\mathbf{0.105}_{\pm.003}$ \\
& $w_{\text{in}}=0, w_{\text{out}}=0$ & $242.58_{\pm36.19}$ & $\phantom{0}10.37_{\pm\phantom{0}7.06}$ & $0.219_{\pm.017}$ \\
\bottomrule
\end{tabular}
\end{adjustbox}
\end{table}

\findingbox{\mocapointensify's cost objective, primarily the input token weight, is crucial to control prompt cost.}

\paragraph{Further Ablations \& Sensitivity Analysis}
We further ablate the influence of the parent selection (see \S\ref{app:further-ablations}), which shows rather limited impact.
We also conduct a study on the sensitivity of \mocapointensify{} to its hyperparameters (block size, population size, number of crossovers; see \S\ref{app:hp}), demonstrating that \mocapointensify{} is relatively robust w.r.t. the examined hyperparameters, as performance remains stable across a broad range of configurations.

\section{Conclusion \& Future Work}

\subsection{\mocapointensify{}: Summary, Implications \& Future Directions}

We introduced \mocapointensify{}, a multi-objective prompt optimization algorithm that jointly optimizes performance and inference cost while explicitly accounting for the high evaluation cost inherent in prompt optimization. The method extends CAPO with true multi-objective optimization and integrates a budget-aware intensification mechanism inspired by MO-SMAC, enabling efficient exploration of the prompt search space under limited evaluation budgets. In addition, we proposed a deployment-oriented cost objective that captures both input and output token consumption, providing a more realistic representation of inference cost than common prompt-length proxies.

Extensive experiments across three LLMs and four tasks demonstrated that \mocapointensify{} consistently discovers well-performing Pareto front approximations.
The method outperforms an NSGA-II-based baseline in the majority of evaluated settings, mostly with substantial improvements over initial, unoptimized instructions, as measured by multi-objective metrics such as nR2 and optimistic and pessimistic HV. We demonstrated that the discovered Pareto fronts exhibit high robustness in most cases as measured by the approximation gap.
Moreover, we showcased the efficiency of \mocapointensify{}, which finds competitive solution sets at much lower budgets compared to the NSGA-II baseline.
We found that \mocapointensify{} discovers a wide variety of performance-cost-trade-offs with both competitive low-cost solutions and more costly but highly performant prompts. The analyzed solution sets contain diverse prompt compositions in terms of instruction length, the number of few-shot examples, and the input/output token ratio.
Importantly, incorporating cost does not degrade performance. \mocapointensify{} remains competitive to SOTA single-objective optimizer and sometimes even superior in terms of accuracy rankings.
Our ablation studies confirmed that intensification is key to the method's efficiency, enabling broader exploration and faster convergence under a given budget. Furthermore, the proposed cost objective, particularly the input token weight, proved essential for controlling prompt cost.

\paragraph{Practical Implications} The cost-efficient multi-objective optimization provided by \mocapointensify{} is particularly useful for practitioners, who can now choose from a diverse set of solutions. In single-objective settings, they are constrained to a single solution focused on accuracy. However, in many practical settings, marginal accuracy gains do not justify a considerable increase in inference costs. Practitioners can now select the most appropriate trade-off for their use case themselves, from a diverse set of trade-offs found efficiently.

\paragraph{Future Work}
Future research can extend this line of work in several directions. First, instead of scalarizing input and output tokens, these could be treated as separate objectives, yielding higher-dimensional Pareto fronts. Second, additional objectives such as robustness across evaluation subsets, stability under stochastic decoding, inference latency, or fairness considerations could be incorporated. Importantly, \mocapointensify{} extends, in principle, to such other objectives without modifying the algorithm itself. Last, integrating surrogate modeling may further improve search efficiency in large prompt spaces. 

\subsection{Reflection on Metrics for Multi-Objective Generalization}
Beyond the algorithmic contribution, a central aspect of our work was the application of novel, not widely used metrics for measuring multi-objective generalization: nR2~\citep{branke-ieeetec25a} and the so-called optimistic and pessimistic HV~\citep{feurer-ida23a}. In our case, this evaluation better reflected the conditions practitioners face when selecting prompts in practice by explicitly accounting for solution robustness and stochastic variability.

More generally, we find that nR2 yields a single, easy-to-interpret number, but when comparing two methods, it is unclear how their Pareto front approximations compare. Similarly, comparing the pessimistic HV of one method with the optimistic HV of another provides no information about specific solutions on the Pareto front. However, this method appears to provide more conservative results than nR2. We hypothesize that, in many practical settings, when A's pessimistic HV exceeds B's optimistic HV, then method A will also tend to achieve a lower nR2 than method B. Dominance of A's pessimistic Pareto front approximation over B's optimistic Pareto front approximation most likely is even stronger.

\paragraph{Future Work}
Based on our findings, we think these metrics for multi-objective generalization demand further study, as machine learning is moving beyond solely performance-driven towards multi-objective settings. Additionally, the empirical attainment surfaces require extension to aggregate multiple optimistic and pessimistic Pareto fronts.

\begin{acks}
Tom Zehle received funding by the European Union. This work was supported by the European Union’s Horizon Europe research and innovation programm under grant agreement No. 101214398 (ELLIOT).
\end{acks}

\subsection*{Author Contribution Statement}
JB conceived the research idea, developed the algorithm, implemented a prototype, conducted preliminary experiments, designed the evaluation and visualizations, and drafted the manuscript. MS led the execution of the experiments and the structured collection and evaluation of results. TH led the writing of the manuscript and implemented the experimental setup. TZ implemented the proposed algorithm and baseline methods, and designed and implemented the benchmarks. MF conceived the research idea, conducted literature research, and supervised the project. JB continued to contribute to the processes led by the other authors throughout the project. All authors contributed to reviewing and editing the manuscript.
This work originated in the Master's thesis of JB, completed under the supervision of MF.

\bibliographystyle{ACM-Reference-Format}
\bibliography{bibliography/strings,bibliography/bib_prompt_tuning,bibliography/bib_automl,bibliography/proc,bibliography/my_proc}

\appendix

\newpage
\section{Appendix}

\subsection{Additional Background on Multi-Objective-Metrics} \label{app:add}

\subsubsection{Hypervolume Indicator}\label{app:hv}
To assess approximation quality without requiring the true front, the Hypervolume (HV) Indicator is commonly used \citep{zitzler-ieeeevocomp03a}. For approximation $\paretosetapprox$ and reference point $r \in \mathbb{R}^m$:
\begin{equation}
    \text{HV}_r(\paretosetapprox) := \text{vol}_m \left( \bigcup_{\prompt{} \in \paretosetapprox} \text{domHC}_r(\prompt{}) \right),
\end{equation}
where $\text{domHC}_r(\prompt{}) := \{u \in \mathbb{R}^m \mid f_j(\prompt{}) \leq u_j \leq r_j \;\forall j \in \{1, \ldots, m\}\}$ and $r$ dominates all $\paretosetapprox$ points. HV is strictly Pareto-compliant, meaning that any Pareto improvement strictly increases the HV, and promotes convergence and diversity~\citep{karl-telo23a}.

\subsubsection{Non-dominated Sorting}\label{app:nds}
Non-Dominated Sorting (NDS) partitions a set of candidate solutions into successive fronts according to Pareto dominance~\citep{deb-ieeeevocomp02a}.  
Given a set of solutions $P$, the first front $F_1$ consists of all non-dominated solutions in $P$. After removing $F_1$ from $P$, the second front $F_2$ contains the non-dominated solutions among the remaining candidates, and the process is repeated until all solutions are assigned to a front. A solution $\prompt{1}$ is dominating  another solution $\prompt{2}$ 
(denoted $\prompt{1} \prec \prompt{2}$) iff:
\begin{equation}
f_j(\prompt{1}) \le f_j(\prompt{2}) \;\; \forall j \in \{1,\dots,m\}
\quad \text{and} \quad
\exists j : f_j(\prompt{1}) < f_j(\prompt{2}).
\end{equation}

\subsubsection{Crowding Distance}\label{app:cd}
Since NDS does not provide a strict ordering within fronts, Crowding Distance (CD) is commonly applied to break potential ties inside fronts and to promote diversity~\citep{deb-ieeeevocomp02a}. The crowding distance of a solution $i$ is computed as
\begin{equation}
\text{CD}_i = \sum_{j=1}^{m}
\frac{f_j (\prompt{+}) - f_j(\prompt{-})}
     {f_j^{\max} - f_j^{\min}},
\end{equation}
where $\prompt{+}$ and $\prompt{-}$ denote the neighboring solutions of $\prompt{}$ in the sorted order for objective $j$. For a set of solutions within the same non-dominated front, the crowding distance of each solution is computed as the sum of normalized objective-wise distances to adjacent solutions. Solutions located in less crowded regions obtain larger crowding distances. During selection, candidates with higher CD are preferred, encouraging a well-distributed approximation of the Pareto front.

\subsection{Algorithm Details}\label{app:algo-details}

\subsubsection{\mocaponsgaII{} Algorithm Details} \label{app:nsgaii}
The full pseudo-code for our \mocaponsgaII{} baseline algorithm is given in Algorithm~\ref{algo:nsga-ii}.
Initialization via \HyperCall{initialize\_pop}{}{func:initialize_population} (see Algorithm~\ref{algo:initialization}) with the few-shot example generation is identical to CAPO. 
For few-shot creation, this means that the input of each sampled example is supplemented by reasoning as the corresponding output for the example, providing richer information than a label alone. To generate reasoning, the evaluation-LLM is prompted with the initial instruction to solve the example input. The LLM response typically contains both reasoning and prediction, which is used as output for the example. In case the LLM fails to produce a correct prediction, the true label $y$ is used as example output instead.

Furthermore, the evolutionary operators \HyperCall{crossover}{}{func:crossover} and \HyperCall{mutate}{}{func:mutate} (see Algorithm~\ref{algo:evol-operators}) are adapted from \citet{zehle-automl25a}.
To accommodate binary tournament selection, the crossover operation is slightly modified to take the objective vectors in $\mathcal{F}_\mu$ of the candidates as additional input.
Binary tournament selection can be imagined as a simplified form of Algorithm~\ref{algo:tournament_inten}, where $\incset=\emptyset$ and all candidates have the same (full) evaluation level. Then, the selection criterion in each tournament reduces to the standard NSGA-II criteria: best front rank with largest CD as a tie-breaker.
The \Call{evaluate}{}-function simply returns the objective vectors of all provided candidates on given instances. In case of \mocaponsgaII{}, all instances from the development set $\devset$ are used.
The function \Call{environmental\_selection}{} can be imagined as a simplified form of Algorithm~\ref{algo:env-selection}, where $\incset=\emptyset$ and all candidates have the same (full) evaluation level. Thus, the process reduces to selection based on front rank with CD as a tie-breaker.
\begin{algorithm}[h]
    \small
    \caption{\mocaponsgaII: Multi-Objective Prompt Optimization with NSGA-II}
    \begin{algorithmic}[1]
        \Require datasets $\devset$ \& $\fewshotset$,
        meta-LLM $\metallm$,
        evaluation-LLM $\evalllm$,
        initial instructions $\initialinstructions = \{\instruction{1}, \dots, \instruction{\populationsize}\}$,
        population size $\populationsize$,
        no. iterations $\niters$, 
        no. crossovers per iteration $\ncrossovers$,
        max. no. few-shots $\maxshots$,
        input \& output token-weights $\weightin$ \& $\weightout$,
        crossover-meta-prompt $\crossoverprompt$,
        mutation-meta-prompt $\mutationprompt$
        \State $\population \gets \HyperCall{initialize\_pop}{\initialinstructions, \fewshotset, \maxshots, \evalllm}{func:initialize_population}$
        \State $\mathcal{F}_\mu \gets \Call{evaluate}{\population, \devset, \evalllm, \weightin, \weightout}$
        \For{$\iter=1$ to $\niters$}
            \State $\offspringpromptset \gets \HyperCall{crossover}{\population, \mathcal{F}_\mu, \metallm, \crossoverprompt, \ncrossovers}{func:cross_over}$
            \State $\offspringpromptset \gets \HyperCall{mutate}{\offspringpromptset, \metallm, \evalllm, \mutationprompt, \fewshotset, \maxshots}{func:mutate}$
            \State $\mathcal{F}_\mu \gets \Call{evaluate}{\offspringpromptset, \devset, \evalllm, \weightin, \weightout}$
            \State $\population \leftarrow \Call{environmental\_selection}{\population \cup \offspringpromptset, \mathcal{F}_\mu, \populationsize}$
        \EndFor
        \State \Return $\population$
    \end{algorithmic}
    \label{algo:nsga-ii}
\end{algorithm}

\subsubsection{\mocapointensify{} Algorithm Details}

\paragraph{Boundary Cases}
For readability and conciseness when introducing the \mocapointensify{} algorithm, we omitted the handling of boundary cases during the optimization process in \S\ref{sec:mocapo}. For completeness' sake, we include them here:

\begin{description}
    \item[Parent selection:] When both tournaments yield the same winner, new tournaments are started until a second distinct parent is yielded.
    \item[Intensification:] In the rare case that a challenger already exists in $\incset$, its intensification is skipped to avoid duplicates. If it exists in $\population$ but not in $\incset$, it is reactivated and intensified again, starting from its cached evaluations in $\runhistory$.  
\end{description}

\paragraph{Function Specifications}
Below, we provide detailed pseudo-code for the functions of the \mocapointensify{} algorithm. For further descriptions, see \S\ref{app:nsgaii}.

\begin{algorithm}[H]
    \small
    \caption{\mocapointensify{} \& \mocaponsgaII{}: Initialization (from \citet{zehle-automl25a})}
    \begin{algorithmic}[1]
        \hypertarget{func:initialize_population}{}
        \Function{initialize\_pop}{$\initialinstructions, \fewshotset, \maxshots, \evalllm$}
            \For{$\instruction{} \in \initialinstructions$}
                \State $\nshots \sim \text{Unif}(\{0, \dots, \maxshots\})$ \Comment{Sample number of few-shots}
                \State $\fewshots{} \gets \Call{create\_shots}{\fewshotset, \nshots, \instruction{},\evalllm}$ \Comment{Create few-shots (cf. \citet{zehle-automl25a})}
                \State $\prompt{} \gets (\instruction{}, \fewshots{})$
                \State $\population \gets \Call{append}{\prompt{}, \population}$
            \EndFor
            \State \Return $\population$
        \EndFunction
    \end{algorithmic}
    \label{algo:initialization}
\end{algorithm}

\begin{algorithm}[H]
    \small
    \caption{\mocapointensify{} \& \mocaponsgaII{}: Evolutionary Operators (adapted from \citet{zehle-automl25a})}
    \begin{algorithmic}[1]
        \hypertarget{func:cross_over}{}
        \Function{crossover}{$\population, \runhistory, \metallm, \crossoverprompt, \ncrossovers$}
            \State $\offspringpromptset \gets  []$
            \For{$j=1$ to $\ncrossovers$}
                \State $\prompt{a}, \prompt{b} \gets \HyperCall{tournament\_selection}{\population, \runhistory}{func:tournament_selection}$
                \Comment{$\prompt{a} = (\instruction{a}, \fewshots{a})$, $\prompt{b} = (\instruction{b}, \fewshots{b})$}
                \State $\offspringinstruction \gets \metallm(\crossoverprompt ||\instruction{a} || \instruction{b})$ \Comment{Let meta-LLM cross the parent instructions}
                \State $\offspringfewshots \gets \Call{sample}{\fewshots{a} \cup \fewshots{b}, \left\lfloor\frac{|\fewshots{a}|+|\fewshots{b}|}{2}\right\rfloor}$ \Comment{Sample offspring shots from parents' shots}
                \State $\offspringprompt \gets (\offspringinstruction, \offspringfewshots)$
                \State $\offspringpromptset \gets \Call{append}{\offspringprompt, \offspringpromptset}$
            \EndFor
            \State \Return $\offspringpromptset$
        \EndFunction
        \Statex
        \hypertarget{func:mutate}{}
        \Function{mutate}{$\offspringpromptset, \metallm, \evalllm, \mutationprompt, \fewshotset, \maxshots$}
            \State $\mutatedpromptset \gets []$
            \For{$\offspringprompt \in \offspringpromptset$}
                \State $\mutatedinstruction \gets \metallm(\mutationprompt \,\|\, \offspringinstruction)$ \Comment{Let meta-LLM mutate the instruction}
                \State $r \sim \text{Unif}(\{0, 1, 2\})$ \Comment{Select randomly one of three cases}
                \If{$r = 0 \ \wedge \ |\offspringfewshots| < \maxshots$} \Comment{Case 1: Create a new few-shot example}
                    \State $\fewshots{\text{new}} \gets \offspringfewshots \cup \Call{create\_shots}{\fewshotset, 1, \mutatedinstruction, \evalllm}$
                \ElsIf{$r = 1 \ \wedge \ |\offspringfewshots|>0$} \Comment{Case 2: Remove a few-shot example}
                    \State $\fewshots{\text{new}} \gets \Call{sample}{\offspringfewshots, |\offspringfewshots|-1}$
                \EndIf
                \Statex \Comment{Case 3: Keep no. few-shot examples unchanged}
                \State $\mutatedprompt \gets \bigl(\,\mutatedinstruction,\; \Call{shuffle}{\fewshots{\text{new}}}\bigr)$ \Comment{In each case: shuffle example order}
                \State $\mutatedpromptset \gets \Call{append}{\mutatedprompt, \mutatedpromptset}$
            \EndFor
            \State \Return $\mutatedpromptset$
        \EndFunction    
    \end{algorithmic}
    \label{algo:evol-operators}
\end{algorithm}

\begin{algorithm}[H]
\caption{\mocapointensify{}: Tournament Selection}
\label{alg:capo_smac}
\small
\begin{algorithmic}[1]
\hypertarget{func:tournament_selection}{}
\Function{tournament\_selection}{$\population, \incset, \runhistory$}
    \For{$i \in \{a,b\}$}
        \State $\prompt{1}, \prompt{2} \gets \Call{sample}{\population, 2}$
        \If{($\prompt{1} \in \incset$) xor ($\prompt{2} \in \incset$)} \Comment{Case (1): only one prompt is incumbent}
            \State $\prompt{i} \gets \{\prompt{1},\prompt{2}\} \cap \incset$  \Comment{Incumbent is selected}
        \ElsIf{$\prompt{1}, \prompt{2} \in \incset$} \Comment{Case (2): both prompts are incumbents}
            \State $\prompt{i} \gets \argmax_{\prompt{} \in \{\prompt{1}, \prompt{2}\}} \Call{cd}{\prompt{}}$ \Comment{Select prompt with larger CD}
        \Else 
            \State $\blockset_1,\blockset_2 \gets \Call{get\_evaluated\_blocks}{\prompt{1},\runhistory},\Call{get\_evaluated\_blocks}{\prompt{2}, \runhistory}$
            \If{$\blockset_1 = \blockset_2$}\Comment{Case (3): non-incumbents with same evaluation level}
                \State $\prompt{i} \gets \Call{get\_best\_by\_nds\_cd}{\{\prompt{1}, \prompt{2}\},\runhistory}$
                \Comment{Select by better front rank / larger CD}
            \ElsIf{$\blockset_1 \subset \blockset_2 \land f(\prompt{2}; \blockset_1) \prec f(\prompt{1};\blockset_1)$} \Comment{Case (4): weaker dominance}
                \State $\prompt{i} \gets \prompt{2}$
            \ElsIf{$\blockset_2 \subset \blockset_1\land f(\prompt{1}; \blockset_2) \prec f(\prompt{2};\blockset_2)$} \Comment{Case (4): weaker dominance}
                \State $\prompt{i} \gets \prompt{1}$
            \Else \Comment{Case (5): random selection}
                \State $\prompt{i} \gets \Call{Sample}{\prompt{1}, \prompt{2}}$ 
            \EndIf
        \EndIf
    \EndFor
    \State \Return $\prompt{a}, \prompt{b}$
\EndFunction
\end{algorithmic}
\label{algo:tournament_inten}
\end{algorithm}

\begin{algorithm}[ht]
\small
\caption{\mocapointensify{}: Advance Incumbents}
\begin{algorithmic}[1]
\hypertarget{func:mocapo_advinc}{}
\Function{advance\_incumbents}{$\incset, \blockset, \runhistory, \evalllm, \weightin, \weightout$}
    \State $\incumbent \gets \argmin_{\prompt{} \in \incset} \left| \Call{get\_evaluated\_blocks}{\prompt{}, \runhistory} \right|$ \Comment{Least evaluated incumbent}
    \State $\blocksmin \gets \Call{get\_evaluated\_blocks}{\incumbent, \runhistory}$

    \If{$(\forall \prompt{i} \in \incset : \Call{get\_evaluated\_blocks}{\prompt{i}, \runhistory} = \blocksmin)$} \Comment{Incumbents share same block sets}
        \State $\blockstoeval \gets \blockset \setminus \blocksmin$  \Comment{Expand: select previously unseen blocks}
    \Else \Comment{$\incumbent$ lags behind}
        \State $\blockstoeval \gets \bigcup_{\prompt{} \in \incset}
        \Call{get\_evaluated\_blocks}{\prompt{}, \runhistory} \setminus \blocksmin$ \Comment{Catch up: select missing blocks}
    \EndIf
    \If{$\blockstoeval \neq \emptyset$}
        \State $\block{} \gets \Call{sample}{\blockstoeval}$ \Comment{Sample block from selection}
        \State $\runhistory \gets \Call{evaluate}{\incumbent, \block{}, \evalllm, \weightin, \weightout}$ \Comment{Advance $\incumbent$ on this block}
    \EndIf
    \State \Return $\runhistory$
\EndFunction
\end{algorithmic}
\end{algorithm}

\begin{algorithm}[ht]
    \small
    \caption{\mocapointensify{}: Environmental Selection}
    \begin{algorithmic}[1]
        \hypertarget{func:mocapo_envselection}{}
        \Function{environmental\_selection}{$\population, \incset, \runhistory, \populationsize$}
            \While{$|\population| > \populationsize$} \Comment{As long as population size is too large}
                \If{$|\incset| < |\population|$} \Comment{If non-incumbents exist}
                \Statex  \Comment{Case 1: all non-incumbents have equal evaluation levels}
                    \If{$\exists \blockset_\text{sub}\subseteq\blockset: \forall \prompt{} \in \population \setminus \incset, \Call{get\_evaluated\_blocks}{\prompt{}, \runhistory} = \blockset_\text{sub}$}
                        \State $\prompt{\text{worst}} \gets \Call{get\_worst\_by\_nds\_cd}{\population \setminus \incset, \runhistory}$ \Comment{Remove by front rank/CD}
                    \Else \Comment{Case 2: non-incumbents have heterogeneous evaluation levels}
                        \State $\prompt{\text{worst}} \gets \Call{sample}{\argmin_{\prompt{} \in \population \setminus \incset}\Call{get\_evaluated\_blocks}{\prompt{}, \runhistory}}$ \Comment{Eliminate randomly}
                    \EndIf
                \Else \Comment{If all remaining candidates are incumbents}
                    \State $\prompt{\text{worst}} \gets \Call{get\_worst\_by\_cd}{\incset, \runhistory}$ \Comment{Eliminate by lowest CD}
                    \State $\incset \gets \incset \setminus \{\prompt{\text{worst}}\}$
                \EndIf
                \State $\population \gets \population \setminus \{\prompt{\text{worst}}\}$
            \EndWhile
            \State \Return $\population, \incset$
        \EndFunction
    \end{algorithmic}
    \label{algo:env-selection}
\end{algorithm}

\newpage
\subsection{Experiment Details}\label{app:experiment-details}

\subsubsection{Model Details}\label{app:model-details}

We report detailed IDs and revisions of the utilized LLMs from HuggingFace in Table~\ref{tab:model-details}. To locally host the LLMs, we use vLLM 0.13.0~\citep{kwon-sosp23a} as a fast and easy-to-use library for LLM inference and serving since it efficiently manages the required memory and allows the usage of quantized models. Note that we restrict the maximum output length to 3000, which is long enough for almost all generations while still allowing for reasonably large batch sizes. The optimal batch size is chosen by vLLM depending on available memory.

\begin{table}[H]
    \centering
    \caption{Overview of the utilized LLMs.}
    \footnotesize
    \begin{tabularx}{\textwidth}{@{}p{0.175\textwidth}p{0.375\textwidth}p{0.45\textwidth}@{}}
        \toprule
        \textbf{Model} & \textbf{Huggingface ID} & \textbf{Revision} \\
        \midrule
        \textbf{\mistral} & \href{https://huggingface.co/mistralai/Mistral-Small-3.2-24B-Instruct-2506}{mistralai/Mistral-Small-3.2-24B-Instruct-2506} & 95a6d26c4bfb886c58daf9d3f7332c857cb27b43 \\
        \textbf{\qwen} & \href{https://huggingface.co/Qwen/Qwen3-30B-A3B-Instruct-2507}{Qwen/Qwen3-30B-A3B-Instruct-2507} & 0d7cf23991f47feeb3a57ecb4c9cee8ea4a17bfe \\
       \textbf{\gpt} & \href{https://huggingface.co/openai/gpt-oss-120b}{openai/gpt-oss-120b} & b5c939de8f754692c1647ca79fbf85e8c1e70f8a \\
        \bottomrule
    \end{tabularx}
    \label{tab:model-details}
\end{table}

\subsubsection{Dataset Details}\label{app:dataset-details}

In our experiments, we utilize four datasets, all retrieved from HuggingFace:

\begin{enumerate}[label=(\arabic*), itemsep=0pt, parsep=0pt, topsep=0pt, partopsep=0pt]
    \item AG News~\citep{zhang-neurips15a}: topic classification dataset with titles and descriptions of news articles that are to be assigned to either \textit{World, Sports, Business} or \textit{Sci/Tech}. The input $x$ is taken from the column ``text'', the labels $y$ from the column ``label\_text''. Performance is evaluated using accuracy.
    \item Subj~\citep{pang-acl04a}: subjectivity classification dataset with movie reviews that are to be classified as either \textit{subjective} or \textit{objective}. The input $x$ is taken from the column ``text'', the labels $y$ from the column ``label\_text''. Performance is evaluated using accuracy.
    \item GSM8K~\citep{cobbe-arxiv21a}: grade school math word problems requiring multi-step reasoning. We utilize the train and test split of the ``main'' subset, from which the column ``question'' is used as input $x$, the label $y$ is extracted from the ``answer'' after \texttt{\#\#\#\#}. Performance is evaluated using accuracy of exactly correct solutions.
    \item MBPP~\citep{austin-arxiv21}:  crowd-sourced Python programming problems, designed to be solvable by entry-level programmers, covering programming fundamentals and standard library functionality. Each problem consists of a ``task description'' used as input $x$, a code solution used as label $y$, and 3 automated test cases. Performance is measured by a reward function that runs the test cases associated with each instance in MBPP on the generated code and returns the fraction of passed tests. Note that this reward function does not require a label $y$, which is only used for few-shot example creation.
\end{enumerate}

We provide detailed IDs and revisions of the utilized datasets in Table~\ref{tab:dataset-details}. For $\fewshotset$ and $\devset$, 400 instances are sampled from the train split without replacement. The first 300 points are used for $\devset$, the remaining 100 for $\fewshotset$. For $\testset$, 500 instances are sampled from the test split without replacement. These sampling procedures are seeded using the random seed of the corresponding experiment run.

\begin{table}[H]
    \centering
    \caption{Overview of the utilized HuggingFace datasets.}
    \footnotesize
    \begin{tabularx}{\textwidth}{@{}p{0.08\textwidth}p{0.26\textwidth}p{0.38\textwidth}p{0.04\textwidth}p{0.04\textwidth}p{0.04\textwidth}@{}}
        \toprule
        \textbf{Dataset} & \textbf{Huggingface ID} & \textbf{Revision} & $\mathbf{n_\text{train}}$ & $\mathbf{n_\text{test}}$ & \textbf{\#classes} \\
        \midrule
         \textbf{AG News} & \href{https://huggingface.co/datasets/SetFit/ag_news}{SetFit/ag\_news} & ca5ba619eb034211db5-f70932b6702efd21e7c73 & 120k & 7.6k & 4\\
         \textbf{Subj} & \href{https://huggingface.co/datasets/SetFit/subj}{SetFit/subj} & f3c1162e678417f664d-76b21864fdb87b0615fcf & 8.0k & 2.0k & 2\\
         \textbf{GSM8K} & \href{https://huggingface.co/datasets/openai/gsm8k}{openai/gsm8k} & e53f048856ff4f594e95-9d75785d2c2d37b678ee & 7.5k & 1.3k & -\\
         \textbf{MBPP} & \href{https://huggingface.co/datasets/google-research-datasets/mbpp}{google-research-datasets/mbpp} & 4bb6404fdc6cacfda99d4ac4205087b89d32030c & 474 & 500 & -\\
         \bottomrule
    \end{tabularx}
    \label{tab:dataset-details}
\end{table}

\subsubsection{Hardware Details}\label{app:hardware-details}

All computations are performed on a GPU cluster. For each experiment configuration, only a single GPU with at least 40GB of RAM (NVIDIA A100 or NVIDIA H100) is used to host the corresponding LLM. Experiments are distributed across multiple instances for parallel execution. %

\subsubsection{Implementation Details}\label{app:implementation-details}

\paragraph{Answer Extraction}
To reliably extract information from LLM output in our experiments, we utilize marker-based extraction. Concretely, we parse the information in html-style tags: offspring/mutated prompts are extracted between \texttt{<prompt></prompt>} markers and predictions between \texttt{<final\_answer></final\_answer>} markers in the LLM output. This information is also included in the initial instructions and task descriptions.

\paragraph{Optimizer Parametrization}
For our experiments, we use the following default hyperparameters:
We parametrize our \mocapointensify{} algorithm with $\ncrossovers=4$, $\maxshots=5$ and $\blocksize=30$. By passing $10$ sampled initial prompts in the beginning, we fix $\populationsize = 10$.
For \mocaponsgaII{} we use the same $\ncrossovers$, $\populationsize$, and $\maxshots$, while the blocksize $\blocksize$ is not required. The model-specific input- and output token weights used in the cost objective of \mocapointensify{} and \mocaponsgaII{} are those from Tab.~\ref{tab:cost-weights}.

For CAPO~\citep{zehle-automl25a}, we use the same $\ncrossovers$, $\populationsize$, $\maxshots$, and $\blocksize$. Additionally, we set $\alpha = 0.2$ and $\maxblockevals=10$ (i.e., $\blocksize \cdot \maxblockevals = |\devset|$), and $\lengthpenalty=0.05$ (a prompt with the same length as the longest initial prompt (instruction + examples) is penalized by 5\%p) as in the original paper.

For EvoPromptGA~\citep{guo-iclr24a}, we also use the same $\populationsize$, following the recommendations of the original paper. For GEPA~\citep{agrawal-arxiv25a}, we use the original parametrization and provide one randomly sampled instruction from our pool.
For all optimizers, an iteration limit of 2000 steps prevents excessively long runs when algorithm convergence prevents budget exhaustion.

\paragraph{Optimizer Implementation}
For CAPO and EvoPromptGA, we use (re-)implementations that are available in the public prompt optimization library \texttt{promptolution}~\citep{zehle-arxiv25a}. For GEPA, we wrap their original implementation\footnote{\url{https://github.com/gepa-ai/gepa} (accessed: 2026-02-19)} with small adaptations to ensure we can stop the optimization loop within our budget constraints.

\paragraph{Budget Computation}  We compute the input token budget usage by applying each LLM's respective tokenizer and counting the resulting number of tokens. Similarly, the input and output costs are computed based on the number of tokens produced by the respective tokenizer.

\subsection{Input Specifications and Templates}

\subsubsection{Task Descriptions}\label{app:task-descriptions}
~\\

\begin{table}[H]
    \centering
    \caption{Manually created task descriptions used for CAPO, \mocapointensify{} and \mocaponsgaII{}}
    \footnotesize
    \begin{tabularx}{\textwidth}{@{}p{\textwidth}@{}}
        \toprule
    \textbf{AG News}: \\ The dataset contains news articles categorized into four classes: World, Sports, Business, and Sci/Tech. The task is to classify each news article into one of the four categories. The class will be extracted between the markers \verb|<final_answer>|answer\verb|</final_answer>|.\\
    \midrule
    \textbf{Subj}: \\ The dataset contains sentences labeled as either subjective or objective. The task is to classify each sentence as either subjective or objective. The class will be extracted between the markers \verb|<final_answer>|answer\verb|</final_answer>|.\\
    \midrule
    \textbf{GSM8K}: \\ The dataset consists of grade school math word problems that require multi-step reasoning to solve. The task is to solve each word problem and provide the final answer. The final solution will be extracted between the markers \verb|<final_answer>|answer\verb|</final_answer>|.\\
    \midrule
    \textbf{MBPP}: \\ The dataset consists of Python programming problems described in natural language. The task is to generate a Python function that correctly solves the given problem. The solution code will be extracted between the markers \verb|<final_answer>|code\verb|</final_answer>|.\\
    \bottomrule
    \end{tabularx}
    \label{tab:task-descriptions}
\end{table}

\subsubsection{Initial Instructions}\label{app:init-instructions}

Since all employed optimizers require initial instructions to start from, we create a set of 15 initial instructions for each task. To demonstrate that this requirement of initial instructions is not a major limiting factor of the algorithms, we produce them in an automated manner, prompting Anthropic's Claude Sonnet (\url{https://claude.ai/}) to create a diverse set of initial instructions, making use of our task descriptions in Appendix~\ref{app:task-descriptions}. The full prompt template is provided in Table~\ref{tab:init-instr-creation}.

\begin{table}[h]
    \centering
    \caption{Prompt used to generate initial instructions with Anthropic's Claude Sonnet. The <task\_description> placeholder is replaced with our task description.}
    \footnotesize
    \begin{tabularx}{\textwidth}{@{}p{\textwidth}@{}}
        \toprule
        Please create diverse prompts for the following task. They should be linguistically diverse (but always in English) and have varying lengths and complexities. This means some consist only of a short sentence with a rather high-level description while others elaborate on the task in little more detail. \\ Task: \verb|<task_description>| \\ Explicitly state this expected format as part of the prompts. Create overall 15 prompts within quotes as an array: \\
        \bottomrule
    \end{tabularx}
    \label{tab:init-instr-creation}
\end{table}

\subsubsection{Meta-Prompt Templates}\label{app:prompt-templates}
~\\

\begin{table}[h]
    \centering
    \caption{List of all meta-prompt templates used in CAPO, \mocapointensify{} and \mocaponsgaII{}.}
    \footnotesize
    \begin{tabularx}{\textwidth}{@{}p{\textwidth}@{}}
        \toprule
        \textbf{CAPO cross-over meta-prompt template}: \\ You receive two prompts for the following task: \textcolor{EvoPurple}{\verb|<task_description>|} \\ Please merge the two prompts into a single coherent prompt. Maintain the key linguistic features from both original prompts: \\ Prompt 1: \textcolor{EvoPurple}{\verb|<mother>|} \\ Prompt 2: \textcolor{EvoPurple}{\verb|<father>|} \\ \\ Return the new prompt in the following format: \\ {\verb|<prompt>|}new prompt{\verb|</prompt>|}.\\
        \midrule
        \textbf{CAPO mutation meta-prompt template}: \\ You receive a prompt for the following task: \textcolor{EvoPurple}{\verb|<task_description>|} \\ Please rephrase the prompt, preserving its core meaning while substantially varying the linguistic style. \\ Prompt: \textcolor{EvoPurple}{\verb|<instruction>|} \\ \\ Return the new prompt in the following format: \\ {\verb|<prompt>|}new prompt {\verb|</prompt>|} \\
        \bottomrule
    \end{tabularx}
    \label{tab:meta-prompt-templates}
\end{table}

\newpage
\subsection{Token Costs}
\label{app:pricing}
To parameterize the input- and output costs of our utilized models we average the according prices from different vendors, when using the models via an API interface. All prices are taken from OpenRouter.\footnote{https://openrouter.ai/ (accessed: 2026-01-19, 2026-03-01)}

\begin{table}[H]
    \centering
    \footnotesize
    \setlength{\tabcolsep}{4pt}
    \renewcommand{\arraystretch}{1}
    \caption{\mistral{} pricing per million tokens across 
    vendors on OpenRouter (as of January 19, 2026). Prices in USD.}
    \vspace{-10pt}
    \begin{tabular}{lcc}
        \toprule
        \textbf{Vendor} & \textbf{Input Token Price} & \textbf{Output Token Price} \\
        \midrule
        Chutes & \$0.06 & \$0.18 \\
        DeepInfra & \$0.075 & \$0.20 \\
        Parasail & \$0.09 & \$0.60 \\
        Mistral & \$0.10 & \$0.30 \\
        \midrule
        \textbf{Average} & \textbf{\$0.08} & \textbf{\$0.32} \\
        \bottomrule
    \end{tabular}
\end{table}

\begin{table}[H]
    \centering
    \footnotesize
    \caption{\qwen{} pricing per million tokens across 
    vendors on OpenRouter (as of January 19, 2026). Prices in USD.}
    \vspace{-10pt}
    \begin{tabular}{lcc}
        \toprule
        \textbf{Vendor} & \textbf{Input Token Price} & \textbf{Output Token Price} \\
        \midrule
        Chutes & \$0.08 & \$0.33 \\
        AtlasCloud & \$0.09 & \$0.30 \\
        SiliconFlow & \$0.09 & \$0.30 \\
        Nebius Token Factory & \$0.10 & \$0.30 \\
        Alibaba Cloud Int. & \$0.20 & \$0.80 \\
        \midrule
        \textbf{Average} & \textbf{\$0.11} & \textbf{\$0.41} \\
        \bottomrule
    \end{tabular}
\end{table}

\begin{table}[H]
    \centering
    \footnotesize
    \setlength{\tabcolsep}{4pt}
    \renewcommand{\arraystretch}{1}
    \caption{\gpt{} pricing per million tokens across 
    vendors on OpenRouter (as of January 19, 2026). Prices in USD.}
    \vspace{-10pt}
    \begin{tabular}{lcc}
        \toprule
        \textbf{Vendor} & \textbf{Input Token Price} & \textbf{Output Token Price} \\
        \midrule
        DeepInfra & \$0.039 & \$0.19 \\
        Chutes & \$0.04 & \$0.18 \\
        GMICloud & \$0.05 & \$0.25 \\
        NovitaAI & \$0.05 & \$0.25 \\
        nCompass & \$0.05 & \$0.28 \\
        SiliconFlow & \$0.05 & \$0.45 \\
        Google Vertex & \$0.09 & \$0.36 \\
        Clarifai & \$0.09 & \$0.36 \\
        AtlasCloud & \$0.10 & \$0.20 \\
        Phala & \$0.10 & \$0.49 \\
        Baseten & \$0.10 & \$0.50 \\
        Parasail & \$0.12 & \$0.75 \\
        SambaNova & \$0.14 & \$0.95 \\
        Fireworks & \$0.15 & \$0.60 \\
        Amazon Bedrock & \$0.15 & \$0.60 \\
        Together & \$0.15 & \$0.60 \\
        Nebius Token Factory & \$0.15 & \$0.60 \\
        Weights \& Biases & \$0.15 & \$0.60 \\
        Groq & \$0.15 & \$0.60 \\
        Crusoe & \$0.15 & \$0.60 \\
        DeepInfra (Turbo) & \$0.15 & \$0.60 \\
        Cerebras & \$0.35 & \$0.75 \\
        \midrule
        \textbf{Average} & \textbf{\$0.12} & \textbf{\$0.49} \\
        \bottomrule
    \end{tabular}
\end{table}

\begin{table}[H]
    \centering
    \footnotesize
    \caption{Claude Opus 4.6 pricing per million tokens across 
    vendors on OpenRouter (as of March 1, 2026). Prices in USD.}
    \vspace{-10pt}
    \begin{tabular}{lcc}
        \toprule
        \textbf{Vendor} & \textbf{Input Token Price} & \textbf{Output Token Price} \\
        \midrule
        Amazon Bedrock & \$2.42 & \$25.47 \\
        Google & \$2.85 & \$25.21 \\
        Anthropic & \$2.88 & \$25.62 \\
        \midrule
        \textbf{Average} & \textbf{\$2.72} & \textbf{\$25.43} \\
        \bottomrule
    \end{tabular}
\end{table}

\begin{table}[H]
    \centering
    \footnotesize
    \caption{GPT-5.2 Pro pricing per million tokens across 
    vendors on OpenRouter (as of March 1, 2026). Prices in USD.}
    \vspace{-10pt}
    \begin{tabular}{lcc}
        \toprule
        \textbf{Vendor} & \textbf{Input Token Price} & \textbf{Output Token Price} \\
        \midrule
        OpenAI & \$21.00 & \$168.00 \\
        \midrule
        \textbf{Average} & \textbf{\$21.00} & \textbf{\$168.00} \\
        \bottomrule
    \end{tabular}
\end{table}

\subsection{Further Ablation Studies}\label{app:further-ablations}

With the ablation configuration $w_{\text{in}}=0, w_{\text{out}}=0$, MO-CAPO effectively reduces to a single-objective optimizer. Table \ref{tab:ablation-so-acc} compares its performance to other single-objective prompt optimizers. This setting provides a direct comparison between the intensification procedure used in \mocapointensify{} and the racing mechanism employed by CAPO. As shown in the table, both approaches achieve very similar overall performance, outperforming the other optimizers and  the initial instructions.

\begin{table}[h]
\centering
\caption{Accuracy achieved by all single-objective optimizers at a budget of 7.5M tokens (bold = best accuracy per dataset).  The best prompt is selected based on development performance and evaluated on the test set. All experiments use \qwen.}
\label{tab:ablation-so-acc}
\begin{tabular}{lcc}
\toprule
 & \textbf{GSM8K} & \textbf{Subj} \\
\midrule
Initial & $0.540_{\pm.022}$ & $0.754_{\pm.044}$ \\
GEPA & $0.659_{\pm.024}$ & $0.775_{\pm.078}$ \\
EvoPromptGA & $0.537_{\pm.010}$ & $0.783_{\pm.019}$ \\
CAPO & $0.665_{\pm.025}$ & $\mathbf{0.909}_{\pm.033}$ \\
\mocapointensify{} ($w_{\text{in}}=0, w_{\text{out}}=0$) & $\mathbf{0.667}_{\pm.020}$ & $0.905_{\pm.012}$ \\
\bottomrule
\end{tabular}
\end{table}

Another minor design choice in \mocapointensify{} is the parent selection via binary tournament with a weaker dominance criterion compared to CAPO's random selection.
We ablate the parent selection mechanism by comparing \mocapointensify's tournament-based strategy against a variant without tournament selection (random selection instead) and a variant without the weaker dominance criterion in the tournament.
These ablations are conducted on GSM8K and Subj wih \qwen{} and a budget of 7.5M tokens.
The results in table~\ref{tab:ablation} do not show any substantial performance differences compared to \mocapointensify{}. On GSM8K, all variants achieve nearly identical nR2 values (0.338–0.340) and very similar HV scores. On Subj, differences are likewise minor: while the variant without weaker dominance attains slightly higher optimistic and pessimistic HV, the default configuration achieves the smallest approximation gap and comparable nR2.

\begin{table}[H]
\centering
\caption{Ablation study results for \mocapointensify{} on \qwen{} at 7.5M token budget.
nR2 (lower is better).
HV$_{opt}$ (optimistic) and HV$_{pes}$ (pessimistic) measure Pareto front quality (higher is better).
Gap measures HV uncertainty (lower is better).
Bold indicates the best value per metric and dataset.}
\label{tab:ablation}
\vspace{-0.25cm}
\begin{adjustbox}{max width=\textwidth}
\setlength{\tabcolsep}{2pt}
\renewcommand{\arraystretch}{0.9}
\begin{tabular}{l cccc cccc}
\toprule
 & \multicolumn{4}{c}{\textbf{GSM8K}} & \multicolumn{4}{c}{\textbf{Subj}} \\
\cmidrule(lr){2-5} \cmidrule(lr){6-9}
\textbf{Parametrization} & \textbf{nR2} $\downarrow$ & \textbf{HV}$_{opt}$ $\uparrow$ & \textbf{HV}$_{pes}$ $\uparrow$ & \textbf{Gap} $\downarrow$ & \textbf{nR2} $\downarrow$ & \textbf{HV}$_{opt}$ $\uparrow$ & \textbf{HV}$_{pes}$ $\uparrow$ & \textbf{Gap} $\downarrow$ \\
\midrule
\mocapointensify{} (default) & $0.340_{\pm.006}$ & $\mathbf{0.410}_{\pm.018}$ & $0.406_{\pm.021}$ & $0.004_{\pm.004}$ & $\mathbf{0.106}_{\pm.005}$ & $0.930_{\pm.015}$ & $0.925_{\pm.022}$ & $\mathbf{0.005}_{\pm.006}$ \\
w/o tournament & $\mathbf{0.338}_{\pm.005}$ & $0.409_{\pm.011}$ & $\mathbf{0.408}_{\pm.014}$ & $\mathbf{0.002}_{\pm.003}$ & $0.112_{\pm.020}$ & $0.922_{\pm.063}$ & $0.914_{\pm.051}$ & $0.009_{\pm.014}$ \\
w/o weaker dom. & $0.339_{\pm.004}$ & $0.409_{\pm.011}$ & $0.407_{\pm.007}$ & $0.002_{\pm.004}$ & $0.106_{\pm.015}$ & $\mathbf{0.954}_{\pm.058}$ & $\mathbf{0.939}_{\pm.060}$ & $0.016_{\pm.026}$ \\
\bottomrule
\end{tabular}
\end{adjustbox}
\end{table}

\subsection{Hyperparameter Analysis}\label{app:hp}

We investigate the sensitivity of the hyperparameters in \mocapointensify{} on GSM8K and Subj with \qwen{} and a budget of 7.5M tokens. Results are reported in Table~\ref{tab:hp}.

Reducing the block size from the default $b{=}30$ to $b{=}15$ yields a marginal improvement in nR2 and a smaller approximation gap on GSM8K, while slightly degrading performance on Subj. However, all differences are within one standard deviation, indicating no meaningful impact.

Varying the population size around the default $\mu{=}10$ shows a weak positive trend with larger populations. Increasing to $\mu{=}12$ produces the best mean results on GSM8K across all metrics and modest gains on Subj, whereas $\mu{=}8$ slightly underperforms. The improvements are small, suggesting limited sensitivity within this range. One might expect MOO to require a larger population size than single-objective optimization to explore the increased number of objectives. However, we are not aware of any studies on this, leaving studies of even larger population sizes for future work. 

Increasing the cross-over rate from $c{=}4$ to $c{=}10$ has a negligible effect on GSM8K but yields small improvements on Subj in nR2 and HV metrics, at the cost of a slightly larger gap.

Overall, the method appears robust to moderate variations in hyperparameters. No single change consistently improves performance across both datasets, and most differences are small relative to the observed variability, supporting the adequacy of the default configuration.

\begin{table}[h]
\centering
\caption{Hyperparameter analysis for \mocapointensify{} on GSM8K and Subj using \qwen{} with a 7.5M token budget. All other hyperparameters are kept default. We report nR2 (lower is better), HV$_{opt}$, HV$_{pes}$ (higher is better), and the gap between those values (lower is better). Bold indicates best value within each subgroup.}
\label{tab:hp}
\vspace{-0.25cm}
\begin{adjustbox}{max width=\textwidth}
\setlength{\tabcolsep}{2pt}
\renewcommand{\arraystretch}{0.9}
\begin{tabular}{l cccc cccc}
\toprule
 & \multicolumn{4}{c}{\textbf{GSM8K}} & \multicolumn{4}{c}{\textbf{Subj}} \\
\cmidrule(lr){2-5} \cmidrule(lr){6-9}
\textbf{Parametrization} & \textbf{nR2} $\downarrow$ & \textbf{HV}$_{opt}$ $\uparrow$ & \textbf{HV}$_{pes}$ $\uparrow$ & \textbf{Gap} $\downarrow$ & \textbf{nR2} $\downarrow$ & \textbf{HV}$_{opt}$ $\uparrow$ & \textbf{HV}$_{pes}$ $\uparrow$ & \textbf{Gap} $\downarrow$ \\
\midrule
$b$=15 & $\mathbf{0.337}_{\pm.003}$ & $0.403_{\pm.016}$ & $0.401_{\pm.017}$ & $\mathbf{0.002}_{\pm.003}$ & $0.126_{\pm.022}$ & $0.882_{\pm.055}$ & $0.878_{\pm.051}$ & $\mathbf{0.003}_{\pm.006}$ \\
$b$=30 (default) & $0.340_{\pm.006}$ & $\mathbf{0.410}_{\pm.018}$ & $\mathbf{0.406}_{\pm.021}$ & $0.004_{\pm.004}$ & $\mathbf{0.106}_{\pm.005}$ & $\mathbf{0.930}_{\pm.015}$ & $\mathbf{0.925}_{\pm.022}$ & $0.005_{\pm.006}$ \\
\midrule
$\mu$=8 & $0.342_{\pm.008}$ & $0.403_{\pm.012}$ & $0.401_{\pm.007}$ & $0.003_{\pm.005}$ & $0.116_{\pm.013}$ & $0.911_{\pm.031}$ & $0.900_{\pm.032}$ & $0.010_{\pm.012}$ \\
$\mu$=10 (default) & $0.340_{\pm.006}$ & $0.410_{\pm.018}$ & $0.406_{\pm.021}$ & $0.004_{\pm.004}$ & $0.106_{\pm.005}$ & $0.930_{\pm.015}$ & $0.925_{\pm.022}$ & $\mathbf{0.005}_{\pm.006}$ \\
$\mu$=12 & $\mathbf{0.335}_{\pm.007}$ & $\mathbf{0.413}_{\pm.002}$ & $\mathbf{0.413}_{\pm.002}$ & $\mathbf{0.000}_{\pm.000}$ & $\mathbf{0.105}_{\pm.022}$ & $\mathbf{0.946}_{\pm.070}$ & $\mathbf{0.934}_{\pm.063}$ & $0.012_{\pm.019}$ \\
\midrule
$c$=4 (default) & $\mathbf{0.340}_{\pm.006}$ & $\mathbf{0.410}_{\pm.018}$ & $0.406_{\pm.021}$ & $0.004_{\pm.004}$ & $0.106_{\pm.005}$ & $0.930_{\pm.015}$ & $0.925_{\pm.022}$ & $\mathbf{0.005}_{\pm.006}$ \\
$c$=10 & $0.341_{\pm.005}$ & $0.408_{\pm.009}$ & $\mathbf{0.407}_{\pm.008}$ & $\mathbf{0.001}_{\pm.001}$ & $\mathbf{0.091}_{\pm.014}$ & $\mathbf{0.986}_{\pm.051}$ & $\mathbf{0.976}_{\pm.050}$ & $0.010_{\pm.006}$ \\
\bottomrule
\end{tabular}
\end{adjustbox}
\end{table}

\clearpage

\subsection{Prompt Examples}\label{app:prompt-examples}

\begin{table*}[h]
\centering
\small
\caption{Prompt examples from the Pareto front obtained with \mocapointensify{} on the Subj dataset using \mistral{}, together with the best single-objective prompt produced by CAPO. Each entry reports the instruction and the first few-shot example of the corresponding prompt (if existent), along with its test accuracy and inference cost. The Pareto front is shown in Fig.~\ref{fig:intro_pareto}, and the labeled solutions correspond to the prompts listed below.}
\begin{tabular}{p{0.98\textwidth}}
\toprule
\textbf{Best Performance} \hfill Test Acc: 89.0\% \hfill Input Costs: 22.69 \hfill Output Costs: 3.93 \\
\cmidrule(lr){1-1}
\textbf{Instruction:} \\
For each sentence provided, determine whether it conveys verifiable information (objective) or reflects personal feelings, beliefs, or interpretations (subjective). Respond with only the appropriate label—`subjective` or `objective`—enclosed within \texttt{<final\_answer>} and \texttt{</final\_answer>} tags. \\

\textbf{Few-Shot Example 1:} \\
Input: however to keep the image of him as a sex guru going he has to get more lessons from sharonna whom he begins to fall for despite using her . \\
Output: \texttt{<final\_answer>objective</final\_answer>} \\

\textit{+2 more examples}\\

\midrule

\textbf{Balanced} \hfill Test Acc: 76.2\% \hfill Input Costs: 10.37 \hfill Output Costs: 3.98 \\
\cmidrule(lr){1-1}

\textbf{Instruction:} \\
Classify the given sentence as either subjective or objective. Provide your response as solely the word `subjective` or `objective`, enclosed within \texttt{<final\_answer>} and \texttt{</final\_answer>} tags. \\

\textbf{Few-Shot Example 1:} \\
Input: the kind of film that grows on you upon reflection \\
Output: \texttt{<final\_answer>subjective</final\_answer>} \\

\midrule

\textbf{Lowest Cost} \hfill Test Acc: 64.0\% \hfill Input Costs: 5.31 \hfill Output Costs: 4.04 \\
\cmidrule(lr){1-1}

\textbf{Instruction:} \\
Determine if this sentence is subjective or objective and put your answer between \texttt{<final\_answer>} tags. \\

\midrule

\textbf{Best SO Prompt} \hfill Test Acc: 87.2\% \hfill Input Costs: 27.43 \hfill Output Costs: 3.93 \\
\cmidrule(lr){1-1}

\textbf{Instruction:} \\
Draw upon your sensitivity to language subtleties to determine whether each provided sentence conveys verifiable, neutral information (objective) or reflects personal perspectives, feelings, evaluations, or subjective viewpoints (subjective). Does the sentence present facts or convey opinion? Respond with your classification enclosed within the tags: \texttt{<final\_answer>your\_answer</final\_answer>}. \\

\textbf{Few-Shot Example 1:} \\
Input: the film serves as a valuable time capsule to remind us of the devastating horror suffered by an entire people . \\
Output: \texttt{<final\_answer>subjective</final\_answer>} \\

\textit{+4 more examples}\\

\bottomrule
\end{tabular}
\label{tab:subjectivity_prompts}
\end{table*}

\begin{table*}[h]
\centering
\small
\caption{Prompt examples from the Pareto front obtained with \mocapointensify{} on MBPP using \qwen{}, showing the highest- and lowest-performing prompts identified during optimization. Each entry reports the instruction, a representative example, test accuracy, and input/output costs. The corresponding Pareto plot is shown in Fig.~\ref{fig:qwen_fs}.}
\begin{tabular}{p{0.98\textwidth}}
\toprule
\textbf{Highest Performance} \hfill Test Acc: 37.9\% \hfill Input Costs: 23.99 \hfill Output Costs: 22.64 \\
\cmidrule(lr){1-1}
\textbf{Instruction:} \\
Examine the natural language specification of a Python coding task. Your objective is to write a concise, correct, and efficient function that fully satisfies all stated conditions. Thoroughly analyze the requirements, account for potential edge cases, and deliver a solution that is both high-performing and easy to understand. Ensure the function is standalone, relies solely on Python's built-in features, and adheres to clean coding standards. Provide a compact, error-free implementation enclosed precisely between the tags \texttt{<final\_answer>} and \texttt{</final\_answer>}. Do not include any additional commentary---only the function code should appear within the markers. \\

\textbf{Few-Shot Example 1:} \\
Input: Write a function to multiply all the numbers in a list and divide with the length of the list. \\
Expected function name is \texttt{multiply\_num}. \\
Output: \texttt{<final\_answer>def multiply\_num(numbers):}\\
\texttt{\ \ \ \ total = 1}\\
\texttt{\ \ \ \ for x in numbers:}\\
\texttt{\ \ \ \ \ \ \ \ total *= x}\\
\texttt{\ \ \ \ return total/len(numbers)</final\_answer>} \\

\midrule

\textbf{Lowest Performance} \hfill Test Acc: 35.3\% \hfill Input Costs: 15.52 \hfill Output Costs: 37.13 \\
\cmidrule(lr){1-1}
\textbf{Instruction:} \\
Here is a natural language description of a Python coding challenge. Your objective is to write a function in Python that accurately fulfills the specified task. After developing your solution, enclose the full code within \texttt{<final\_answer>} and \texttt{</final\_answer>} tags to ensure proper extraction. \\

\textbf{Few-Shot Example 1:} \\
Input: Write a function to return true if the given number is even else return false. \\
Expected function name is \texttt{even\_num}. \\
Output: \texttt{<final\_answer>def even\_num(x):}\\
\texttt{\ \ if x\%2==0:}\\
\texttt{\ \ \ \ \ return True}\\
\texttt{\ \ else:}\\
\texttt{\ \ \ \ return False</final\_answer>} \\

\bottomrule
\end{tabular}
\label{tab:mbpp_qwen_prompt_examples}
\end{table*}

\clearpage
\subsection{Further Results}\label{app:further-results}

\begin{figure}[h]
\centering

\includegraphics[width=0.6\textwidth]{graphics/traj_nR2/legend_wide_reduced.pdf}

\begin{minipage}[t]{0.26\textwidth}\centering
\includegraphics[width=\linewidth]{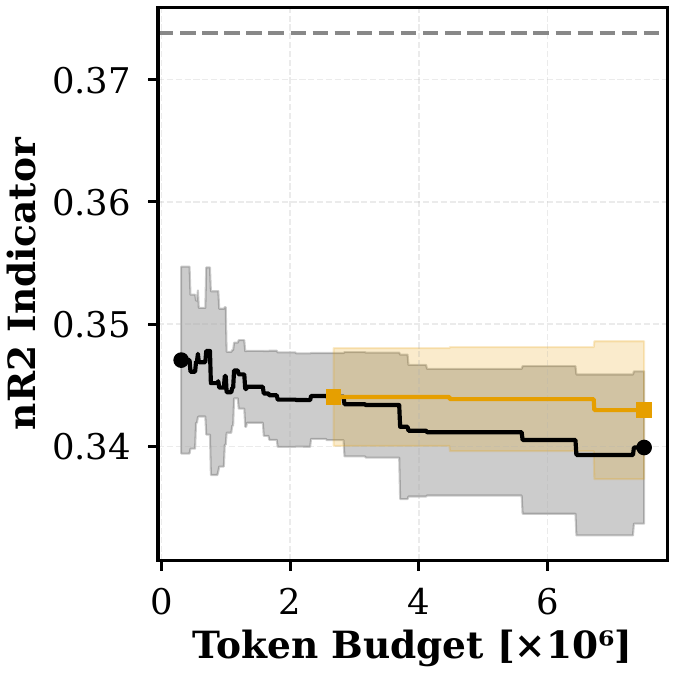}
\Description[Optimization trace for Qwen3-30B on GSM8K.]{Optimization trace for Qwen3-30B on GSM8K for the nR2 indicator.}
\end{minipage}\hfill
\begin{minipage}[t]{0.26\textwidth}\centering
\includegraphics[width=\linewidth]{graphics/traj_nR2/gsm8k_gpt-oss-120b_nR2.pdf}
\Description[Optimization trace for GPT-OSS-120B on GSM8K.]{Optimization trace for GPT-OSS-120B on GSM8K for the nR2 indicator.}
\end{minipage}\hfill
\begin{minipage}[t]{0.26\textwidth}\centering
\includegraphics[width=\linewidth]{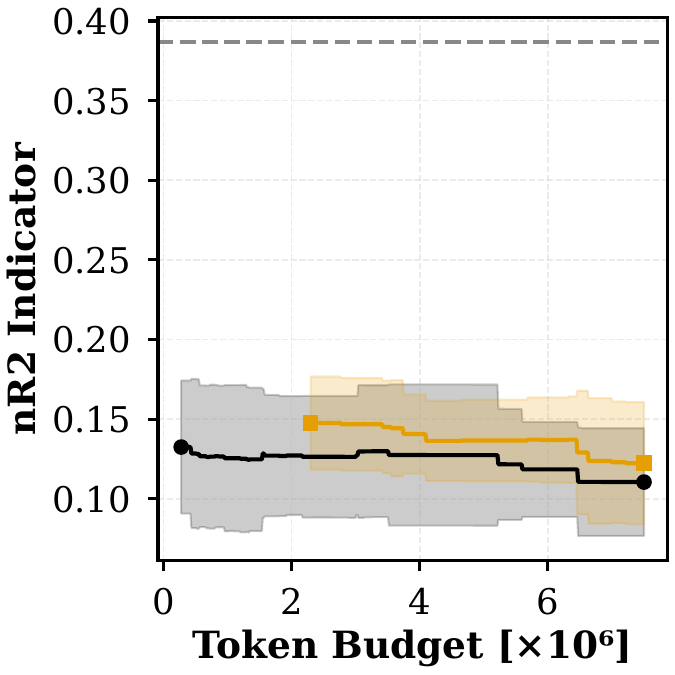}
\Description[Optimization trace for Mistral-3-24B on GSM8K.]{Optimization trace for Qwen3-30B on GSM8K for the nR2 indicator.}
\end{minipage}

\begin{minipage}[t]{0.26\textwidth}\centering\small (a) Qwen-3-30B on GSM8K \end{minipage}\hfill
\begin{minipage}[t]{0.26\textwidth}\centering\small (b) GPT-OSS-120B on GSM8K \end{minipage}\hfill
\begin{minipage}[t]{0.26\textwidth}\centering\small (c) Mistral-3-24B on GSM8K \end{minipage}

\begin{minipage}[t]{0.26\textwidth}\centering
\includegraphics[width=\linewidth]{graphics/traj_nR2/subj_qwen-3-30b_nR2.pdf}
\Description[Optimization trace for Qwen3-30B on Subj.]{Optimization trace for Qwen3-30B on Subj for the nR2 indicator.}
\end{minipage}\hfill
\begin{minipage}[t]{0.26\textwidth}\centering
\includegraphics[width=\linewidth]{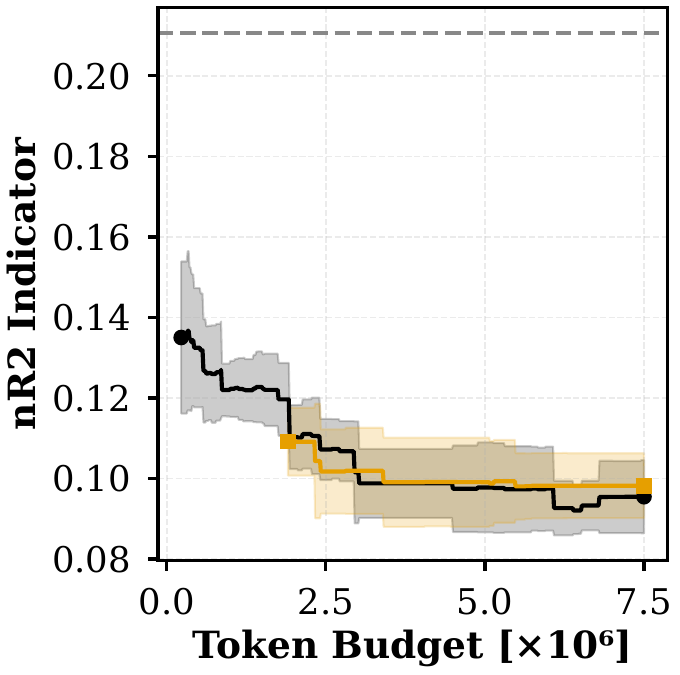}
\Description[Optimization trace for GPT-OSS-120B on Subj.]{Optimization trace for GPT-OSS-120B on Subj for the nR2 indicator.}
\end{minipage}\hfill
\begin{minipage}[t]{0.26\textwidth}\centering
\includegraphics[width=\linewidth]{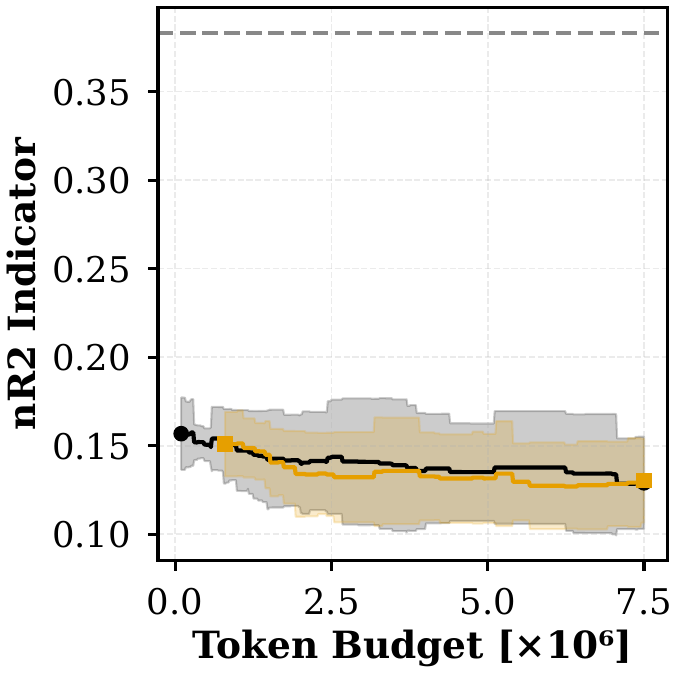}
\Description[Optimization trace for Mistral-3-24B on Subj.]{Optimization trace for Qwen3-30B on Subj for the nR2 indicator.}
\end{minipage}

\begin{minipage}[t]{0.26\textwidth}\centering\small (d) Qwen-3-30B on Subj \end{minipage}\hfill
\begin{minipage}[t]{0.26\textwidth}\centering\small (e) GPT-OSS-120B on Subj \end{minipage}\hfill
\begin{minipage}[t]{0.26\textwidth}\centering\small (f) Mistral-3-24B on Subj \end{minipage}

\begin{minipage}[t]{0.26\textwidth}\centering
\includegraphics[width=\linewidth]{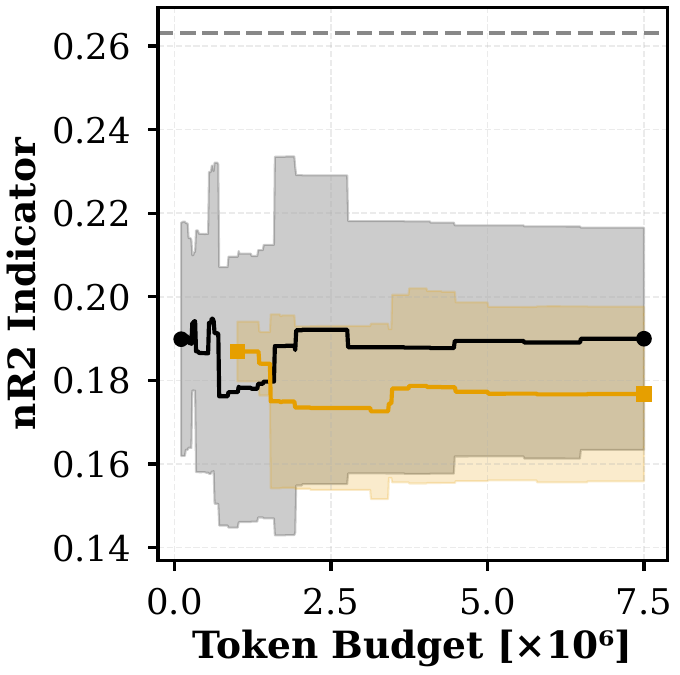}
\Description[Optimization trace for Qwen3-30B on AGNews.]{Optimization trace for Qwen3-30B on AGNews for the nR2 indicator.}
\end{minipage}\hfill
\begin{minipage}[t]{0.26\textwidth}\centering
\includegraphics[width=\linewidth]{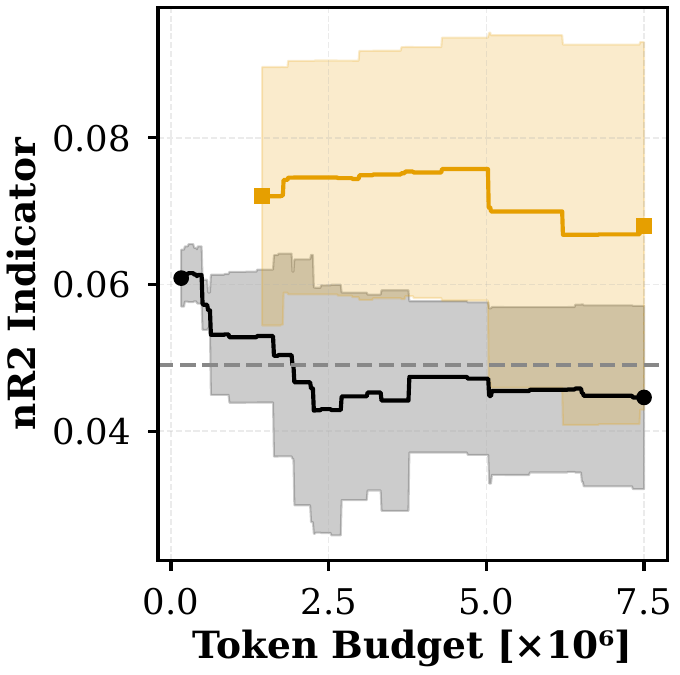}
\Description[Optimization trace for GPT-OSS-120B on AGNews.]{Optimization trace for GPT-OSS-120B on AGNews for the nR2 indicator.}
\end{minipage}\hfill
\begin{minipage}[t]{0.26\textwidth}\centering
\includegraphics[width=\linewidth]{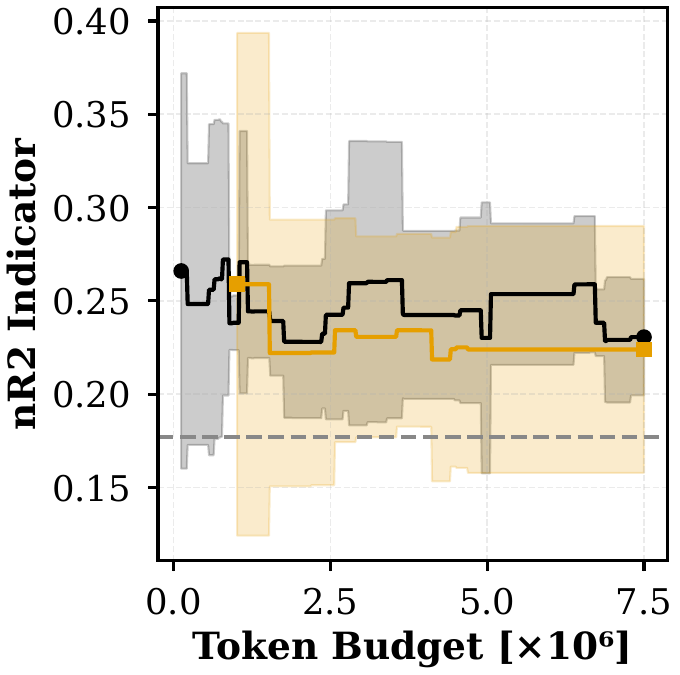}
\Description[Optimization trace for Mistral-3-24B on AGNews.]{Optimization trace for Qwen3-30B on AGNews for the nR2 indicator.}
\end{minipage}

\begin{minipage}[t]{0.26\textwidth}\centering\small (g) Qwen-3-30B on AG News \end{minipage}\hfill
\begin{minipage}[t]{0.26\textwidth}\centering\small (h) GPT-OSS-120B on AG News \end{minipage}\hfill
\begin{minipage}[t]{0.26\textwidth}\centering\small (i) Mistral-3-24B on AG News \end{minipage}

\begin{minipage}[t]{0.26\textwidth}\centering
\includegraphics[width=\linewidth]{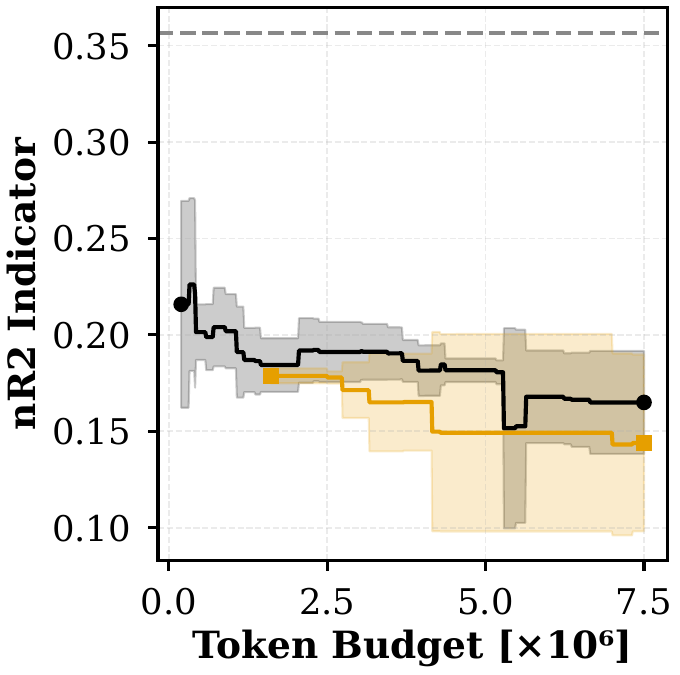}
\Description[Optimization trace for Qwen3-30B on MBPP.]{Optimization trace for Qwen3-30B on MBPP for the nR2 indicator.}
\end{minipage}\hfill
\begin{minipage}[t]{0.26\textwidth}\centering
\includegraphics[width=\linewidth]{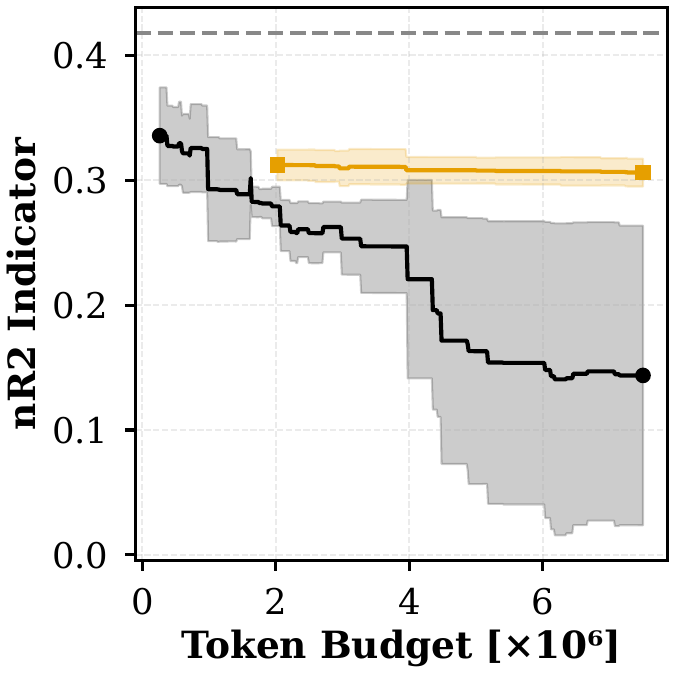}
\Description[Optimization trace for GPT-OSS-120B on MBPP.]{Optimization trace for GPT-OSS-120B on MBPP for the nR2 indicator.}
\end{minipage}\hfill
\begin{minipage}[t]{0.26\textwidth}\centering
\includegraphics[width=\linewidth]{graphics/traj_nR2/mbpp_mistral-3-24b_nR2.pdf}
\Description[Optimization trace for Mistral-3-24B on MBPP.]{Optimization trace for Qwen3-30B on MBPP for the nR2 indicator.}
\end{minipage}

\begin{minipage}[t]{0.26\textwidth}\centering\small (j) Qwen-3-30B on MBPP \end{minipage}\hfill
\begin{minipage}[t]{0.26\textwidth}\centering\small (k) GPT-OSS-120B on MBPP \end{minipage}\hfill
\begin{minipage}[t]{0.26\textwidth}\centering\small (l) Mistral-3-24B on MBPP \end{minipage}

\caption{Optimization trajectories for the nR2 metric across datasets and models. Lines and shaded regions denote mean $\pm$ std across three independent runs.}

\end{figure}

\begin{figure}[h]
\centering

\includegraphics[width=0.95\textwidth]{graphics/eaf/legend_wide.pdf}

\begin{minipage}[t]{0.26\textwidth}\centering
\includegraphics[width=\linewidth]{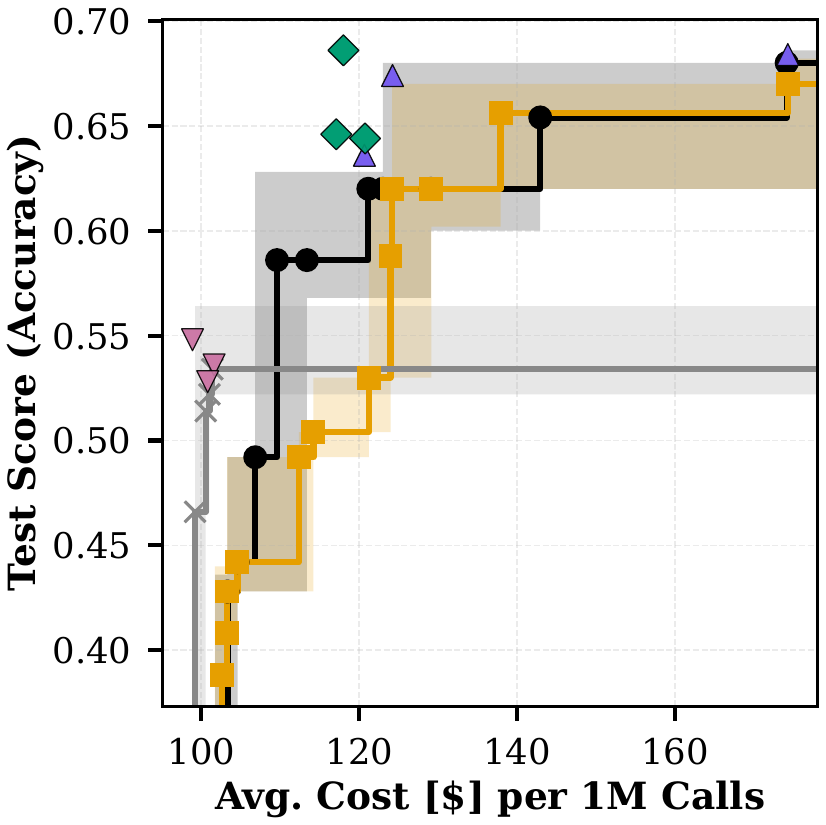}
\Description[Empirical attainment surface for Qwen3-30B on GSM8K.]{Empirical attainment surface for Qwen3-30B on GSM8K.}
\end{minipage}\hfill
\begin{minipage}[t]{0.26\textwidth}\centering
\includegraphics[width=\linewidth]{graphics/eaf/gsm8k_gpt-oss-120b_eaf.pdf}
\Description[Empirical attainment surface for GPT-OSS-120B on GSM8K.]{Empirical attainment surface for GPT-OSS-120B on GSM8K.}
\end{minipage}\hfill
\begin{minipage}[t]{0.26\textwidth}\centering
\includegraphics[width=\linewidth]{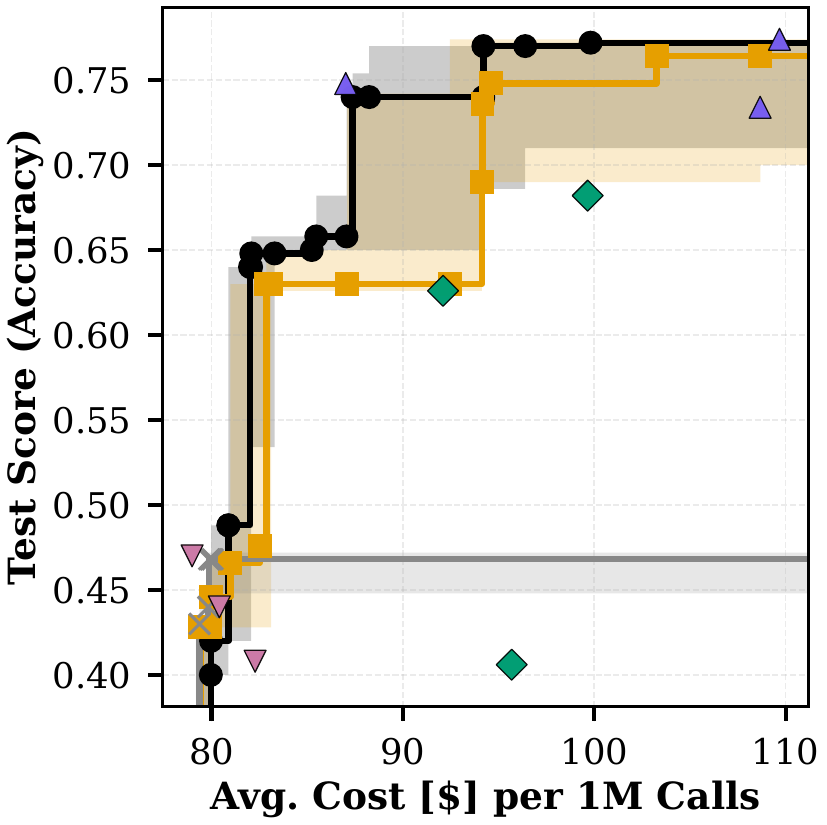}
\Description[Empirical attainment surface for Mistral-3-24B on GSM8K.]{Empirical attainment surface for Mistral-3-24B on GSM8K.}
\end{minipage}

\begin{minipage}[t]{0.26\textwidth}\centering\small (a) Qwen-3-30B on GSM8K \end{minipage}\hfill
\begin{minipage}[t]{0.26\textwidth}\centering\small (b) GPT-OSS-120B on GSM8K \end{minipage}\hfill
\begin{minipage}[t]{0.26\textwidth}\centering\small (c) Mistral-3-24B on GSM8K \end{minipage}

\begin{minipage}[t]{0.26\textwidth}\centering
\includegraphics[width=\linewidth]{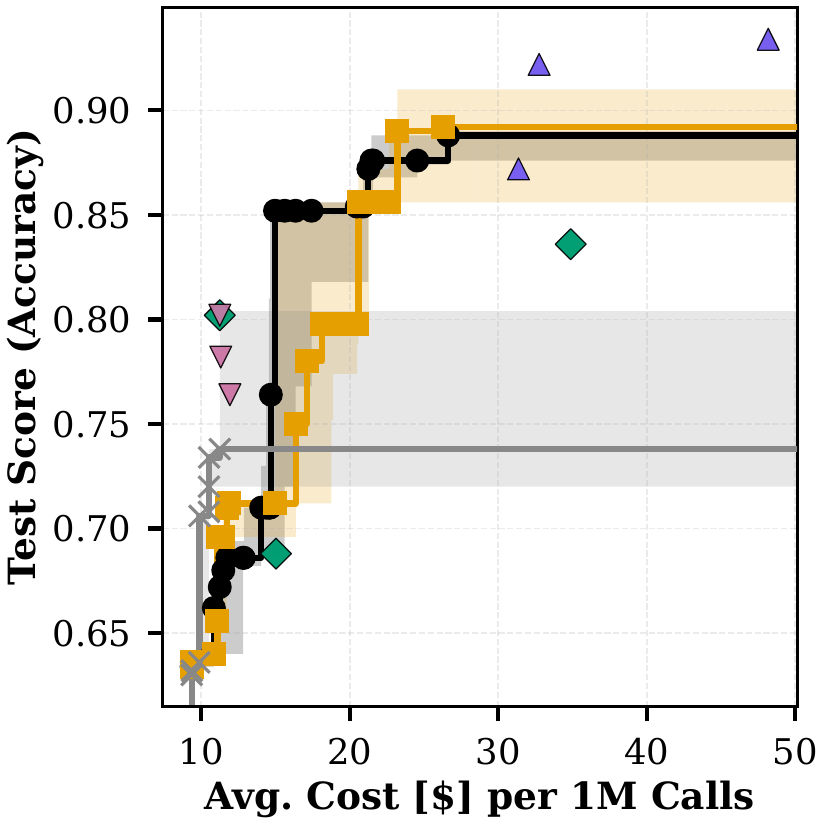}
\Description[Empirical attainment surface for Qwen3-30B on Subj.]{Empirical attainment surface for Qwen3-30B on Subj.}
\end{minipage}\hfill
\begin{minipage}[t]{0.26\textwidth}\centering
\includegraphics[width=\linewidth]{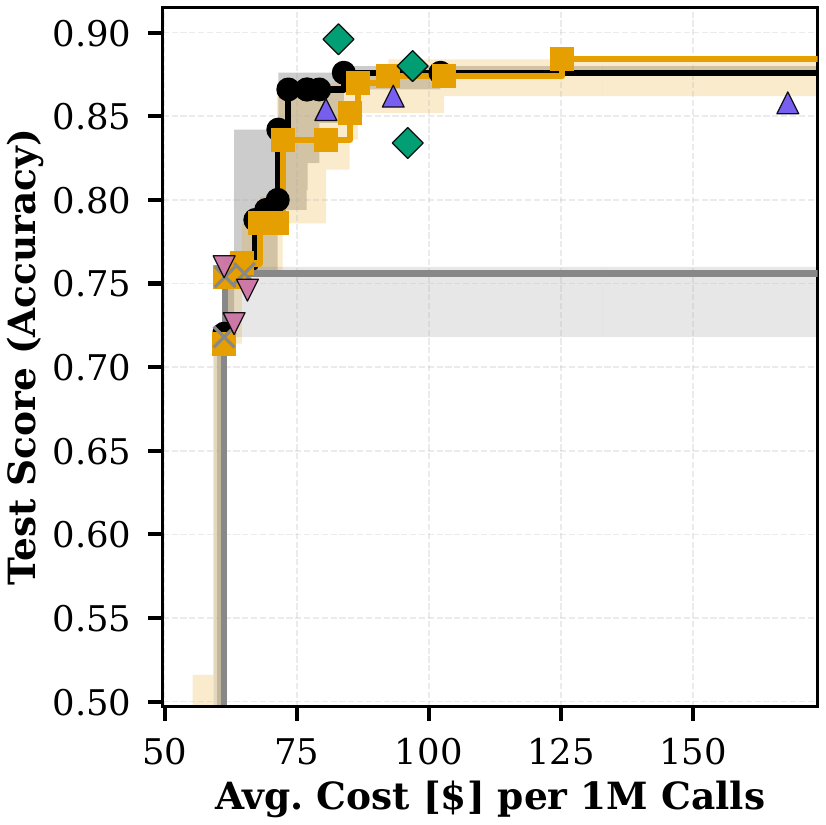}
\Description[Empirical attainment surface for GPT-OSS-120B on Subj.]{Empirical attainment surface for GPT-OSS-120B on Subj.}
\end{minipage}\hfill
\begin{minipage}[t]{0.26\textwidth}\centering
\includegraphics[width=\linewidth]{graphics/eaf/subj_mistral-3-24b_eaf.pdf}
\Description[Empirical attainment surface for Mistral-3-24B on Subj.]{Empirical attainment surface for Mistral-3-24B on Subj.}
\end{minipage}

\begin{minipage}[t]{0.26\textwidth}\centering\small (d) Qwen-3-30B on Subj \end{minipage}\hfill
\begin{minipage}[t]{0.26\textwidth}\centering\small (e) GPT-OSS-120B on Subj \end{minipage}\hfill
\begin{minipage}[t]{0.26\textwidth}\centering\small (f) Mistral-3-24B on Subj \end{minipage}

\begin{minipage}[t]{0.26\textwidth}\centering
\includegraphics[width=\linewidth]{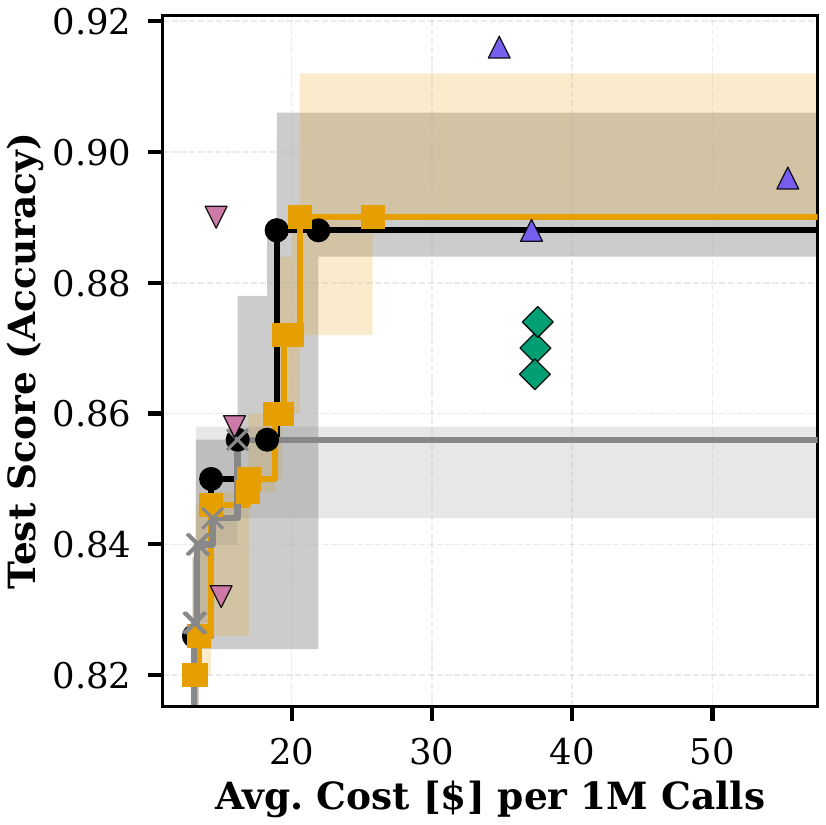}
\Description[Empirical attainment surface for Qwen3-30B on AGNews.]{Empirical attainment surface for Qwen3-30B on AGNews.}
\end{minipage}\hfill
\begin{minipage}[t]{0.26\textwidth}\centering
\includegraphics[width=\linewidth]{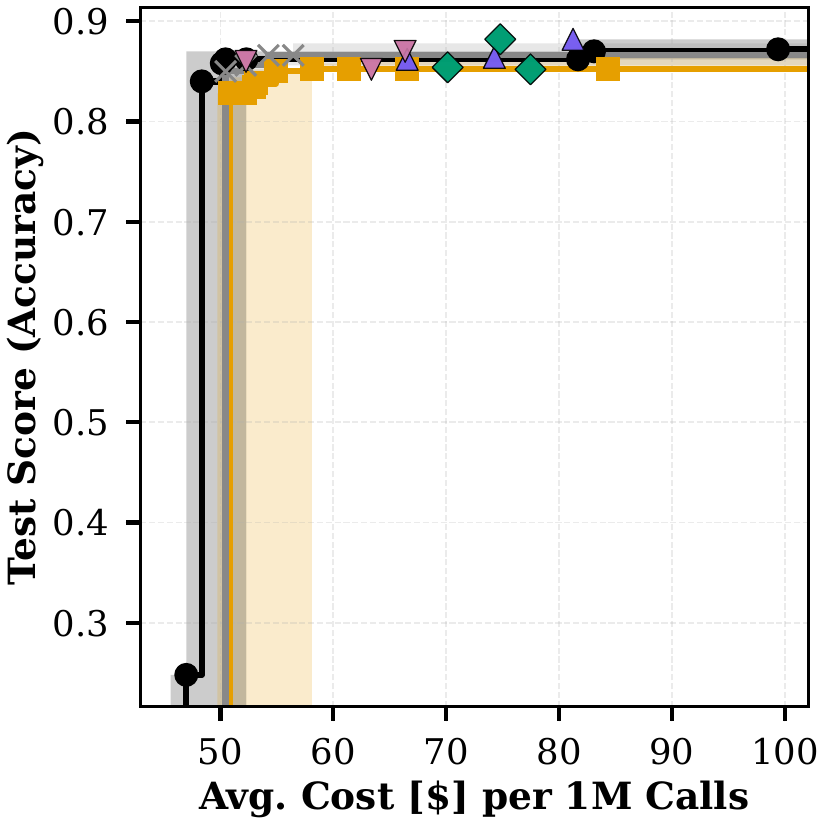}
\Description[Empirical attainment surface for GPT-OSS-120B on AGNews.]{Empirical attainment surface for GPT-OSS-120B on AGNews.}
\end{minipage}\hfill
\begin{minipage}[t]{0.26\textwidth}\centering
\includegraphics[width=\linewidth]{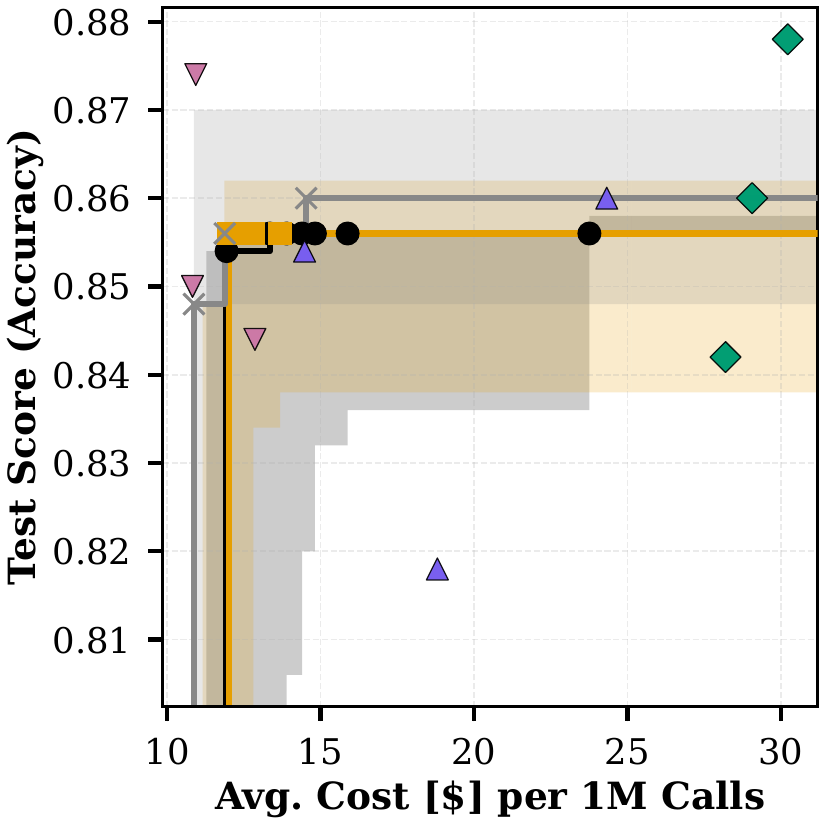}
\Description[Empirical attainment surface for Mistral-3-24B on AGNews.]{Empirical attainment surface for Mistral-3-24B on AGNews.}
\end{minipage}

\begin{minipage}[t]{0.26\textwidth}\centering\small (g) Qwen-3-30B on AG News \end{minipage}\hfill
\begin{minipage}[t]{0.26\textwidth}\centering\small (h) GPT-OSS-120B on AG News \end{minipage}\hfill
\begin{minipage}[t]{0.26\textwidth}\centering\small (i) Mistral-3-24B on AG News \end{minipage}

\begin{minipage}[t]{0.26\textwidth}\centering
\includegraphics[width=\linewidth]{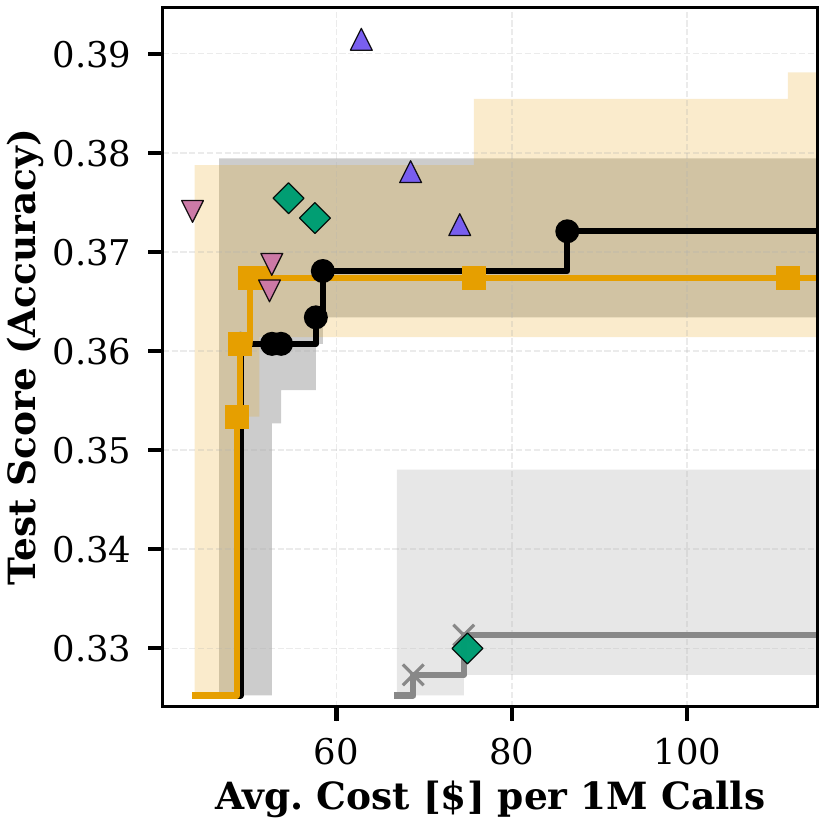}
\Description[Empirical attainment surface for Qwen3-30B on MBPP.]{Empirical attainment surface for Qwen3-30B on MBPP.}
\end{minipage}\hfill
\begin{minipage}[t]{0.26\textwidth}\centering
\includegraphics[width=\linewidth]{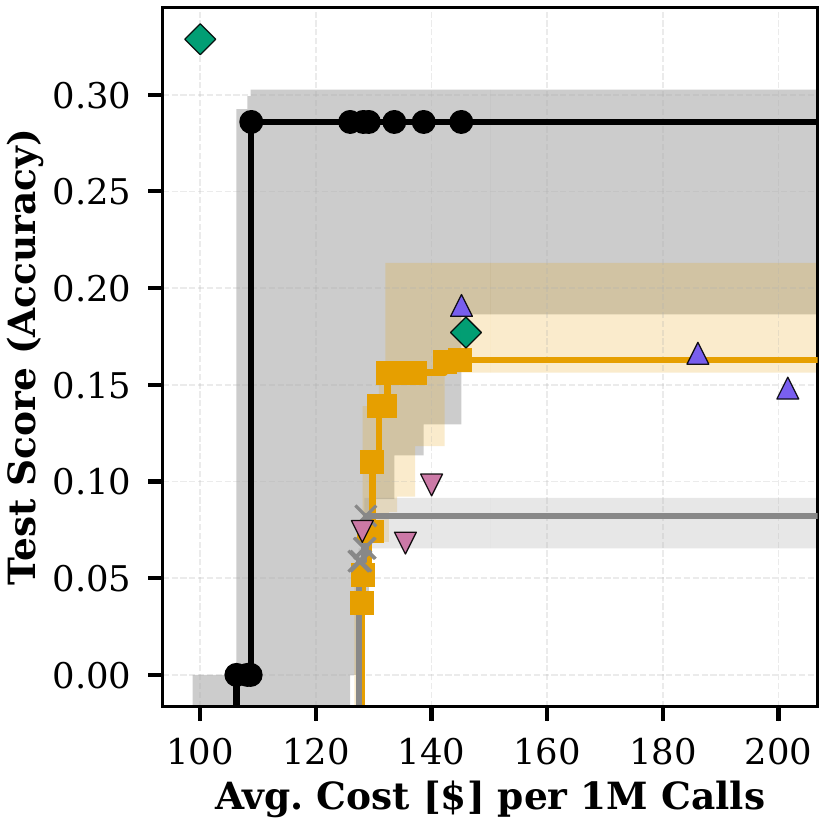}
\Description[Empirical attainment surface for GPT-OSS-120B on MBPP.]{Empirical attainment surface for GPT-OSS-120B on MBPP.}
\end{minipage}\hfill
\begin{minipage}[t]{0.26\textwidth}\centering
\includegraphics[width=\linewidth]{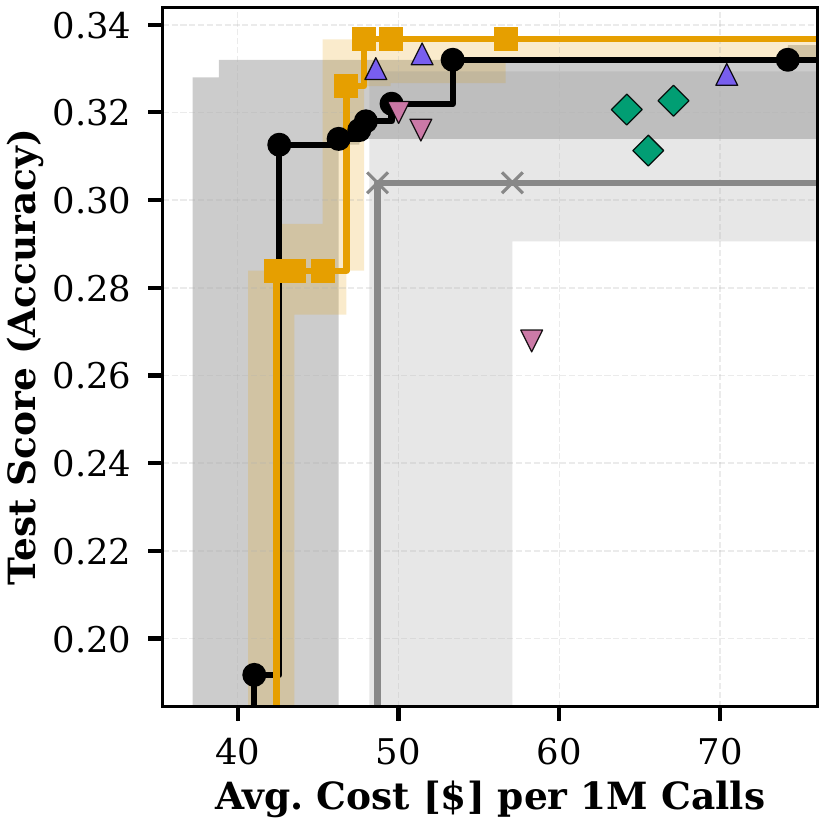}
\Description[Empirical attainment surface for Mistral-3-24B on MBPP.]{Empirical attainment surface for Mistral-3-24B on MBPP.}
\end{minipage}

\begin{minipage}[t]{0.26\textwidth}\centering\small (j) Qwen-3-30B on MBPP \end{minipage}\hfill
\begin{minipage}[t]{0.26\textwidth}\centering\small (k) GPT-OSS-120B on MBPP \end{minipage}\hfill
\begin{minipage}[t]{0.26\textwidth}\centering\small (l) Mistral-3-24B on MBPP \end{minipage}

\caption{Empirical attainment surfaces for solution sets across datasets and models. Lines indicate median attainment across three independent seeds. Shaded bands span minimum to maximum attainment.}

\end{figure}

\end{document}